\PassOptionsToPackage{table}{xcolor}
\RequirePackage{fix-cm}
\documentclass{article}
\usepackage[T1]{fontenc}
\usepackage{cellaudit_arxiv,times}
\usepackage{xcolor}
\usepackage{amsmath,amsfonts,bm}

\def\eqref#1{equation~\ref{#1}}
\def\1{\bm{1}}

\DeclareMathAlphabet{\mathsfit}{\encodingdefault}{\sfdefault}{m}{sl}
\SetMathAlphabet{\mathsfit}{bold}{\encodingdefault}{\sfdefault}{bx}{n}

\usepackage{hyperref}
\usepackage{url}
\usepackage{booktabs}
\usepackage{graphicx}
\usepackage{subcaption}
\usepackage{multirow}
\usepackage{amsmath}
\usepackage{xspace}
\usepackage{placeins}
\usepackage{float}

\definecolor{csAccent}{HTML}{176B73}
\definecolor{csAccentLight}{HTML}{EAF5F4}
\definecolor{csHeader}{HTML}{EEF2F4}
\definecolor{csReference}{HTML}{F6F7F8}
\definecolor{csCP}{HTML}{EDF4F7}
\definecolor{csL1000}{HTML}{F8F1E8}
\usepackage{array}
\usepackage{tabularx}
\usepackage{makecell}
\newcommand{\tblfont}{\fontsize{8}{9.2}\selectfont}
\newcommand{\tblcompact}{\fontsize{7.5}{8.6}\selectfont}
\newcommand{\tblsmall}{\fontsize{7}{8}\selectfont}
\newcommand{\tblsetup}{%
  \tblfont
  \setlength{\tabcolsep}{3pt}%
  \renewcommand{\arraystretch}{1.12}%
}
\newcommand{\tblpm}[2]{#1\,{\normalfont\tblsmall$\pm$\,#2}}
\newcommand{\tblci}[2]{#1\,{\normalfont\tblsmall[#2]}}
\newcommand{\tblstack}[2]{\makecell[r]{#1\\{\normalfont\tblsmall#2}}}
\newcommand{\tblhdr}[1]{\shortstack[c]{\bfseries#1}}
\newcommand{\tblnote}[1]{\par\vspace{2pt}\noindent\parbox{\linewidth}{\tblsmall #1}}
\newcolumntype{L}[1]{>{\raggedright\arraybackslash}p{#1}}
\newcolumntype{C}[1]{>{\centering\arraybackslash}p{#1}}
\newcolumntype{R}[1]{>{\raggedleft\arraybackslash}p{#1}}

\newcommand{\method}{\textsc{CellAudit}\xspace}
\newcommand{\discoverymethod}{\textsc{CellScientist}\xspace}
\newcommand{\pcc}{\textsc{PCC}\xspace}
\newcommand{\secondbest}[1]{\underline{#1}}
\newcommand{\na}{--}

\title{Discover, Falsify, Revise: Auditing Input-Use Claims from Source Code to Predictive Contribution in Agent-Discovered Cell Models}

\author{%
  Mengran Li\textsuperscript{1}\quad
  Bo Li\textsuperscript{2}\quad
  Chengyang Zhang\textsuperscript{3}\\[2pt]
  Yang Yan\textsuperscript{4}\quad
  Jinfeng Xu\textsuperscript{5}\quad
  Zhenchao Tang\textsuperscript{6}%
}
\affiliations{%
  \textsuperscript{1}Sun Yat-sen University\quad
  \textsuperscript{2}University of Macau\\
  \textsuperscript{3}Sichuan University\quad
  \textsuperscript{4}Zhejiang University\\
  \textsuperscript{5}University of British Columbia\quad
  \textsuperscript{6}Tencent AI Lab%
}

\hypersetup{hidelinks,
  pdftitle={Discover, Falsify, Revise: Auditing Input-Use Claims from Source Code to Predictive Contribution in Agent-Discovered Cell Models},
  pdfauthor={Mengran Li; Bo Li; Chengyang Zhang; Yang Yan; Jinfeng Xu; Zhenchao Tang},
  pdfsubject={Preprint}
}
\begin{document}
\maketitle

\begin{abstract}
AI virtual cells aim to predict cellular responses to specified interventions, yet held-out predictive performance alone does not establish use of the supplied perturbation information. This prediction--claim gap is especially consequential in agentic model discovery, where language-model agents generate and revise predictors using predominantly score-based feedback. We introduce \method, which audits registered input-use claims through three distinct questions: whether the input can enter the cited computation, whether fitted predictions depend on it, and whether that dependence improves prediction of observed responses. On a paired morphology--transcriptomics perturbation benchmark (BBBC047), an agent-selected predictor achieves a mean held-out Global Pearson correlation coefficient (PCC) of 0.3153 yet remains exactly invariant to compound replacement; a control-profile-only predictor reaches 0.3142. Source inspection identifies a compound-query pathway blocked by singleton key--value attention; the invariance persists after refitting with physically disjoint control wells for inputs and target references. In a stratified audit of 48 generated candidates across two linked tasks, 47 change predictions under compound replacement on both held-out folds, but only 20 show target-loss gains with intervals above zero on both folds. On BBBC047, falsification-guided revisions recover positive mean compound contributions while retaining predictive gains over the control-profile-only baseline. In matched sci-Plex searches, audit-enriched feedback shows higher mean held-out predictive performance and larger mean compound and dose contributions across five paired trajectories, although paired intervals span zero. Refitting fixed designs on an independently acquired cohort further shows predictive generalization need not imply generalization of input-use claims: dose contribution persists, whereas support for compound identity does not persist. \method therefore adds a falsification layer to agentic model discovery, moving from generate--score--revise toward discover--falsify--revise.
Project page: \url{https://limengran98.github.io/CellAudit/}.
\end{abstract}

\section{Introduction}

AI virtual cells aim to predict cellular responses to chemical and genetic interventions and ultimately support intervention design \citep{bunne2024virtualcell}. Yet a model can predict well by exploiting cellular context, control profiles, or other systematic structure while making little or no use of the supplied perturbation information \citep{ahlmanneltze2025linear,vinas2026systema}. Predictive performance alone therefore does not establish perturbation use \citep{damour2022underspecification,geirhos2020shortcut,lapuschkin2019clever,degrave2021shortcuts}. We call a mismatch between a registered claim about input use and the model's implementation or fitted behavior the \emph{prediction--claim gap}.

This gap poses an additional challenge for agentic model discovery \citep{huang2024mlagentbench,li2024automated,jiang2025aide,lu2026ai}. Systems such as CellScientist \citep{li2026cellscientist}, CellForge \citep{tang2025cellforge}, and HarmonyCell \citep{huang2026harmonycell} generate executable predictors, train them, observe validation feedback, and iteratively revise or select candidates. When feedback primarily rewards predictive performance, search can favor high-scoring predictors without testing whether they use the inputs specified by their designs. As generated models become more numerous and diverse, manual verification becomes increasingly difficult. Agentic discovery therefore needs scalable tests of the claims attached to its selected models, not only mechanisms for proposing better predictors.

We introduce \method, a framework that links registered input-use claims to evidence from source code and fitted models. It asks three non-equivalent questions: can the input enter the cited computation, do fitted predictions depend on it, and does that dependence improve prediction of the observed response? These correspond to \emph{source consumption}, \emph{fitted dependence}, and \emph{target-relevant predictive contribution}. A cited pathway can be mathematically inactive; an executable pathway may have little effect after fitting; and input dependence need not provide predictive benefit. To distinguish these cases, \method combines source inspection with matched input replacement \citep{fisher2019modelreliance,chamma2023conditional} on frozen checkpoints, testing both whole-model dependence and target-relevant contribution. During development, the same diagnostics can guide revision; final claims are evaluated only after the selected source, checkpoint, and audit rules are frozen.

\method reveals these distinctions in agent-generated cellular-response predictors. On a paired morphology--transcriptomics perturbation task built from BBBC047 \citep{bray2017cellpainting,haghighi2022profiles}, an agent-selected model reaches a mean held-out Global PCC of 0.3153, compared with 0.3142 for a control-profile-only predictor, while remaining exactly invariant to compound replacement. Source inspection identifies a singleton key--value attention pathway whose query cannot transmit compound information. The model remains invariant after refitting with physically disjoint control wells for inputs and target references. A broader audit separates fitted dependence from evidence of predictive contribution. In a stratified sample of 48 generated candidates across the two linked tasks, 47 change predictions under compound replacement on both held-out folds, but only 20 have registered target-loss scaffold intervals that lie entirely above zero on both folds.

These diagnostics also distinguish what a revision needs to address. Making a perturbation pathway executable does not by itself ensure a target-relevant contribution. On BBBC047, falsification-guided revisions learn a residual over a control-profile-only predictor, recovering positive compound contributions together with predictive gains. In matched sci-Plex searches \citep{srivatsan2020sciplex}, audit-enriched feedback produces higher mean held-out predictive performance and larger mean compound and dose contributions than score-only feedback. However, paired confidence intervals for these differences across five trajectories include zero, so the evidence for improved discovery remains suggestive rather than conclusive.

Finally, we distinguish \emph{predictive generalization} from \emph{claim generalization}. The latter asks whether a registered input-use claim remains supported when a fixed model design is refit on an independently acquired cohort. When LINCS-selected designs are fixed and refit on LKCP \citep{keenan2018lincs,subramanian2017l1000,weisbart2024gallery}, predictive gains and dose contribution persist, whereas compound-identity contribution changes substantially. A separate CRISPRa evaluation applies the same auditing principle to unseen genetic combinations \citep{norman2019genetic}. Together, these studies trace input-use claims from source code through fitted dependence to predictive contribution, use the resulting evidence to guide revision, and re-evaluate those claims on new data.

\begin{figure}[!h]
    \centering
    \includegraphics[width=\textwidth]{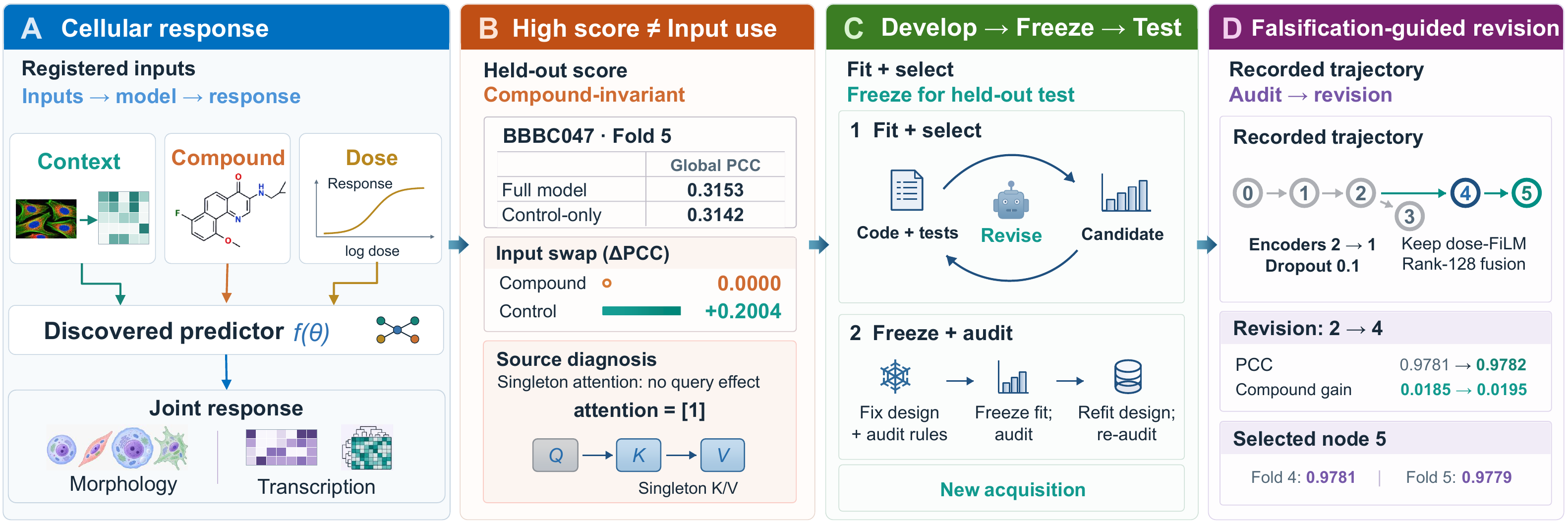}
    \caption{\textbf{Discover, falsify input-use claims, and revise.}
    (A) Registered context, compound, and dose inputs predict morphology and transcriptional responses.
    (B) \method identifies a high-scoring BBBC047 predictor with zero measured dependence on compound identity.
    (C) Development diagnostics guide revision; the selected source, checkpoint, and audit rules are frozen before held-out evaluation.
    (D) A recorded sci-Plex revision reduces encoder count while retaining dose-conditioned FiLM and rank-128 fusion. Node 0 is the shared anchor, node $2\rightarrow4$ the highlighted revision, and node 5 the final selected endpoint for held-out evaluation. Appendix~\ref{app:feedback-case} gives the complete trajectories and source changes.}
    \label{fig:overview}
\end{figure}

\section{Related work}

\paragraph{Perturbation prediction and evaluation.}
Cellular perturbation models use latent transfer, compositional representations, neural transport, genetic graphs, and foundation models to predict chemical or genetic responses \citep{lotfollahi2019scgen,lotfollahi2023cpa,bunne2023cellot,roohani2024gears,cui2024scgpt}. Benchmarks standardize evaluation across splits and baselines \citep{wu2025perturbench,wenteler2025perteval}, while recent studies examine linear references, shared response variation, calibrated metrics, and distribution shifts \citep{ahlmanneltze2025linear,vinas2026systema,miller2025calibrated,mao2026virtualcellbenchmark}. These works ask how well perturbation responses can be predicted and evaluated; \method asks whether registered input-use claims are supported by a model's source code and fitted behavior.

\paragraph{Discovery, reliance, and falsification.}
Scientific agents generate and revise executable hypotheses, programs, and models \citep{boiko2023coscientist,romeraparedes2024funsearch,jiang2025aide}. Permutation reliance, conditional replacement, and behavioral testing probe fitted-model behavior under targeted input interventions \citep{fisher2019modelreliance,chamma2023conditional,ribeiro2020checklist}, while ConceptSMILE studies explanation reliability and POPPER automates hypothesis falsification \citep{mollapour2026conceptsmile,huang2025popper}. Unlike reliance tests that begin from prediction behavior, \method starts with a registered input-use claim and its cited computation, then follows it through source consumption, fitted dependence, and target-relevant predictive contribution. This evidence can also guide model revision. Appendix~\ref{app:related-work} provides additional context.

\section{Problem setup: predictive performance and registered input-use claims}
\label{sec:setting}

Each example contains biological context $c$, perturbation representation $p$, optional attributes $a$ such as dose, and response $y=[y^{(1)},\ldots,y^{(M)}]$. Source $s$ and fitted parameters $\theta$ define $f_s(c,p,a;\theta)$; outputs may include imaging, molecular, or other readouts. Before search, the task registers input coordinates and admissible interventions. Each audit records whether an input-use claim comes from a candidate or the task contract and, after selection, fixes its cited computation and test mapping. A \emph{prediction--claim gap} occurs when the cited source contradicts a registered claim or fitted predictions are invariant to input replacement. Predictive performance affects the importance of the case, not the definition; target-relevant contribution is separate.

Data processing, budgets, evaluation, and selection rules are fixed before search. Folds 1--2 fit candidates; Fold 3 provides feedback and selects checkpoints. Selected checkpoints are evaluated on Folds 4 and 5, except prediction-score discovery also selects a source across policies on Fold 4 and refits it for Fold 5. Norman instead selects sources before Fold 4 and refits the same sources on Fold 5 without reselection. Appendices~\ref{app:protocol} and \ref{app:feedback-comparison} specify study-specific refitting and reuse rules.

\section{Method}
\label{sec:method}

\subsection{Agentic model search with fixed data and evaluation}

The existing \discoverymethod policy \citep{li2026cellscientist} starts from a common predictor $h_0$. It receives the current model, development diagnostics, execution status, and compact history, and proposes a hypothesis and executable revision within the registered permissions. Shared preflight and training code provide up to three repairs per failed proposal; persistent failures consume a slot. Each executable candidate trains once, and a fixed Fold-3 rule selects the endpoint. Records preserve parent links, emitted hypotheses and declarations, source, diagnostics, repairs, checkpoint, and costs. The language model proposes revisions; input-use tests are deterministic.

\subsection{Checking registered input-use claims in source code}

Given registered inputs and cited source locations, \method checks the computation attached to each input-use statement. Its abstract-syntax-tree checker returns \emph{implemented}, \emph{implementation-contradicted}, or \emph{unresolved}. Coverage is the fraction of checked candidates whose cited locations are all resolved; unresolved code remains eligible for behavioral testing. With one key--value pair, attention has constant normalized weight and cannot transmit compound information through its query \citep{vaswani2017attention}. This localizes a recognized implementation defect. Whole-model replacement then establishes whether fitted predictions depend on the input, including through other routes; target-loss contrasts test whether that dependence helps prediction.

\subsection{Testing fitted dependence and predictive contribution}
\label{sec:audit}

With checkpoint and targets fixed, \method replaces one registered input $x_q$ while retaining the others. Prediction distance measures target-blind dependence; zero identifies invariance on the tested replacements. To test predictive benefit, it computes the decrease in a higher-is-better score $m$ and the increase in loss:
\begin{equation}
\Delta m_q=m(f(x),y)-m(f(x_{-q},\tilde x_q),y),\quad
\Delta\mathcal L_q=\mathcal L(f(x_{-q},\tilde x_q),y)-\mathcal L(f(x),y).
\end{equation}
Positive effects indicate benefit from the observed input relative to its registered replacements. Replacement distributions and their observed-support coverage are specified by task. We distinguish single-coordinate contrasts $\Delta m_q$ from factorial effects $E_q$, which average over the other input's two states in the $2\times2$ compound--context audit. Loss effects $E^{\mathcal L}_q$ use the same averaging. The transfer analysis separates compound and dose allocations \citep{lundberg2017shap}. Appendix~\ref{app:factorial-definition} gives the full contrasts.

\paragraph{Effect size, uncertainty, and stability.}
Maps are repeated measurements within a checkpoint, averaged before across-model inference. Continuous contributions have paired training-seed or trajectory intervals; fixed-model source-group resampling tests variation across biological groups. Known-truth checks distinguish fixed-dataset and population uncertainty (Appendix~\ref{app:measurement-validity}). BBBC, LINCS, and LKCP separately report whether the effect quantile interval lies \emph{above}, \emph{below}, or \emph{across} a reference threshold. A positive mean contribution can occur in any category. Appendix~\ref{app:protocol} defines the threshold, quantiles, and original category labels; Norman uses a separate loss-based criterion (Appendix~\ref{app:norman-transfer}).

\subsection{Using falsification evidence to guide the next search}

The audit distinguishes two reasons to revise a model. A contradicted source path motivates an executable route; weak target contribution motivates learning an increment over context. Path-constrained discovery addresses the first by compiling structured design cards into models with an explicit compound route. BBBC falsification-guided discovery addresses the second through residual modeling and input-use feedback:
\begin{equation}
\hat y=g(c)+\alpha h(c,p,a),
\end{equation}
where $g(c)$ is fitted on Folds 1--2 and fixed. The compiler enforces centered perturbation/dose branches, context gates, and bias-free readouts, giving zero residual at their joint reference input. Training and Fold-3 feedback use prediction error and replacement loss changes. BBBC047 selects by predictive score among qualifying candidates; BBBC036 prioritizes compound loss gain among predictively noninferior candidates. Each trajectory evaluates the anchor plus nine proposals; the harness selects scale $\alpha$ and checkpoint on Fold 3 and freezes that model for Folds 4 and 5. These formulations test the joint design changes; Appendix~\ref{app:protocol} specifies their rules.

\paragraph{Separating feedback from changes to the model space.}
The sci-Plex comparison instead shares a context-plus-dose anchor $g(c,a)$ and permissions for direct predictors or anchor residuals. Score feedback supplies metrics and learning curves; Audit adds grouped errors, source checks, and development-set compound/dose replacements. After each fit, \method computes these diagnostics and inserts them into the agent's next prompt, linking input-path findings and measured contributions to model revision. Loss, optimizer, training limits, and PCC-first selection stay fixed. Paired outcomes evaluate the complete audit-enriched feedback package; selected sources and checkpoints are frozen before both held-out evaluations.

\section{Experimental design}
\label{sec:experiments}

\subsection{Development on linked multimodal perturbation tasks}

Development uses linked tasks from one paired cellular-profile release: BBBC036, the known-bioactive subset, and its parent BBBC047 \citep{haghighi2022profiles,bray2017cellpainting,subramanian2017l1000}. Each predicts joint morphology--transcription responses from a same-plate control profile, compound representation, and dose. Response-blind Bemis--Murcko scaffold folds \citep{bemis1996frameworks} hold out compound families. Global \pcc on concatenated responses selects models; MSE and modality-specific metrics diagnose gains. Appendix~\ref{app:data} gives provenance, transformations, fold inventories, and a complementary plate-family split.

Prediction-score discovery compares CellScientist, AIDE, CellForge, and HarmonyCell policies \citep{li2026cellscientist,jiang2025aide,tang2025cellforge,huang2026harmonycell} under a matched initial model, executor, feedback, and ten-candidate budget, with ten trajectories per task and five paired refits per selected model. Both follow-up formulations retain the folds and budget and carry all ten selected designs through Folds 4 and 5; failures, costs, and repeated designs are retained (Appendix~\ref{app:receipts}).

\paragraph{Separating the control input from the target reference.}
For BBBC047, each plate's control wells are split into equal-size banks A/B. With the target reference fixed, shared and disjoint arms take the input profile from the same or the other bank, so their contrast isolates physical control-well overlap. Both orientations refit the score-selected source, a control-only source, and one fixed guided design under five seeds; all 60 checkpoints are fixed before evaluation on Folds 4 and 5 (Appendix~\ref{app:control-decoupling}).

\paragraph{Matched feedback on sci-Plex.}
We predict 2,000-gene well-level responses in sci-Plex3 \citep{srivatsan2020sciplex}, holding out compound--cell-line combinations, with whole treatment wells in one partition, while both marginals occur in training; gene selection and scaling use training wells only. Both conditions use the same DeepSeek serving model, five paired trajectory seeds, five new candidate slots per trajectory, and a shared anchor. All ten selected checkpoints are evaluated on two held-out combination partitions of this source. Compound replacements match dose and experimental context; dose replacements retain compound identity (Appendix~\ref{app:feedback-comparison}).

\subsection{Testing input-use claims on new cohorts and perturbation types}

The fixed falsification-guided protocol first runs on LINCS--Pilot1 \citep{keenan2018lincs,subramanian2017l1000}. Its ten selected designs and sources are then frozen and refit on independently acquired LKCP Batch 2 \citep{weisbart2024gallery}, with response- and fold-blind interface adaptation fixed before training. Compound permutations retain dose; dose permutations stay within compound (Appendix~\ref{app:external-transfer}).
The Norman CRISPRa task predicts 5,045-gene responses for unseen double-perturbation combinations \citep{norman2019genetic}. Its registered input bundles multi-hot pair identity with the matching mean single-perturbation anchor. GEARS supplies a task-native reference \citep{roohani2024gears}.

\section{Results}
\label{sec:results}

\subsection{\method detects compound invariance in a high-scoring predictor}

\method's automated audit identifies the BBBC047 source selected on Fold~4 as compound-invariant. Its source-selection score is $0.3035$; mean Global \pcc reaches $0.3153$ on held-out replication Fold~5. Compound shuffling changes \pcc by $0.0000$ in every refit, whereas control replacement lowers \pcc by $0.1932$ on Fold~4 and $0.2004$ on Fold~5 (Table~\ref{tab:main-evidence-ladder}). Control-only reaches $0.3142$ on Fold 5; Full$-$control-only is $+0.0011$ (95\% CI $[-0.0005,+0.0027]$). Prediction-space distance is exactly zero under compound replacement and large under control replacement (Appendix~\ref{app:prediction-space}).

The source checker identifies a concrete computation to revise: the cited compound-conditioned cross-attention has one key and one value, making its normalized weight constant. The behavioral test establishes complete-model invariance; the source result localizes an inactive cited pathway. A compound-only positive control produces a shuffle effect of $0.0111\pm0.0010$, with all five refits above the reference threshold.

A source-stratified sample of 48 real candidates separates these outcomes further (Figure~\ref{fig:candidate-audit-story}a,b). An implemented multi-tower source is exactly invariant; 47 candidates show fitted dependence on both folds, but only 20 have positive target-loss scaffold intervals on both. Source consumption and fitted dependence thus leave predictive contribution unresolved for many candidates (Appendix~\ref{app:real-source-behavior}). \discoverymethod supplies a competitive discovery setting: mean Best@10 and frontier AUC lead both linked tasks, with intervals versus HarmonyCell crossing zero (Table~\ref{tab:main-discovery}).

\begin{table}[H]
\centering
\caption{\textbf{Prediction, source implementation, and target-relevant compound use are distinct.} BBBC047: post-search Fold 4 and held-out replication Fold 5. Full/Control are mean Global PCC. $E_{\rm cmpd}$ and $E_{\rm ctrl}$ are factorial PCC effects averaged over the other input's two states; $E^{\mathcal L}_{\rm cmpd}$ is the corresponding loss effect. Reference counts locate effect quantile intervals relative to their thresholds. Rows summarize five prediction-score refits or ten endpoints per constrained search.}
\label{tab:main-evidence-ladder}
\tblsetup
\renewcommand{\arraystretch}{1.0}
\setlength{\tabcolsep}{2.2pt}
\begin{tabular}{@{}L{68pt}L{49pt}C{21pt}R{29pt}R{30pt}R{29pt}R{34pt}R{29pt}C{72pt}@{}}
\toprule
\rowcolor{csHeader}
\textbf{Search} & \textbf{Code check} & \textbf{Fold} & \textbf{Full} & \textbf{Control} & $E_{\rm cmpd}$ & $E^{\mathcal L}_{\rm cmpd}$ & $E_{\rm ctrl}$ & \tblhdr{Reference\\Above/Below/Across} \\
\midrule
\multirow{2}{*}{Prediction-score} & \multirow{2}{*}{Contradicted$^{a}$} & 4 & 0.3035 & 0.3019 & 0.0000 & 0.00000 & 0.1932 & 0/5/0 \\
 & & 5 & 0.3153 & 0.3142 & 0.0000 & 0.00000 & 0.2004 & 0/5/0 \\
\addlinespace[2pt]
\multirow{2}{*}{Path-constrained} & \multirow{2}{*}{Implemented$^{b}$} & 4 & 0.2822 & \na & 0.0083 & 0.00024 & 0.1388 & 0/6/4 \\
 & & 5 & 0.2922 & \na & 0.0074 & 0.00022 & 0.1477 & 0/7/3 \\
\addlinespace[2pt]
\rowcolor{csAccentLight} & & 4 & 0.3060 & 0.3023 & 0.0053 & 0.00018 & 0.1988 & 0/10/0 \\
\rowcolor{csAccentLight} \multirow{-2}{*}{Falsification-guided} & \multirow{-2}{*}{Implemented$^{c}$} & 5 & 0.3173 & 0.3145 & 0.0048 & 0.00017 & 0.2051 & 0/10/0 \\
\bottomrule
\end{tabular}
\tblnote{Control: control-only predictor. Reference: effect quantile interval above/below/across the reference threshold (Appendix~\ref{app:protocol}); distinct from a positive mean contribution. $^{a}$One-key attention cannot transmit compound identity. $^{b}$Compiler-enforced compound path. $^{c}$Explicit compound-residual path.}
\end{table}

\subsection{The mismatch persists with disjoint control references}
\label{sec:control-separation}

\begin{figure}[!t]
    \centering
    \input{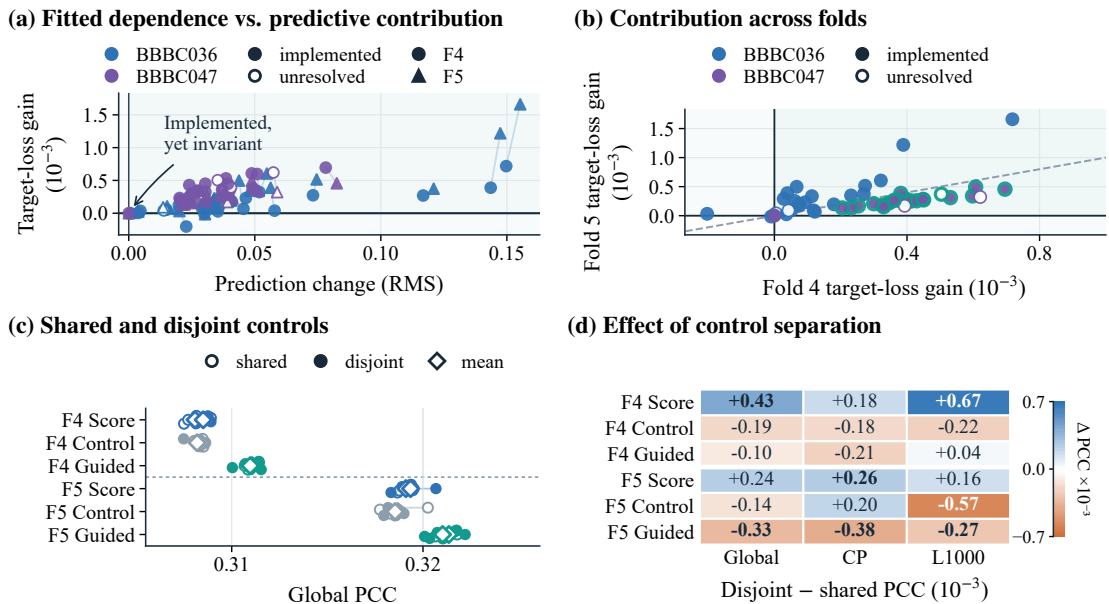}
    \captionsetup{skip=4pt}
    \caption{\textbf{Fitted dependence and predictive contribution are distinct.} (a,b) Forty-eight sources across the two linked tasks; filled/hollow markers denote implemented/unresolved paths, and green outlines mark the 20/48 sources with positive target-loss scaffold intervals on both folds. Lines in (a) pair folds. (c) Five paired seeds, averaging control-bank orientations within seed. (d) Mean disjoint-minus-shared PCC.}
    \label{fig:candidate-audit-story}
\end{figure}

We test whether the mismatch persists when control inputs and CP target references use disjoint physical wells. Newly reconstructed shared and disjoint arms have identical targets within each bank orientation (Figure~\ref{fig:candidate-audit-story}c,d). In the disjoint arm, the score-selected source reaches Global \pcc $0.3085$ and $0.3193$, while control-only reaches $0.3082$ and $0.3185$ on Folds 4 and 5. The score-selected source remains exactly compound-invariant in every refit and evaluation. All six shared--disjoint PCC comparisons have five-seed intervals crossing zero. The detected mismatch therefore persists without physical control-well reuse at this reference-construction layer.

The fixed guided design retains both predictive and compound-use increments under the same separation (Table~\ref{tab:bbbc-control-decoupling}). Relative to disjoint control-only, its PCC gain is $+0.0028$ ($95\%$ paired-seed CI $[+0.0017,+0.0038]$) on Fold 4 and $+0.0025$ ($[+0.0012,+0.0037]$) on Fold 5. Compound replacement lowers PCC by $0.0043$ and $0.0033$ and increases target loss by $1.5106\times10^{-4}$ and $1.1939\times10^{-4}$; each mean effect has a positive five-seed interval. Effects are slightly smaller than in the shared arm. Same-plate context and the original scaffold folds are retained (Appendix~\ref{app:control-decoupling}).

\subsection{An explicit compound route and its fitted contribution}

Path-constrained discovery addresses the source defect by compiling an explicit compound route: all 200 candidate evaluations pass source and interface checks without candidate-code repair. On BBBC047, mean factorial compound effects are $0.0083$ and $0.0074$, with positive mean loss effects on Folds 4 and 5. Relative to the reference threshold, Fold 4 has six below-threshold and four overlapping cases; Fold 5 has seven and three. Control-profile effects exceed the threshold throughout (Table~\ref{tab:main-evidence-ladder}). An executable route therefore admits compound dependence without ensuring a consistently above-reference contribution in each model.

\subsection{Guided residual discovery combines predictive gains with target-relevant compound contribution}

Falsification-guided discovery next asks the perturbation residual to improve prediction beyond a fixed control-profile-only model. On BBBC047, the full predictor exceeds that baseline by $+0.0037$ on Fold 4 and $+0.0028$ on Fold 5. Mean factorial compound effects are $0.0053$ and $0.0048$, with positive loss effects (Table~\ref{tab:main-evidence-ladder}); joint model--scaffold intervals for these increments are above zero. The explicit-route models have larger compound effects but lower predictive scores. Guided discovery combines higher scores with positive target-relevant effects, although none of its ten models exceeds the reference-relative criterion.

For the original guided checkpoints, observed-support replacements retain positive contributions on BBBC047 while holding context and dose fixed. They cover $3341/4380$ and $2927/3878$ rows on Folds 4 and 5: compound PCC drops are $0.00534$ and $0.00404$, and loss gains are $1.84\times10^{-4}$ and $1.44\times10^{-4}$, all with positive source-scaffold intervals. The score-selected model remains exactly invariant on these same eligible rows (Table~\ref{tab:observed-support-replay}).

Fixed residual learners clarify the trade-off: Ridge has strong compound sensitivity but low joint-response scores, and on BBBC047 the residual MLP has larger compound effects but lower scores than the guided models, which lead both references in mean Global \pcc at all four BBBC evaluations. On BBBC036 Fold 5, their gain over the context anchor is $-0.0020$, with model--scaffold intervals for predictive and compound increments crossing zero (Table~\ref{tab:d3-scaffold-bootstrap}).

\paragraph{Checking what these effects measure.}
Known-function experiments separate invariance, sensitivity without benefit, benefit, and harm across 6,400 independent simulated datasets. On the sensitive-without-benefit null, map-only intervals attain near-nominal coverage for the fixed-dataset conditional effect but severely undercover the population effect. Source-group intervals improve population coverage, with finite-group undercoverage remaining. Holding compound benefit fixed while changing reference-context variation can also change qualification. These checks support reporting continuous contribution, across-model uncertainty, and reference-relative status as distinct results (Appendix~\ref{app:measurement-validity}).

\subsection{Audit-enriched feedback yields higher mean held-out prediction}
\label{sec:feedback-results}

\begin{figure}[!t]
    \centering
    \input{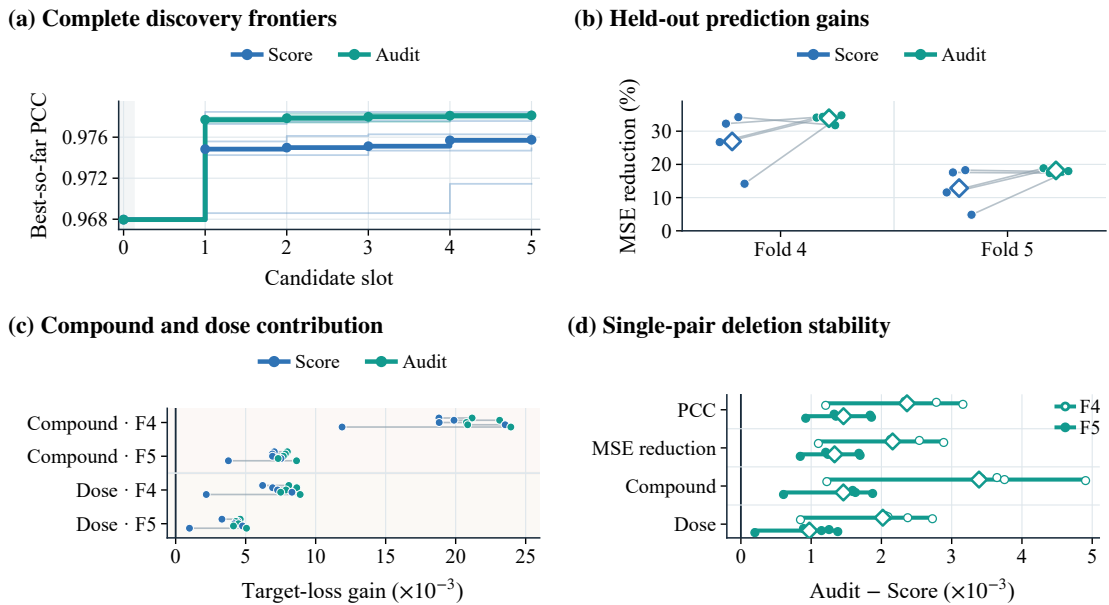}
    \captionsetup{skip=4pt}
    \caption{\textbf{Audit feedback connects search progress to held-out contributions.} (a) Five trajectories per condition and their mean frontiers; slot 0 is the anchor. (b) Relative MSE reduction against each paired anchor. (c) Compound/dose loss gains; lines pair trajectories. (d) Leave-one-pair-out means for the absolute Audit--Score differences; the MSE row is shown as a favorable reduction, so its scale is not the relative percentage scale in (b). Bars span the resulting means, not confidence intervals; diamonds summarize all five pairs.}
    \label{fig:feedback-paired-evidence}
\end{figure}

Do audit measurements help when returned to an agent? Figure~\ref{fig:feedback-paired-evidence}a retains all five paired sci-Plex search frontiers. The context-plus-dose anchor already predicts much of the whole-expression profile, making incremental error reduction informative. Audit feedback yields higher mean PCC and lower mean MSE on both later partitions (Figure~\ref{fig:feedback-paired-evidence}b): MSE is $9.58\%$ lower than Score on Fold 4 and $6.10\%$ lower on Fold 5. Both improve on the anchor.

The higher mean predictive performance of Audit is accompanied by higher mean compound and dose contributions (Figure~\ref{fig:feedback-paired-evidence}c). Compound target-loss gain rises from $0.0186$ to $0.0220$ on Fold 4 and from $0.0064$ to $0.0079$ on Fold 5. Paired $t$ intervals for these Audit--Score contribution contrasts include zero at both folds (Table~\ref{tab:feedback-all-contrasts}). Correct inputs help prediction in all five endpoints of both conditions.
Audit--Score PCC is $+0.0024$ (paired $95\%$ CI $[-0.0013,+0.0060]$) on Fold 4 and $+0.0015$ ($[-0.0005,+0.0034]$) on Fold 5. Both means, and the favorable MSE and compound loss-gain directions, persist after removing any one trajectory pair (Figure~\ref{fig:feedback-paired-evidence}d), whereas prediction-RMS differences can reverse after one omission. Percentile-bootstrap predictive intervals exclude zero even when resampling only trajectories, so uncertainty depends on interval construction at five pairs (Appendix~\ref{app:feedback-robustness}). Both conditions complete all 25 new candidates; Audit uses 38 versus 36 logical model calls and 592,347 versus at least 286,995 reported tokens (Table~\ref{tab:feedback-costs}).

\paragraph{From automatic audit output to a recorded revision.}
In the first registered Audit trajectory, \method returns source checks, compound/dose replacement effects, grouped errors, and learning curves for the bilinear incumbent and a non-improving challenger. The agent returns to that incumbent and proposes simpler encoders with dropout, retaining dose-conditioned FiLM. Parameters decrease while selection PCC and compound loss gain increase (Figure~\ref{fig:overview}D); search continues to a distinct final endpoint (Appendix~\ref{app:feedback-case}).

\subsection{Predictive gains and input-use support generalize differently under independent acquisition}

\begin{figure}[!t]
    \centering
    \input{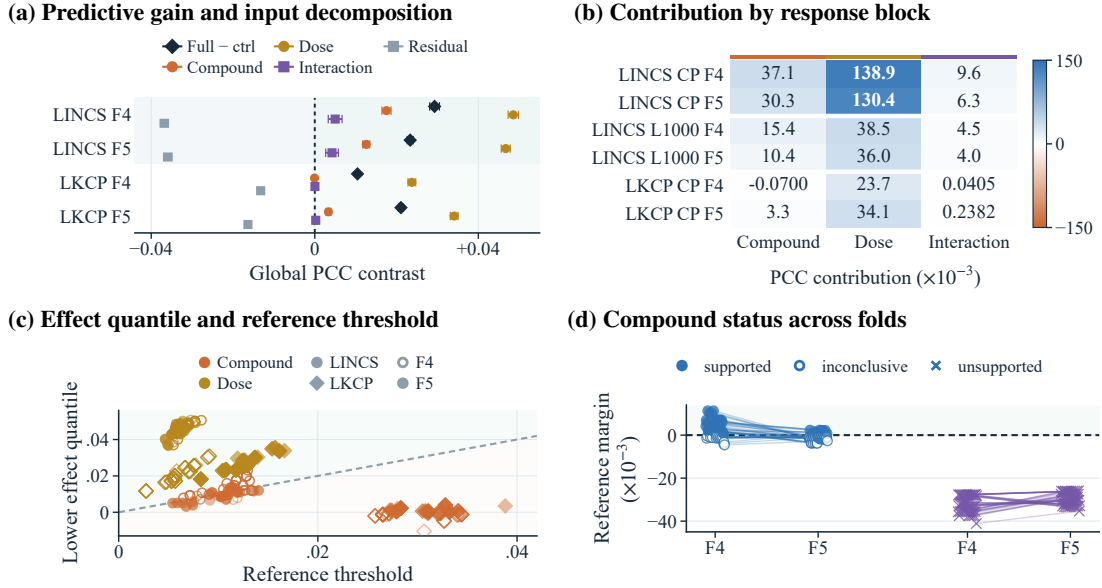}
    \captionsetup{skip=4pt}
    \caption{\textbf{Predictive and input-use evidence generalize differently across acquisitions.} (a) Means and 95\% intervals over ten endpoint instances; interaction is separate from the additive allocations. (b) LINCS CP/L1000 and LKCP morphology. (c) Lower effect quantiles and thresholds for 400 input--checkpoint evaluations. (d) Paired compound reference margins; the left pair is LINCS and the right pair is LKCP, with lines joining the same endpoint--seed refit across folds. Marker states denote supported, inconclusive, and unsupported refits. Each cohort has ten endpoint instances with five refits each.}
    \label{fig:coordinate-input-use}
\end{figure}

To test claim generalization, the frozen LINCS-selected designs are refit on LKCP without source revision, model reselection, or audit-rule changes. They improve over the control-profile-only baseline by $+0.0104$ (95\% CI $[+0.0103,+0.0106]$) on Fold 4 and $+0.0211$ (95\% CI $[+0.0205,+0.0217]$) on Fold 5.
Figure~\ref{fig:coordinate-input-use}a,b separates compound identity from dose. On Folds 4 and 5, the dose-replacement Shapley allocation is $+0.0486$ and $+0.0467$ Global \pcc on LINCS and $+0.0237$ and $+0.0341$ on LKCP, whereas the compound-identity allocation is $+0.0175$ and $+0.0126$ on LINCS but $-0.000070$ and $+0.0033$ on LKCP. All 50 trajectory-endpoint refits pass the dose-use criterion at every cohort--fold boundary, but none passes the compound-identity criterion at either LKCP boundary.
LINCS shows why average effects and per-model decisions must remain separate: its compound-identity criterion is passed by 35/50 refits on Fold 4 but 10/50 on Fold 5 (mean lower effect quantile $0.0139$ to $0.0090$), although average compound contribution remains positive.

\subsection{The audit detects dependence on the perturbation-pair input bundle}

On Norman, all five \discoverymethod refits pass the registered pair-input-bundle criterion on both held-out folds; shuffling the multi-hot pair identity together with its matching single-perturbation anchor reduces \pcc by $0.5578$ and $0.6280$. Mean \pcc exceeds $h_0$ by $+0.0060$ and $+0.0045$ and the observed-single additive baseline by $+0.0209$ and $+0.0114$ (Table~\ref{tab:main-norman}; Appendix~\ref{app:norman-transfer}).

\section{Discussion and conclusion}

\method shows that predictive success and claimed input use should be evaluated separately. Source consumption, fitted dependence, and target-relevant contribution can diverge: on BBBC047, a high-scoring source remains compound-invariant after disjoint control-reference refitting, while falsification-guided revisions recover positive compound contributions with predictive gains. The broader candidate audit shows this is not specific to singleton attention. In sci-Plex, audit-enriched feedback improves mean prediction and input contributions, but paired intervals over five trajectories span zero. Independent-acquisition refits further show that predictive generalization need not imply claim generalization.

These conclusions are conditional on the fitted model, evaluation setting, and registered replacement distribution. Positive contribution is distinct from passing the reference-relative criterion, and support for one fit should not be inherited by later refits. Observed-support replacements avoid unobserved combinations but do not establish biological exchangeability, mechanism, or causal effects. The sci-Plex study evaluates the complete audit-feedback package rather than individual diagnostics. \method therefore adds a falsification layer to agentic model discovery, tracing input-use claims from cited computation to fitted behavior and predictive contribution.

\FloatBarrier
\clearpage

\bibliographystyle{iclr2027_conference}
\bibliography{references}

\appendix
\makeatletter
\setlength{\@fptop}{0pt}
\setlength{\@fpsep}{14pt plus 2pt minus 2pt}
\setlength{\@fpbot}{0pt plus 1fil}
\makeatother
\renewcommand{\floatpagefraction}{0.8}
\setcounter{topnumber}{5}
\setcounter{bottomnumber}{5}
\setcounter{totalnumber}{10}
\raggedbottom
\setlength{\intextsep}{4pt plus 1pt minus 1pt}
\setlength{\textfloatsep}{8pt plus 2pt minus 2pt}
\setlength{\abovecaptionskip}{6pt}
% Appendix-only presentation; main-text typography is unchanged.
\makeatletter
\setlength{\@fptop}{0pt}
\setlength{\@fpsep}{10pt plus 1pt minus 1pt}
\setlength{\@fpbot}{0pt plus 1fil}
\makeatother
\renewcommand{\topfraction}{0.98}
\renewcommand{\bottomfraction}{0.95}
\renewcommand{\textfraction}{0.02}
\renewcommand{\floatpagefraction}{0.80}
\setcounter{topnumber}{8}
\setcounter{bottomnumber}{8}
\setcounter{totalnumber}{16}
\setlength{\floatsep}{8pt plus 1pt minus 1pt}
\setlength{\intextsep}{7pt plus 1pt minus 1pt}
\setlength{\textfloatsep}{9pt plus 1pt minus 1pt}
\captionsetup[table]{font=small,skip=4pt}
\renewcommand{\tblfont}{\fontsize{8}{9.2}\selectfont}
\renewcommand{\tblcompact}{\tblfont}
\renewcommand{\tblsmall}{\fontsize{7.5}{8.6}\selectfont}
\renewcommand{\tblsetup}{\tblfont\setlength{\tabcolsep}{3pt}\renewcommand{\arraystretch}{1.06}}
\newcommand{\tblinlinepm}[2]{#1\,{\normalfont\tblsmall #2}}
\newsavebox{\AppendixTableBox}
\newcommand{\AppendixTableOutput}[1]{%
  \typeout{APPENDIX-TABLE #1 WIDTH=\the\wd\AppendixTableBox LIMIT=\the\linewidth}%
  \ifdim\wd\AppendixTableBox>\dimexpr\linewidth+0.5pt\relax
    \typeout{APPENDIX-TABLE-OVERWIDTH #1}%
  \fi
  \usebox{\AppendixTableBox}}

\FloatBarrier
\section{Data provenance, transformations, and task statistics}
\label{app:data}

BBBC036 is the known-bioactive subset of its parent BBBC047 screen; the two tasks are therefore linked cohorts from one paired cellular-profile release rather than independent acquisitions. Their source collections in \texttt{cpg0003-rosetta} are CDRPBIO-BBBC036-Bray and CDRP-BBBC047-Bray, and the abbreviated task names retain the formal BBBC accession mapping \citep{ljosa2012bbbc}. Inputs come from released replicate-level Cell Painting and L1000 profile tables \citep{weisbart2024gallery,haghighi2022profiles}. The morphology screen originates from the CDRP/BBBC047 Cell Painting resource \citep{bray2017cellpainting}, and the transcriptional profiles use the L1000 platform \citep{subramanian2017l1000}. Both assays apply matched compound--dose conditions on parallel plates and are joined at the condition level.

Table~\ref{tab:paired-cohort-statistics} reports both the released profile rows and the matched samples used for modeling. BBBC036 begins with 21,122 CP and 6,929 L1000 replicate profiles and retains 1,916 matched compound--dose conditions; BBBC047 begins with 153,386 CP and 68,120 L1000 profiles and retains 20,081 matched conditions. The matched subsets span 47 CP/22 L1000 treatment plates for BBBC036 and 273 CP/360 L1000 treatment plates for BBBC047. Before aggregation, 143 BBBC047 L1000 treatment profiles without a same-plate control are removed.

For each condition, $c$ is the featurewise median of same-plate control-only CP profiles. Compound identity $p$ is represented by a 2,048-bit radius-2 Morgan fingerprint \citep{rogers2010ecfp}, while the separate attribute $a$ is the log-transformed observed dose. Before treatment filtering and condition aggregation, numeric profile columns are converted to float32 and any column containing a non-finite value in the released profile table is removed. CP treatment responses undergo the empirical same-plate-control mid-CDF transform, are centered by subtracting $0.5$, and are aggregated by condition median; L1000 responses are centered by the same-plate control median and likewise aggregated by condition median. The resulting joint targets contain 591 CP and 977 L1000 outputs for BBBC036, and 701 CP and 977 L1000 outputs for BBBC047, yielding 1,568 and 1,678 dimensions. For training, each target coordinate is centered and scaled using its Folds 1--2 mean and population standard deviation; scales at or below $10^{-8}$ are replaced by one. Reported metrics invert this standardization, retaining the control-transformed response scale. Bemis--Murcko groups \citep{bemis1996frameworks} are assigned without response values and remain disjoint across all five folds. Table~\ref{tab:paired-fold-statistics} lists the complete fold inventory and experimental roles.

For the complementary plate-family/context transfer, connected components are formed from released CP treatment-plate identifiers, their same-plate control contexts, linked L1000 treatment plates, and matched-condition counts, without reading responses. The construction yields six components for BBBC036 and 64 for BBBC047, assigning every CP/context and linked L1000 plate to exactly one role. Fit/selection/audit/replication condition counts are 640/320/637/319 for BBBC036 and 8,264/3,868/3,826/4,123 for BBBC047. BBBC036 therefore provides a six-family transfer case, while BBBC047 supplies broader 64-family coverage.

% Appendix presentation derived from ../generated_dataset_statistics.tex
\begin{table}[!htbp]
\centering
\caption{\textbf{Paired morphology--transcriptomic task cohorts after consistent preprocessing.} Source counts are replicate-level profile rows from the released tables, not image or cell counts; parentheses give treatment/control rows. Matched conditions are compound--dose samples retained after condition-level matching across assays.}
\label{tab:paired-cohort-statistics}
\tblsetup
\begin{lrbox}{\AppendixTableBox}
\begin{tabular}{@{}lrr@{}}
\toprule
\rowcolor{csHeader}
\tblhdr{Cohort} & \tblhdr{CP source profiles} & \tblhdr{L1000 source profiles} \\
\midrule
\multicolumn{3}{@{}l}{\textbf{A. Released source profiles}} \\
BBBC036 & \tblstack{21,122}{(17,594/3,528)} & \tblstack{6,929}{(3,451/3,478)} \\
BBBC047 & \tblstack{153,386}{(126,814/26,572)} & \tblstack{68,120}{(64,642/3,478)} \\
\bottomrule
\end{tabular}
\end{lrbox}
\AppendixTableOutput{generated_dataset_statistics-1}
\tblnote{Parentheses: treatment/control profile rows.}
\par\vspace{5pt}
\begin{lrbox}{\AppendixTableBox}
\begin{tabular}{@{}lrrrrrr@{}}
\toprule
\rowcolor{csHeader}
\tblhdr{Cohort} & \tblhdr{Matched\\conditions} & \tblhdr{Unique\\SMILES} & \tblhdr{Murcko\\groups} & \tblhdr{CP/context\\dim.} & \tblhdr{L1000\\dim.} & \tblhdr{Joint target\\dim.} \\
\midrule
\multicolumn{7}{@{}l}{\textbf{B. Matched samples and model dimensions}} \\
BBBC036 & 1,916 & 1,916 & 1,196 & 591 & 977 & 1,568 \\
BBBC047 & 20,081 & 20,068 & 4,540 & 701 & 977 & 1,678 \\
\bottomrule
\end{tabular}
\end{lrbox}
\AppendixTableOutput{generated_dataset_statistics-2}
\end{table}

\begin{table}[!htbp]
\centering
\caption{\textbf{Fold-level inventory for the paired task cohorts.} Folds 1--2 fit models; Fold 3 supplies search feedback and within-policy selection. Fold 4 is held out from discovery and supplies the disclosed across-policy source selection in the prediction-score panel. That panel independently refits the selected source for Fold 5; constrained searches evaluate the same frozen checkpoint on both later folds (Appendix~\ref{app:protocol}). Scaffold groups and assay replicates are counted after filtering and condition matching.}
\label{tab:paired-fold-statistics}
\tblsetup
\begin{lrbox}{\AppendixTableBox}
\begin{tabular}{@{}lrrrrr@{}}
\toprule
\rowcolor{csHeader}
\tblhdr{Statistic} & \tblhdr{Fold 1} & \tblhdr{Fold 2} & \tblhdr{Fold 3} & \tblhdr{Fold 4} & \tblhdr{Fold 5} \\
\midrule
\rowcolor{csHeader}\multicolumn{6}{@{}l}{\textbf{BBBC036}} \\
Matched conditions & 285 & 509 & 356 & 411 & 355 \\
Murcko groups & 206 & 244 & 258 & 265 & 223 \\
CP replicate profiles & 2,222 & 3,995 & 2,779 & 3,213 & 2,803 \\
L1000 replicate profiles & 519 & 919 & 626 & 739 & 648 \\
\rowcolor{csHeader}\multicolumn{6}{@{}l}{\textbf{BBBC047}} \\
Matched conditions & 3,788 & 3,921 & 4,114 & 4,380 & 3,878 \\
Murcko groups & 853 & 912 & 941 & 913 & 921 \\
CP replicate profiles & 15,891 & 17,060 & 17,263 & 18,390 & 16,429 \\
L1000 replicate profiles & 10,722 & 10,692 & 11,531 & 12,252 & 10,815 \\
\bottomrule
\end{tabular}
\end{lrbox}
\AppendixTableOutput{generated_dataset_statistics-3}
\end{table}

% Appendix presentation derived from ../generated_plate_context_generalization.tex
\begin{table}[!htbp]
\centering
\caption{Generalization of selected predictors to unseen plate families. The split is constructed without response values and models are selected before evaluation; each value is the mean [95\% CI] over five paired refits. The same-plate control profile remains available as an input, while all Cell Painting treatment/control plates and linked L1000 plates are disjoint across split roles.}
\label{tab:plate-context-generalization}
\tblsetup
\begin{lrbox}{\AppendixTableBox}
\begin{tabular}{@{}lccrrr@{}}
\toprule
\rowcolor{csHeader}
\tblhdr{Frozen source} & \tblhdr{Audit\\Global PCC} & \tblhdr{Replication\\Global PCC} & \tblhdr{Audit\\CP PCC} & \tblhdr{Audit\\L1000 PCC} & \tblhdr{Parameters} \\
\midrule
\rowcolor{csHeader}\multicolumn{6}{@{}l}{\textbf{BBBC036}} \\
$h_0$ & \tblstack{-0.0030}{[-0.0400, 0.0339]} & \tblstack{0.0484}{[0.0223, 0.0746]} & 0.0255 & -0.0130 & 1,144,864 \\
 CellScientist & \tblstack{\textbf{0.1921}}{[0.1903, 0.1939]} & \tblstack{\textbf{0.1567}}{[0.1517, 0.1617]} & 0.2758 & 0.1568 & 3,999,776 \\
AIDE & \tblstack{\secondbest{0.1746}}{[0.1495, 0.1998]} & \tblstack{\secondbest{0.1497}}{[0.1193, 0.1800]} & 0.2520 & 0.1400 & 4,419,872 \\
CellForge & \tblstack{-0.0091}{[-0.0397, 0.0215]} & \tblstack{0.0640}{[0.0187, 0.1093]} & 0.0821 & -0.0435 & 3,824,672 \\
HarmonyCell & \tblstack{0.0657}{[0.0378, 0.0936]} & \tblstack{0.0102}{[-0.0426, 0.0629]} & 0.0486 & 0.0719 & 2,061,343 \\
\addlinespace[4pt]
\rowcolor{csHeader}\multicolumn{6}{@{}l}{\textbf{BBBC047}} \\
$h_0$ & \tblstack{\secondbest{0.1805}}{[0.1769, 0.1840]} & \tblstack{\secondbest{0.2014}}{[0.1941, 0.2087]} & 0.2688 & 0.0494 & 1,201,294 \\
 CellScientist & \tblstack{0.1781}{[0.1716, 0.1846]} & \tblstack{\textbf{0.2036}}{[0.1940, 0.2132]} & 0.2615 & 0.0469 & 1,777,486 \\
AIDE & \tblstack{\secondbest{0.1805}}{[0.1769, 0.1840]} & \tblstack{\secondbest{0.2014}}{[0.1941, 0.2087]} & 0.2688 & 0.0494 & 1,201,294 \\
CellForge & \tblstack{0.1771}{[0.1716, 0.1826]} & \tblstack{0.2011}{[0.1963, 0.2059]} & 0.2652 & 0.0444 & 2,531,982 \\
HarmonyCell & \tblstack{\textbf{0.1808}}{[0.1699, 0.1916]} & \tblstack{0.1963}{[0.1921, 0.2006]} & 0.2689 & 0.0484 & 2,284,875 \\
\bottomrule
\end{tabular}
\end{lrbox}
\AppendixTableOutput{generated_plate_context_generalization-1}
\tblnote{Bold/underline: best/second distinct displayed predictive mean within each task or fold. Ties share a rank; interval bounds and diagnostic/resource columns are unranked.}
\end{table}

\paragraph{Plate-family $2\times2$ audit.}
The plate-family audit crosses the observed or permuted compound representation with the observed or an alternative control profile from the same plate family. Each selected model is refit under five fixed seeds, and each refit averages 32 permutations constructed without response values. Compounds are permuted within the observed control-profile group whenever possible, covering every held-out BBBC036 row and $98.8\%$ of BBBC047 rows; each remaining row receives a fixed donor with a different compound from the same held-out partition. Table~\ref{tab:plate-2x2} separates compound, control-profile, and interaction effects on Folds 4 and 5.

% Appendix presentation derived from ../generated_plate_2x2.tex
\begin{table}[!htbp]
\centering
\caption{Compound/control audit on unseen plate families. Values are means [95\% CI] across five paired training refits; each effect averages 32 fixed matched permutations within a refit. Positive compound and control effects denote a loss of Global PCC after the corresponding input intervention. Scientific notation preserves nonzero values below the four-decimal rounding threshold of $5\times10^{-5}$.}
\label{tab:plate-2x2}
\tblsetup
\begin{lrbox}{\AppendixTableBox}
\begin{tabular}{@{}lcccc@{}}
\toprule
\rowcolor{csHeader}
\tblhdr{Model} & \tblhdr{Global PCC} & \tblhdr{Compound effect} & \tblhdr{Control effect} & \tblhdr{Interaction} \\
\midrule
\multicolumn{5}{@{}l}{\textbf{BBBC036 / Audit}} \\
$h_0$ & \tblstack{-0.0030}{[-0.0400, 0.0339]} & \tblstack{0.0025}{[0.0020, 0.0031]} & \tblstack{0.0012}{[-0.0126, 0.0150]} & \tblstack{-0.0005}{[-0.0008, -0.0002]} \\
 CellScientist & \tblstack{0.1921}{[0.1903, 0.1939]} & \tblstack{0.0000}{[0.0000, 0.0000]} & \tblstack{0.0146}{[0.0114, 0.0178]} & \tblstack{0.0000}{[0.0000, 0.0000]} \\
AIDE & \tblstack{0.1746}{[0.1495, 0.1998]} & \tblstack{0.0024}{[0.0019, 0.0028]} & \tblstack{0.0260}{[0.0076, 0.0445]} & \tblstack{-0.0008}{[-0.0012, -0.0003]} \\
CellForge & \tblstack{-0.0091}{[-0.0397, 0.0215]} & \tblstack{0.0018}{[0.0011, 0.0025]} & \tblstack{-0.0086}{[-0.0169, -0.0004]} & \tblstack{-0.0005}{[-0.0010, 0.0001]} \\
HarmonyCell & \tblstack{0.0657}{[0.0378, 0.0936]} & \tblstack{0.0019}{[0.0002, 0.0037]} & \tblstack{0.0768}{[0.0632, 0.0905]} & \tblstack{-0.0028}{[-0.0044, -0.0013]} \\
\addlinespace[3pt]
\multicolumn{5}{@{}l}{\textbf{BBBC047 / Audit}} \\
$h_0$ & \tblstack{0.1805}{[0.1769, 0.1840]} & \tblstack{0.0046}{[0.0039, 0.0053]} & \tblstack{0.0249}{[0.0228, 0.0271]} & \tblstack{0.0008}{[0.0005, 0.0012]} \\
 CellScientist & \tblstack{0.1781}{[0.1716, 0.1846]} & \tblstack{0.0000}{[0.0000, 0.0000]} & \tblstack{0.0295}{[0.0243, 0.0346]} & \tblstack{0.0000}{[0.0000, 0.0000]} \\
AIDE & \tblstack{0.1805}{[0.1769, 0.1840]} & \tblstack{0.0046}{[0.0039, 0.0053]} & \tblstack{0.0249}{[0.0228, 0.0271]} & \tblstack{0.0008}{[0.0005, 0.0012]} \\
CellForge & \tblstack{0.1771}{[0.1716, 0.1826]} & \tblstack{0.0050}{[0.0039, 0.0061]} & \tblstack{0.0208}{[0.0163, 0.0252]} & \tblstack{0.0014}{[0.0004, 0.0023]} \\
HarmonyCell & \tblstack{0.1808}{[0.1699, 0.1916]} & \tblstack{$6.892\!\times\!10^{-9}$}{$[-1.390,2.768]\!\times\!10^{-8}$} & \tblstack{0.0365}{[0.0245, 0.0484]} & \tblstack{$1.211\!\times\!10^{-8}$}{$[-2.216,4.637]\!\times\!10^{-8}$} \\
\addlinespace[3pt]
\multicolumn{5}{@{}l}{\textbf{BBBC036 / Held-out replication}} \\
$h_0$ & \tblstack{0.0484}{[0.0223, 0.0746]} & \tblstack{0.0026}{$[9.873\!\times\!10^{-6},0.0053]$} & \tblstack{0.0043}{[-0.0117, 0.0203]} & \tblstack{0.0002}{[-0.0006, 0.0009]} \\
 CellScientist & \tblstack{0.1567}{[0.1517, 0.1617]} & \tblstack{0.0000}{[0.0000, 0.0000]} & \tblstack{-0.0070}{[-0.0088, -0.0053]} & \tblstack{0.0000}{[0.0000, 0.0000]} \\
AIDE & \tblstack{0.1497}{[0.1193, 0.1800]} & \tblstack{0.0013}{[-0.0004, 0.0031]} & \tblstack{-0.0016}{[-0.0168, 0.0136]} & \tblstack{-0.0001}{[-0.0004, 0.0002]} \\
CellForge & \tblstack{0.0640}{[0.0187, 0.1093]} & \tblstack{0.0034}{[0.0026, 0.0043]} & \tblstack{0.0018}{[-0.0087, 0.0123]} & \tblstack{0.0001}{[-0.0006, 0.0009]} \\
HarmonyCell & \tblstack{0.0102}{[-0.0426, 0.0629]} & \tblstack{-0.0063}{[-0.0100, -0.0025]} & \tblstack{-0.0038}{[-0.0350, 0.0274]} & \tblstack{-0.0037}{[-0.0080, 0.0005]} \\
\addlinespace[3pt]
\multicolumn{5}{@{}l}{\textbf{BBBC047 / Held-out replication}} \\
$h_0$ & \tblstack{0.2014}{[0.1941, 0.2087]} & \tblstack{0.0030}{[0.0023, 0.0038]} & \tblstack{0.0136}{[0.0077, 0.0194]} & \tblstack{0.0002}{[-0.0010, 0.0013]} \\
 CellScientist & \tblstack{0.2036}{[0.1940, 0.2132]} & \tblstack{0.0000}{[0.0000, 0.0000]} & \tblstack{0.0231}{[0.0168, 0.0293]} & \tblstack{0.0000}{[0.0000, 0.0000]} \\
AIDE & \tblstack{0.2014}{[0.1941, 0.2087]} & \tblstack{0.0030}{[0.0023, 0.0038]} & \tblstack{0.0136}{[0.0077, 0.0194]} & \tblstack{0.0002}{[-0.0010, 0.0013]} \\
CellForge & \tblstack{0.2011}{[0.1963, 0.2059]} & \tblstack{0.0037}{[0.0030, 0.0044]} & \tblstack{0.0088}{[0.0029, 0.0148]} & \tblstack{-0.0004}{[-0.0010, 0.0002]} \\
HarmonyCell & \tblstack{0.1963}{[0.1921, 0.2006]} & \tblstack{$3.073\!\times\!10^{-9}$}{$[-5.460,11.61]\!\times\!10^{-9}$} & \tblstack{0.0205}{[0.0142, 0.0268]} & \tblstack{$2.608\!\times\!10^{-9}$}{$[-4.632,9.848]\!\times\!10^{-9}$} \\
\bottomrule
\end{tabular}
\end{lrbox}
\AppendixTableOutput{generated_plate_2x2-1}
\end{table}

\paragraph{Joint residual references.}
Fixed residual baselines isolate a perturbation-specific function class while keeping the same task. They retain the selected control-profile predictor $g(c)$ and learn one joint residual $h(p,a)$ using either multi-output Ridge or a shallow multi-output MLP. Both baselines use Fold~3 for selection and are refit under the same five seeds. Their Fold-4 and Fold-5 predictive scores, compound effects, and target-loss gains are reported together in Table~\ref{tab:d3-residual-frontier}.

Two additional small-molecule cohorts test the final procedure beyond BBBC development. We first apply it to paired morphology--transcription LINCS--Pilot1, then transfer its ten trajectory-selected design instances to independently acquired Cell Painting LKCP Batch 2 under the same compound, dose, and control-profile inputs. Appendix~\ref{app:external-transfer} records data provenance, the order in which models and rules were fixed, the input permutations, and the compound--dose decomposition.

\FloatBarrier
\subsection{Physical separation of control inputs and target references}
\label{app:control-decoupling}
% Appendix presentation derived from ../generated_bbbc_decoupling_main.tex
% Generated by tools/render_validation_evidence.py; do not edit by hand.
\begin{table}[!htbp]
\centering
\caption{BBBC047 with shared control wells or disjoint control wells. Each arm uses newly constructed data and the same fixed model design.}
\label{tab:bbbc-control-decoupling}
\tblsetup
\begin{lrbox}{\AppendixTableBox}
\begin{tabular}{@{}L{43pt}C{17pt}R{48pt}R{48pt}R{48pt}R{48pt}R{48pt}R{48pt}@{}}
\toprule
\rowcolor{csHeader}
Model & Fold & \tblhdr{Shared\\PCC} & \tblhdr{Disjoint\\PCC} & \tblhdr{Shared\\PCC drop} & \tblhdr{Disjoint\\PCC drop} & \tblhdr{Shared\\Loss gain} & \tblhdr{Disjoint\\Loss gain} \\
\midrule
Score & F4 & $0.3081$ & $0.3085$ & $0$ & $0$ & $0$ & $0$ \\
Control only & F4 & $0.3084$ & $0.3082$ & $0$ & $0$ & $0$ & $0$ \\

Guided & F4 & $0.3111$ & $0.3110$ & $0.0047$ & $0.0043$ & $1.65\!\times\!10^{-4}$ & $1.51\!\times\!10^{-4}$ \\
\midrule
Score & F5 & $0.3191$ & $0.3193$ & $0$ & $0$ & $0$ & $0$ \\
Control only & F5 & $0.3187$ & $0.3185$ & $0$ & $0$ & $0$ & $0$ \\

Guided & F5 & $0.3214$ & $0.3210$ & $0.0036$ & $0.0033$ & $1.31\!\times\!10^{-4}$ & $1.19\!\times\!10^{-4}$ \\
\bottomrule
\end{tabular}
\end{lrbox}
\AppendixTableOutput{generated_bbbc_decoupling_main-1}
\tblnote{Means of five seeds, averaging directions A/B within seed; compound replacement uses 32 fixed maps. PCC drop and loss gain are correct-minus-replacement PCC and replacement-minus-correct MSE. Guided minus control-only disjoint-arm PCC: F4 $0.0028$ $[0.0017,\,0.0038]$, F5 $0.0025$ $[0.0012,\,0.0037]$ (paired-seed 95\% $t$ CIs).}
\end{table}

% Appendix presentation derived from ../generated_bbbc_decoupling_appendix.tex
% Generated by tools/render_validation_evidence.py; do not edit by hand.
\paragraph{Construction and scope.}
The shared and disjoint arms are newly constructed BBBC047 comparisons; the historical shared-control result is not their absolute baseline. Physical control wells are hash-sorted within each plate and well-row stratum and allocated to balanced, disjoint A/B banks (seed 2026091401). Within direction A, the target uses bank A as reference: shared context uses A, disjoint context uses B; direction B reverses these roles. CP targets are treated-well empirical mid-CDF values relative to the direction's reference bank, centered by $0.5$ and aggregated by condition medians. The 701 CP features are fixed from the historical schema; no new feature selection is performed. Shared/disjoint arms have byte-identical targets and row identities within direction, and L1000 targets are unchanged. A/B are averaged within each seed, not counted as independent replicas. This intervention separates physical wells at the reconstructed CP reference layer; it does not establish that all upstream preprocessing or biological dependence has been removed. No historical qualification label is changed.

\begin{table}[!htbp]
\centering
\caption{Construction and inference units for the new control-well comparison.}
\label{tab:bbbc-decoupling-construction}
\tblsetup
\begin{lrbox}{\AppendixTableBox}
\begin{tabular}{@{}L{121pt}L{88pt}L{166pt}@{}}
\toprule
\rowcolor{csHeader}
Quantity & Value & Interpretation \\
\midrule
Predictor refits / anchors & 60 / 20 & 80 training stages; 120 fold evaluations \\
Seeds / directions & 5 / 2 & A/B are paired inside each seed \\
F4 retained support & 3341 / 4380 & 76.28\% of evaluation rows \\
F5 retained support & 2927 / 3878 & 75.48\% of evaluation rows \\
F4 / F5 source scaffolds & 913 / 921 & Bootstrap over source-scaffold labels \\
Maps / scaffold draws & 32 / 2000 & All refits and donor maps held fixed \\
Donor kernel total variation & 0 (both folds) & Every eligible compound contributes one row \\
Guided coordinate signs & 40 / 40 positive & RMS, loss gain and PCC drop, each record \\
Score/control coordinates & 80 / 80 exact zero & All three compound coordinates \\
Physical controls, A / B & 9144 / 9144 & Unique plate/well identities; banks disjoint \\
CP plates / retained conditions & 273 / 20081 & 0 conditions excluded \\
\bottomrule
\end{tabular}
\end{lrbox}
\AppendixTableOutput{generated_bbbc_decoupling_appendix-1}
\tblnote{Only row-uniform maps were evaluated. An independent metadata calculation proves equality of row-uniform and compound-balanced donor probabilities on all 6268 retained source pools; pre-generated compound-balanced maps are not an additional evaluated sensitivity analysis. Kernel equality is not biological exchangeability.}
\end{table}

\paragraph{Target transformation check.} Separately from the direction-specific bank targets, a full-bank CP mid-CDF reconstruction was compared with historical normalized CP ranks. It recorded a maximum absolute discrepancy of $0.19140625$ and mean absolute discrepancy $2.76\!\times\!10^{-6}$, with $99.9799\%$ agreement at $10^{-6}$. This discrepancy was retained, not repaired by choosing a favorable target transform. The portable metadata retains the complete check.

\begin{table}[!htbp]
\centering
\caption{Complete predictive means for shared and disjoint control banks at both held-out folds (five seeds per row).}
\label{tab:bbbc-decoupling-predictive-f4}
\label{tab:bbbc-decoupling-predictive-f5}
\tblsetup
\begin{lrbox}{\AppendixTableBox}
\begin{tabular}{@{}L{43pt}C{16pt}C{12pt}L{33pt}R{38pt}R{38pt}R{38pt}R{38pt}R{38pt}R{38pt}@{}}
\toprule
\rowcolor{csHeader}
Model & Fold & Dir. & Arm & \tblhdr{Global\\PCC} & \tblhdr{MSE} & \tblhdr{CP\\PCC} & \tblhdr{CP\\MSE} & \tblhdr{L1000\\PCC} & \tblhdr{L1000\\MSE} \\
\midrule
\rowcolor{csHeader}\multicolumn{10}{@{}l}{\textbf{Fold 4}} \\
Score & F4 & A & Shared & $0.3082$ & $0.0460$ & $0.3331$ & $0.0532$ & $0.2802$ & $0.0408$ \\
Score & F4 & A & Disjoint & $0.3087$ & $0.0460$ & $0.3334$ & $0.0532$ & $0.2807$ & $0.0408$ \\
Score & F4 & B & Shared & $0.3079$ & $0.0460$ & $0.3317$ & $0.0532$ & $0.2810$ & $0.0408$ \\
Score & F4 & B & Disjoint & $0.3084$ & $0.0460$ & $0.3317$ & $0.0532$ & $0.2818$ & $0.0408$ \\
Control only & F4 & A & Shared & $0.3093$ & $0.0460$ & $0.3327$ & $0.0532$ & $0.2828$ & $0.0408$ \\
Control only & F4 & A & Disjoint & $0.3086$ & $0.0460$ & $0.3322$ & $0.0533$ & $0.2818$ & $0.0408$ \\
Control only & F4 & B & Shared & $0.3074$ & $0.0460$ & $0.3303$ & $0.0533$ & $0.2813$ & $0.0408$ \\
Control only & F4 & B & Disjoint & $0.3077$ & $0.0460$ & $0.3304$ & $0.0533$ & $0.2818$ & $0.0408$ \\
Guided & F4 & A & Shared & $0.3122$ & $0.0459$ & $0.3384$ & $0.0530$ & $0.2823$ & $0.0408$ \\
Guided & F4 & A & Disjoint & $0.3115$ & $0.0459$ & $0.3373$ & $0.0531$ & $0.2821$ & $0.0408$ \\
Guided & F4 & B & Shared & $0.3100$ & $0.0459$ & $0.3350$ & $0.0531$ & $0.2814$ & $0.0408$ \\
Guided & F4 & B & Disjoint & $0.3105$ & $0.0459$ & $0.3356$ & $0.0531$ & $0.2817$ & $0.0408$ \\
\addlinespace[3pt]
\rowcolor{csHeader}\multicolumn{10}{@{}l}{\textbf{Fold 5}} \\
Score & F5 & A & Shared & $0.3219$ & $0.0454$ & $0.3492$ & $0.0536$ & $0.2895$ & $0.0396$ \\
Score & F5 & A & Disjoint & $0.3222$ & $0.0454$ & $0.3500$ & $0.0535$ & $0.2889$ & $0.0396$ \\
Score & F5 & B & Shared & $0.3163$ & $0.0455$ & $0.3389$ & $0.0537$ & $0.2894$ & $0.0396$ \\
Score & F5 & B & Disjoint & $0.3165$ & $0.0455$ & $0.3386$ & $0.0537$ & $0.2903$ & $0.0396$ \\
Control only & F5 & A & Shared & $0.3217$ & $0.0454$ & $0.3480$ & $0.0536$ & $0.2904$ & $0.0396$ \\
Control only & F5 & A & Disjoint & $0.3218$ & $0.0455$ & $0.3493$ & $0.0536$ & $0.2889$ & $0.0396$ \\
Control only & F5 & B & Shared & $0.3156$ & $0.0455$ & $0.3373$ & $0.0538$ & $0.2897$ & $0.0396$ \\
Control only & F5 & B & Disjoint & $0.3153$ & $0.0456$ & $0.3364$ & $0.0539$ & $0.2901$ & $0.0396$ \\
Guided & F5 & A & Shared & $0.3249$ & $0.0454$ & $0.3536$ & $0.0534$ & $0.2905$ & $0.0396$ \\
Guided & F5 & A & Disjoint & $0.3236$ & $0.0454$ & $0.3525$ & $0.0535$ & $0.2892$ & $0.0396$ \\
Guided & F5 & B & Shared & $0.3178$ & $0.0455$ & $0.3416$ & $0.0537$ & $0.2895$ & $0.0396$ \\
Guided & F5 & B & Disjoint & $0.3184$ & $0.0455$ & $0.3420$ & $0.0536$ & $0.2903$ & $0.0396$ \\
\bottomrule
\end{tabular}
\end{lrbox}
\AppendixTableOutput{generated_bbbc_decoupling_appendix-2}
\end{table}

\begin{table}[!htbp]
\centering
\caption{Complete compound-coordinate means for shared and disjoint control banks at both held-out folds (five seeds per row).}
\label{tab:bbbc-decoupling-coordinate-f4}
\label{tab:bbbc-decoupling-coordinate-f5}
\tblsetup
\begin{lrbox}{\AppendixTableBox}
\begin{tabular}{@{}L{43pt}C{16pt}C{12pt}L{33pt}R{79pt}R{79pt}R{79pt}@{}}
\toprule
\rowcolor{csHeader}
Model & Fold & Dir. & Arm & \tblhdr{Compound\\RMS} & \tblhdr{Compound\\loss gain} & \tblhdr{Compound\\PCC drop} \\
\midrule
\rowcolor{csHeader}\multicolumn{7}{@{}l}{\textbf{Fold 4}} \\
Score & F4 & A & Shared & $0$ & $0$ & $0$ \\
Score & F4 & A & Disjoint & $0$ & $0$ & $0$ \\
Score & F4 & B & Shared & $0$ & $0$ & $0$ \\
Score & F4 & B & Disjoint & $0$ & $0$ & $0$ \\
Control only & F4 & A & Shared & $0$ & $0$ & $0$ \\
Control only & F4 & A & Disjoint & $0$ & $0$ & $0$ \\
Control only & F4 & B & Shared & $0$ & $0$ & $0$ \\
Control only & F4 & B & Disjoint & $0$ & $0$ & $0$ \\
Guided & F4 & A & Shared & $0.0126$ & $1.75\!\times\!10^{-4}$ & $0.0050$ \\
Guided & F4 & A & Disjoint & $0.0118$ & $1.48\!\times\!10^{-4}$ & $0.0041$ \\
Guided & F4 & B & Shared & $0.0118$ & $1.55\!\times\!10^{-4}$ & $0.0045$ \\
Guided & F4 & B & Disjoint & $0.0118$ & $1.54\!\times\!10^{-4}$ & $0.0044$ \\
\addlinespace[3pt]
\rowcolor{csHeader}\multicolumn{7}{@{}l}{\textbf{Fold 5}} \\
Score & F5 & A & Shared & $0$ & $0$ & $0$ \\
Score & F5 & A & Disjoint & $0$ & $0$ & $0$ \\
Score & F5 & B & Shared & $0$ & $0$ & $0$ \\
Score & F5 & B & Disjoint & $0$ & $0$ & $0$ \\
Control only & F5 & A & Shared & $0$ & $0$ & $0$ \\
Control only & F5 & A & Disjoint & $0$ & $0$ & $0$ \\
Control only & F5 & B & Shared & $0$ & $0$ & $0$ \\
Control only & F5 & B & Disjoint & $0$ & $0$ & $0$ \\
Guided & F5 & A & Shared & $0.0125$ & $1.43\!\times\!10^{-4}$ & $0.0039$ \\
Guided & F5 & A & Disjoint & $0.0116$ & $1.17\!\times\!10^{-4}$ & $0.0032$ \\
Guided & F5 & B & Shared & $0.0117$ & $1.19\!\times\!10^{-4}$ & $0.0033$ \\
Guided & F5 & B & Disjoint & $0.0117$ & $1.21\!\times\!10^{-4}$ & $0.0034$ \\
\bottomrule
\end{tabular}
\end{lrbox}
\AppendixTableOutput{generated_bbbc_decoupling_appendix-3}
\end{table}

\begin{table}[!htbp]
\centering
\caption{Predictive contrasts under control separation. Seed and source-scaffold uncertainty are reported separately; the two metric blocks retain their respective scale factors.}
\label{tab:bbbc-decoupling-paired}
\label{tab:bbbc-decoupling-paired-mse}
\tblsetup
\begin{lrbox}{\AppendixTableBox}
\begin{tabular}{@{}L{96pt}C{17pt}R{40pt}R{100pt}R{100pt}@{}}
\toprule
\rowcolor{csHeader}
Contrast & Fold & Mean & Seed 95\% CI & Scaffold 95\% CI \\
\midrule
\rowcolor{csHeader}\multicolumn{5}{@{}l}{\textbf{Global PCC ($\times 10^{3}$)}} \\
Score: D$-$S & F4 & $0.427$ & $[-0.130,\,0.984]$ & $[-0.070,\,0.945]$ \\
Score: D$-$S & F5 & $0.243$ & $[-0.583,\,1.069]$ & $[-0.158,\,0.638]$ \\
Control only: D$-$S & F4 & $-0.186$ & $[-0.769,\,0.398]$ & $[-0.794,\,0.338]$ \\
Control only: D$-$S & F5 & $-0.136$ & $[-1.679,\,1.407]$ & $[-0.687,\,0.393]$ \\
Guided: D$-$S & F4 & $-0.099$ & $[-0.972,\,0.773]$ & $[-0.442,\,0.225]$ \\
Guided: D$-$S & F5 & $-0.327$ & $[-0.963,\,0.309]$ & $[-0.640,\,0.004]$ \\
Score$-$control, S & F4 & $-0.295$ & $[-1.223,\,0.633]$ & $[-1.724,\,1.066]$ \\
Score$-$control, S & F5 & $0.428$ & $[-0.662,\,1.517]$ & $[-0.938,\,1.539]$ \\
Guided$-$control, S & F4 & $2.688$ & $[2.059,\,3.317]$ & $[1.012,\,4.541]$ \\
Guided$-$control, S & F5 & $2.674$ & $[1.609,\,3.739]$ & $[0.272,\,5.761]$ \\
Score$-$control, D & F4 & $0.318$ & $[-0.514,\,1.149]$ & $[-0.853,\,1.491]$ \\
Score$-$control, D & F5 & $0.807$ & $[-0.037,\,1.650]$ & $[-0.554,\,1.931]$ \\
Guided$-$control, D & F4 & $2.775$ & $[1.731,\,3.818]$ & $[1.264,\,4.468]$ \\
Guided$-$control, D & F5 & $2.483$ & $[1.227,\,3.739]$ & $[0.187,\,5.548]$ \\
\addlinespace[3pt]
\rowcolor{csHeader}\multicolumn{5}{@{}l}{\textbf{MSE ($\times 10^{5}$)}} \\
Score: D$-$S & F4 & $-1.079$ & $[-2.884,\,0.726]$ & $[-2.847,\,0.536]$ \\
Score: D$-$S & F5 & $-0.564$ & $[-3.258,\,2.131]$ & $[-1.846,\,0.775]$ \\
Control only: D$-$S & F4 & $1.923$ & $[-1.011,\,4.858]$ & $[0.232,\,3.870]$ \\
Control only: D$-$S & F5 & $1.490$ & $[-4.983,\,7.964]$ & $[-0.383,\,3.538]$ \\
Guided: D$-$S & F4 & $1.047$ & $[-2.859,\,4.954]$ & $[-0.192,\,2.309]$ \\
Guided: D$-$S & F5 & $1.780$ & $[-0.662,\,4.222]$ & $[0.555,\,2.980]$ \\
Score$-$control, S & F4 & $-0.598$ & $[-3.662,\,2.465]$ & $[-4.858,\,4.209]$ \\
Score$-$control, S & F5 & $-2.943$ & $[-7.456,\,1.569]$ & $[-6.772,\,1.784]$ \\
Guided$-$control, S & F4 & $-7.236$ & $[-11.010,\,-3.463]$ & $[-13.811,\,-1.332]$ \\
Guided$-$control, S & F5 & $-7.329$ & $[-14.377,\,-0.282]$ & $[-18.153,\,0.990]$ \\
Score$-$control, D & F4 & $-3.601$ & $[-6.091,\,-1.110]$ & $[-7.476,\,0.520]$ \\
Score$-$control, D & F5 & $-4.998$ & $[-8.488,\,-1.507]$ & $[-8.982,\,-0.105]$ \\
Guided$-$control, D & F4 & $-8.112$ & $[-12.543,\,-3.681]$ & $[-14.199,\,-2.832]$ \\
Guided$-$control, D & F5 & $-7.040$ & $[-12.443,\,-1.636]$ & $[-17.585,\,0.849]$ \\
\bottomrule
\end{tabular}
\end{lrbox}
\AppendixTableOutput{generated_bbbc_decoupling_appendix-4}

\tblnote{D$-$S: disjoint minus shared wells. Within-arm contrasts subtract control only. Seed CIs use five A/B-averaged paired values ($t_4$). Scaffold CIs share each draw across all arms/models/directions and fix all five refits, 32 maps and the donor pool; 2000 draws, seed $2026091433+\mathrm{fold}$. PCC is recomputed after pooling sufficient statistics, never averaged across scaffold PCCs. These are source-scaffold, not compound-cluster, intervals; scaffold labels do not ensure fully independent biological groups.}
\end{table}

\begin{table}[!htbp]
\centering
\caption{Compound sensitivity and target-loss contrasts under control separation. Each metric block retains its own scale; seed and source-scaffold intervals quantify distinct uncertainty.}
\label{tab:bbbc-decoupling-paired-compound-prediction-rms}
\label{tab:bbbc-decoupling-paired-compound-target-loss-gain}
\tblsetup
\begin{lrbox}{\AppendixTableBox}
\begin{tabular}{@{}L{96pt}C{17pt}R{40pt}R{100pt}R{100pt}@{}}
\toprule
\rowcolor{csHeader}
Contrast & Fold & Mean & Seed 95\% CI & Scaffold 95\% CI \\
\midrule
\rowcolor{csHeader}\multicolumn{5}{@{}l}{\textbf{Compound RMS ($\times 10^{3}$)}} \\
Score: D$-$S & F4 & $0$ & $[0,\,0]$ & $[0,\,0]$ \\
Score: D$-$S & F5 & $0$ & $[0,\,0]$ & $[0,\,0]$ \\
Control only: D$-$S & F4 & $0$ & $[0,\,0]$ & $[0,\,0]$ \\
Control only: D$-$S & F5 & $0$ & $[0,\,0]$ & $[0,\,0]$ \\
Guided: D$-$S & F4 & $-0.414$ & $[-1.567,\,0.738]$ & $[-0.495,\,-0.333]$ \\
Guided: D$-$S & F5 & $-0.406$ & $[-1.563,\,0.750]$ & $[-0.490,\,-0.327]$ \\
Score$-$control, S & F4 & $0$ & $[0,\,0]$ & $[0,\,0]$ \\
Score$-$control, S & F5 & $0$ & $[0,\,0]$ & $[0,\,0]$ \\
Guided$-$control, S & F4 & $12.220$ & $[11.209,\,13.231]$ & $[11.249,\,13.186]$ \\
Guided$-$control, S & F5 & $12.085$ & $[11.011,\,13.158]$ & $[11.456,\,12.679]$ \\
Score$-$control, D & F4 & $0$ & $[0,\,0]$ & $[0,\,0]$ \\
Score$-$control, D & F5 & $0$ & $[0,\,0]$ & $[0,\,0]$ \\
Guided$-$control, D & F4 & $11.805$ & $[9.826,\,13.784]$ & $[10.884,\,12.725]$ \\
Guided$-$control, D & F5 & $11.678$ & $[9.740,\,13.616]$ & $[11.106,\,12.221]$ \\
\addlinespace[3pt]
\rowcolor{csHeader}\multicolumn{5}{@{}l}{\textbf{Compound loss gain ($\times 10^{5}$)}} \\
Score: D$-$S & F4 & $0$ & $[0,\,0]$ & $[-2.36\!\times\!10^{-12},\,2.64\!\times\!10^{-12}]$ \\
Score: D$-$S & F5 & $0$ & $[0,\,0]$ & $[-2.71\!\times\!10^{-12},\,1.87\!\times\!10^{-12}]$ \\
Control only: D$-$S & F4 & $0$ & $[0,\,0]$ & $[-2.91\!\times\!10^{-12},\,1.53\!\times\!10^{-12}]$ \\
Control only: D$-$S & F5 & $0$ & $[0,\,0]$ & $[-1.46\!\times\!10^{-12},\,3.05\!\times\!10^{-12}]$ \\
Guided: D$-$S & F4 & $-1.403$ & $[-4.179,\,1.374]$ & $[-2.134,\,-0.809]$ \\
Guided: D$-$S & F5 & $-1.163$ & $[-3.432,\,1.107]$ & $[-1.938,\,-0.376]$ \\
Score$-$control, S & F4 & $0$ & $[0,\,0]$ & $[-2.43\!\times\!10^{-12},\,2.50\!\times\!10^{-12}]$ \\
Score$-$control, S & F5 & $0$ & $[0,\,0]$ & $[-1.39\!\times\!10^{-12},\,3.26\!\times\!10^{-12}]$ \\
Guided$-$control, S & F4 & $16.509$ & $[13.377,\,19.640]$ & $[10.136,\,24.392]$ \\
Guided$-$control, S & F5 & $13.102$ & $[10.729,\,15.475]$ & $[5.985,\,20.741]$ \\
Score$-$control, D & F4 & $0$ & $[0,\,0]$ & $[-1.32\!\times\!10^{-12},\,3.12\!\times\!10^{-12}]$ \\
Score$-$control, D & F5 & $0$ & $[0,\,0]$ & $[-2.57\!\times\!10^{-12},\,2.01\!\times\!10^{-12}]$ \\
Guided$-$control, D & F4 & $15.106$ & $[9.906,\,20.305]$ & $[9.267,\,22.336]$ \\
Guided$-$control, D & F5 & $11.939$ & $[7.423,\,16.455]$ & $[5.390,\,18.963]$ \\
\bottomrule
\end{tabular}
\end{lrbox}
\AppendixTableOutput{generated_bbbc_decoupling_appendix-5}

\tblnote{D$-$S: disjoint minus shared wells. Within-arm contrasts subtract control only. Seed CIs use five A/B-averaged paired values ($t_4$). Scaffold CIs share each draw across all arms/models/directions and fix all five refits, 32 maps and the donor pool; 2000 draws, seed $2026091433+\mathrm{fold}$. PCC is recomputed after pooling sufficient statistics, never averaged across scaffold PCCs. These are source-scaffold, not compound-cluster, intervals; scaffold labels do not ensure fully independent biological groups.}
\end{table}

\begin{table}[!htbp]
\centering
\caption{Compound PCC drop contrasts ($\times 10^{3}$): seed and source-scaffold uncertainty are distinct.}
\label{tab:bbbc-decoupling-paired-compound-pcc-drop}
\tblsetup
\begin{lrbox}{\AppendixTableBox}
\begin{tabular}{@{}L{96pt}C{17pt}R{40pt}R{100pt}R{100pt}@{}}
\toprule
\rowcolor{csHeader}
Contrast & Fold & Mean & Seed 95\% CI & Scaffold 95\% CI \\
\midrule
Score: D$-$S & F4 & $0$ & $[0,\,0]$ & $[-1.50\!\times\!10^{-13},\,1.83\!\times\!10^{-13}]$ \\
Score: D$-$S & F5 & $0$ & $[0,\,0]$ & $[-1.89\!\times\!10^{-13},\,1.78\!\times\!10^{-13}]$ \\
Control only: D$-$S & F4 & $0$ & $[0,\,0]$ & $[-1.61\!\times\!10^{-13},\,1.83\!\times\!10^{-13}]$ \\
Control only: D$-$S & F5 & $0$ & $[0,\,0]$ & $[-2.33\!\times\!10^{-13},\,1.44\!\times\!10^{-13}]$ \\
Guided: D$-$S & F4 & $-0.424$ & $[-1.191,\,0.342]$ & $[-0.627,\,-0.248]$ \\
Guided: D$-$S & F5 & $-0.337$ & $[-0.944,\,0.270]$ & $[-0.559,\,-0.115]$ \\
Score$-$control, S & F4 & $0$ & $[0,\,0]$ & $[-2.05\!\times\!10^{-13},\,1.11\!\times\!10^{-13}]$ \\
Score$-$control, S & F5 & $0$ & $[0,\,0]$ & $[-2.28\!\times\!10^{-13},\,1.33\!\times\!10^{-13}]$ \\
Guided$-$control, S & F4 & $4.708$ & $[3.832,\,5.583]$ & $[2.949,\,6.840]$ \\
Guided$-$control, S & F5 & $3.625$ & $[2.955,\,4.296]$ & $[1.650,\,5.741]$ \\
Score$-$control, D & F4 & $0$ & $[0,\,0]$ & $[-2.22\!\times\!10^{-13},\,1.33\!\times\!10^{-13}]$ \\
Score$-$control, D & F5 & $0$ & $[0,\,0]$ & $[-1.89\!\times\!10^{-13},\,1.67\!\times\!10^{-13}]$ \\
Guided$-$control, D & F4 & $4.283$ & $[2.823,\,5.744]$ & $[2.673,\,6.238]$ \\
Guided$-$control, D & F5 & $3.288$ & $[2.042,\,4.535]$ & $[1.471,\,5.215]$ \\
\bottomrule
\end{tabular}
\end{lrbox}
\AppendixTableOutput{generated_bbbc_decoupling_appendix-6}
\tblnote{D$-$S: disjoint minus shared wells. Within-arm contrasts subtract control only. Seed CIs use five A/B-averaged paired values ($t_4$). Scaffold CIs share each draw across all arms/models/directions and fix all five refits, 32 maps and the donor pool; 2000 draws, seed $2026091433+\mathrm{fold}$. PCC is recomputed after pooling sufficient statistics, never averaged across scaffold PCCs. These are source-scaffold, not compound-cluster, intervals; scaffold labels do not ensure fully independent biological groups.}
\end{table}

\paragraph{Reading the uncertainty.}
The guided model has positive compound RMS, target-loss gain and PCC drop in all 40 evaluations, whereas Score and control only are exactly invariant. The guided disjoint-arm PCC advantage over control only remains positive under both reported interval methods. However, all guided disjoint-minus-shared coordinate differences include zero under the five-seed interval while their fixed-refit source-scaffold intervals are negative; the two uncertainty scopes do not establish an unchanged effect. The F5 guided-minus-control MSE contrast likewise crosses zero under the source-scaffold interval despite a negative seed interval. A positive prediction RMS indicates input use, not target benefit. RMS is the mean over maps of the square root of mean squared prediction distance, not the square root after averaging maps.

\FloatBarrier
\section{Exact protocol, decision rules, and statistical units}
\label{app:protocol}

The fixed executor owns joint inputs $(c,p,a)$, the response target, data transformations, folds, evaluator, checkpoint rule, and training configuration; candidate code cannot alter data loading, targets, folds, or metrics. Discovery accesses Folds 1--3 only. Global \pcc is one Pearson correlation after flattening all post-search rows and all dimensions of the inverse-standardized joint output. CP and L1000 \pcc use the same operation within each response block. MSE is averaged over every element of the inverse-standardized joint output. At search step $t$, the policy maps the current model, Fold-3 diagnostics, and compact search history to a proposed revision,
\begin{equation}
s_{t+1}=\mathcal{D}(s_t,r_t,\mathcal{H}_t),
\end{equation}
where edits are limited to architecture, fusion, response readout, loss, optimizer, and scheduler. The experiment log stores the proposal's parent, hypothesis, source or structured design, declared input use, execution and repair results, checkpoint, metrics, provider usage, and compute. Within each ten-slot prediction-score trajectory, the incumbent changes only under a strict improvement in Fold-3 Global \pcc. After all ten trajectories finish, the first, within-policy selection chooses one source per search policy from that policy's complete executable candidate pool by maximum Fold-3 Global \pcc, with lower Fold-3 MSE, fewer parameters, and stable candidate identity as tie-breakers. Neither Fold 4 nor Fold 5 enters this within-policy selection. The four policy-selected agent sources and three distinct diagnostic sources are then refit under five paired seeds and evaluated on Fold 4; the $h_0$ source is also displayed under its simple-concatenation role. A second, pre-specified across-policy score rule chooses the agent source with the highest mean Fold-4 \pcc for transfer to Fold 5. Within source selection, Fold 4 enters only this disclosed across-policy choice, while Fold 5 never enters. The path-constrained and falsification-guided studies instead carry all ten Fold-3 winners through Folds 4 and 5 without Fold-4 reselection. Before Fold 4, each study freezes the candidate set and registers the source-check rules, input-intervention construction, map seeds, threshold-construction rule, status rule, and aggregation. The realized model-specific thresholds are computed from the 32 reference scores on the fold being audited. For the prediction-score panel, Fold 5 independently refits the frozen source on Folds 1--2, selects its checkpoint on Fold 3 under the same paired-seed schedule, and evaluates Fold 5 once. Thus Fold 5 does not reuse Fold-4 weights and never changes the selected source. LINCS and LKCP subsequently test the final procedure outside BBBC development.

The compound intervention first seeks a different-fingerprint derangement within rows sharing the same observed control-profile vector. If that class derangement is impossible, a deterministic partition-wide donor with a different fingerprint is used and recorded. Only the compound representation changes: the source control profile and dose remain fixed. The resulting effect is measured under this control-stratified, dose-preserving replacement distribution. The control intervention maps each row to a different observed control-vector class; donor reuse is allowed to ensure that every control vector changes. Dose maps replace only the scalar dose, with observed-dose support checked against the source compound's released BRD identity and available cell/time annotations. None of these maps uses responses, model predictions, or evaluation scores to choose donors.

For the primary scaffold-fold compound maps, within-control-group derangements cover $411/411$ and $355/355$ rows on BBBC036, $4330/4380$ and $3831/3878$ on BBBC047, $847/847$ and $868/868$ on LINCS, and $378/405$ and $378/432$ on LKCP, respectively on Folds 4 and 5. The remaining rows use partition-wide donors. These counts specify control matching; the response contrasts retain the source dose, including when the donor compound was not observed at that dose. The plate-family analysis uses its separately specified maps in Appendix~\ref{app:data}.

For the $2\times2$ audit, the model-specific reference values are derived from the intervened predictions rather than from responses used to choose a permutation. For compound use they are $b^{(\mathrm{chem})}_{j,r}=\tfrac12(m_{\tilde p c,r}+m_{\tilde p\tilde c,r})$; for control-profile use, $b^{(\mathrm{context})}_{j,r}=\tfrac12(m_{p\tilde c,r}+m_{\tilde p\tilde c,r})$; and for their interaction, $b^{(\mathrm{int})}_{j,r}=m_{\tilde p c,r}-m_{\tilde p\tilde c,r}$. Dose uses the prediction score under a within-compound dose permutation. In every case, $\tau_{j,q}$ is the 95th percentile of absolute pairwise differences among the 32 fixed reference values. The lower and upper bounds used for the status are the empirical 2.5th and 97.5th percentiles of the 32 effect values; the interaction uses absolute effects. A model is qualified when the lower bound exceeds $\tau_{j,q}$, unsupported when the upper bound does not exceed $\tau_{j,q}$, and inconclusive otherwise. Inputs without enough legal interventions are reported as not identifiable. For the five-refit prediction-score panel, each refit receives its own status and an aggregate status requires agreement in at least four of five refits; if no category reaches four of five, the aggregate is inconclusive. Path-constrained and falsification-guided results report status counts across ten selected models without a vote. The 50-refit coordinate analysis is descriptive at the refit level and uses the ten trajectory-selected endpoint instances as the policy-distribution units. These instances contain six unique configurations; repeated configurations are retained because they are independent trajectory selections. For the BBBC, LINCS, and LKCP panels, the 32 permutations are repeated measurements within a trained model, not independent sample units. The sci-Plex feedback study uses 32 development maps and 128 final held-out maps, which are likewise repeated measurements; Norman uses a separate set of 32 complete derangements (Appendix~\ref{app:norman-transfer}).

The falsification-guided trainer evaluates $\alpha\in\{0.05,0.10,0.20,0.35,0.50,0.75,1.00\}$ on Fold 3. Predictive noninferiority allows at most a $0.003$ decrease from the control-only predictor in Global \pcc, a $0.005$ decrease in either response block, and a $0.0005$ increase in MSE. Search qualification uses eight fixed Fold-3 replacement maps. The empirical 2.5th percentile of their compound-related $\Delta\mathcal{L}$ values (increases in inverse-standardized joint-output MSE) must exceed both the 95th percentile of absolute pairwise differences among the replaced-input losses and $0.001$ times the reference loss. Here reference loss is the current candidate's Fold-3 MSE with correct inputs. Dose uses the same rule when eligible, with reference loss restricted to eligible rows. These eight development maps are separate from the 32 held-out audit maps. The discovery code groups equal Morgan fingerprints for its replacement maps. Its dose gate requires at least 32 eligible rows and eight fingerprint groups in both the fit partition and Fold 3, with at least $0.1$ log-dose contrast. For each candidate, epoch/scale selection maximizes Fold-3 Global \pcc among qualifying states, breaking ties by lower MSE; if none qualifies, the best predictive state is retained with its qualification status recorded.

Trajectory selection is distinct from this within-candidate checkpoint rule. BBBC047 selects the highest-PCC qualifying candidate, with lower MSE as tie-breaker, or reports no qualified endpoint. BBBC036 prefers predictively noninferior candidates and ranks them by compound-related loss increase, then Global \pcc, lower MSE, and stable candidate identity. If none is noninferior, that ordering retains a predictive-fallback endpoint from the executable candidates. Strict qualification is reported separately for BBBC036. These cohort-specific rules select the ten sources before either post-search boundary; held-out results never reselect them.

Held-out dose testing requires at least 64 eligible rows and coverage of at least 10\% of the partition. The prediction-score and path-constrained tests accept any observed within-compound dose change; the falsification-guided test additionally requires eight compounds and an absolute log-dose difference of $0.1$. Compound identity here is the BRD compound identifier, with sample/batch suffixes removed, rather than a fingerprint-equivalence class. On BBBC047, the resulting coverage is $0/4380$ rows on Fold 4 and $14/3878$ rows from seven compounds on Fold 5 ($0.3610\%$); neither boundary meets the minimum, and neither contains a legal $0.1$-log-dose contrast. Dose is therefore not identifiable in the BBBC047 panels (Table~\ref{tab:all-input-status}). Equal-fingerprint grouping alone would merge distinct released identities. On LINCS and LKCP, all replacement dose values in the 32 frozen maps are observed for the source BRD compound and its available cell/time annotations; the original input tensors and effect estimates are retained. The held-out PCC threshold is computed from fold-specific reference-score variation and is separate from the Fold-3 loss gate.

The prediction-score discovery study contains 80 ten-slot trajectories: two tasks, four search methods, and ten independent trajectory seeds. Each trajectory's budget includes the common initial predictor, and discovery intervals treat paired trajectory identifiers as the statistical unit because all policies share the same candidate-training seed schedule within an identifier. Selected-model comparisons use five paired fitting seeds on the relevant held-out fold. Path-constrained and falsification-guided studies retain all ten trajectory winners and report their distributions. All summaries are computed from fixed experiment logs that retain failed slots, repairs, and provider attempts.

\paragraph{Reproducing tables, designs, and fitted predictions.}
The accompanying package separates three reproducibility targets. Numerical source records and deterministic renderers regenerate reported tables. Frozen-source refitting trains new parameters under the recorded data and selection rules. Checkpoint replay takes the selected source, matching fitted weights, prepared data, and replacement specification to recompute predictions and input-use effects by inference. Its standalone feedback entry point verifies checkpoint and data hashes and writes fresh outputs. Biological matrices and fitted weights are supplied as external inputs; tabulated records include checkpoint identities and counterfactual sufficient statistics. The package's \texttt{docs/REPRODUCIBILITY.md} gives the corresponding commands and required assets.

% Appendix presentation derived from ../generated_all10_paired_discovery.tex
\begin{table}[!htbp]
\centering
\caption{Pre-specified paired comparisons of agentic model discovery. Each comparison pairs methods by the shared trajectory seed ($n=10$ matched blocks). Differences are \discoverymethod minus the listed method; CIs and two-sided $p$-values use paired $t$ tests without multiplicity adjustment.}
\label{tab:all10-paired}
\tblsetup
\begin{lrbox}{\AppendixTableBox}
\begin{tabular}{@{}lcc@{}}
\toprule
\rowcolor{csHeader}
\tblhdr{Comparator} & \tblhdr{Best@10\\$\Delta$ [95\% CI]; $p$} & \tblhdr{Frontier AUC\\$\Delta$ [95\% CI]; $p$} \\
\midrule
\rowcolor{csHeader}\multicolumn{3}{@{}l}{\textbf{BBBC036}} \\
\textsc{AIDE} & \shortstack{+0.0084 {\tblsmall[+0.0044, +0.0124]}\\{\tblsmall $p=9.638\times10^{-4}$}} & \shortstack{+0.0054 {\tblsmall[-0.0003, +0.0111]}\\{\tblsmall $p=6.157\times10^{-2}$}} \\
\textsc{CellForge} & \shortstack{+0.0151 {\tblsmall[+0.0107, +0.0196]}\\{\tblsmall $p=3.24\times10^{-5}$}} & \shortstack{+0.0106 {\tblsmall[+0.0056, +0.0155]}\\{\tblsmall $p=9.601\times10^{-4}$}} \\
\textsc{HarmonyCell} & \shortstack{+0.0046 {\tblsmall[-0.0007, +0.0100]}\\{\tblsmall $p=8.064\times10^{-2}$}} & \shortstack{+0.0025 {\tblsmall[-0.0016, +0.0066]}\\{\tblsmall $p=1.962\times10^{-1}$}} \\
\addlinespace[2pt]
\rowcolor{csHeader}\multicolumn{3}{@{}l}{\textbf{BBBC047}} \\
\textsc{AIDE} & \shortstack{+0.0214 {\tblsmall[+0.0105, +0.0324]}\\{\tblsmall $p=1.636\times10^{-3}$}} & \shortstack{+0.0180 {\tblsmall[+0.0078, +0.0282]}\\{\tblsmall $p=3.18\times10^{-3}$}} \\
\textsc{CellForge} & \shortstack{+0.0181 {\tblsmall[+0.0057, +0.0304]}\\{\tblsmall $p=9.051\times10^{-3}$}} & \shortstack{+0.0159 {\tblsmall[+0.0046, +0.0271]}\\{\tblsmall $p=1.086\times10^{-2}$}} \\
\textsc{HarmonyCell} & \shortstack{+0.0028 {\tblsmall[-0.0168, +0.0223]}\\{\tblsmall $p=7.56\times10^{-1}$}} & \shortstack{+0.0094 {\tblsmall[-0.0033, +0.0220]}\\{\tblsmall $p=1.293\times10^{-1}$}} \\
\bottomrule
\end{tabular}
\end{lrbox}
\AppendixTableOutput{generated_all10_paired_discovery-1}
\end{table}

% Appendix presentation derived from ../generated_all10_paired_frozen.tex
\begin{table}[!htbp]
\centering
\caption{Paired Fold-4 Global-PCC contrasts for frozen sources/designs. Differences are \discoverymethod minus the listed model; each uses five shared fitting seeds evaluated on Fold 4, a two-sided paired $t$ interval, and an unadjusted paired $t$ test.}
\label{tab:all10-frozen-paired}
\tblsetup
\begin{lrbox}{\AppendixTableBox}
\begin{tabular}{@{}lrrrc@{}}
\toprule
\rowcolor{csHeader}
\tblhdr{Comparator} & \tblhdr{$\Delta$ Global PCC} & \tblhdr{95\% CI} & \tblhdr{$p$} & \tblhdr{$n$} \\
\midrule
\rowcolor{csHeader}\multicolumn{5}{@{}l}{\textbf{BBBC036}} \\
\textsc{HarmonyCell} & -0.0015 & [-0.0037, +0.0007] & $1.228\times10^{-1}$ & 5 \\
\textsc{AIDE} & +0.0117 & [+0.0030, +0.0205] & $2.027\times10^{-2}$ & 5 \\
\textsc{CellForge} & +0.0158 & [+0.0095, +0.0222] & $2.307\times10^{-3}$ & 5 \\
TabM & +0.0048 & [+0.0016, +0.0081] & $1.445\times10^{-2}$ & 5 \\
RealMLP & +0.0118 & [-0.0103, +0.0338] & $2.13\times10^{-1}$ & 5 \\
Standard MLP & +0.0193 & [+0.0072, +0.0313] & $1.128\times10^{-2}$ & 5 \\
$h_0$ & +0.0207 & [+0.0130, +0.0283] & $1.733\times10^{-3}$ & 5 \\
Joint TabR & +0.0369 & [+0.0343, +0.0395] & $2.576\times10^{-6}$ & 5 \\
Ridge & +0.2324 & [+0.2301, +0.2347] & $1.033\times10^{-9}$ & 5 \\
\addlinespace[4pt]
\rowcolor{csHeader}\multicolumn{5}{@{}l}{\textbf{BBBC047}} \\
\textsc{HarmonyCell} & +0.0012 & [-0.0083, +0.0108] & $7.394\times10^{-1}$ & 5 \\
\textsc{AIDE} & +0.0337 & [+0.0283, +0.0390] & $6.271\times10^{-5}$ & 5 \\
\textsc{CellForge} & +0.0293 & [+0.0256, +0.0330] & $2.57\times10^{-5}$ & 5 \\
TabM & +0.0181 & [+0.0150, +0.0211] & $7.867\times10^{-5}$ & 5 \\
RealMLP & +0.0202 & [+0.0154, +0.0250] & $3.062\times10^{-4}$ & 5 \\
Standard MLP & +0.0337 & [+0.0275, +0.0398] & $1.103\times10^{-4}$ & 5 \\
$h_0$ & +0.0337 & [+0.0283, +0.0390] & $6.271\times10^{-5}$ & 5 \\
Joint TabR & +0.0500 & [+0.0451, +0.0549] & $9.339\times10^{-6}$ & 5 \\
Ridge & +0.2541 & [+0.2530, +0.2552] & $3.077\times10^{-11}$ & 5 \\
\bottomrule
\end{tabular}
\end{lrbox}
\AppendixTableOutput{generated_all10_paired_frozen-1}
\end{table}

For chemical-generalization contrasts, complete held-out Murcko-scaffold clusters are jointly resampled across all five paired refits, and Global \pcc is recomputed from cluster sufficient statistics. This preserves within-scaffold response dependence and quantifies uncertainty over held-out chemical families. The Holm step-down adjustment applies to the three registered discovery-policy contrasts reported in Table~\ref{tab:holm-discovery}; the broader frozen-predictor comparison is descriptive and reports its unadjusted paired intervals explicitly \citep{holm1979sequential}.

% Appendix presentation derived from ../generated_scaffold_bootstrap.tex
\begin{table}[!htbp]
\centering
\caption{Murcko-scaffold cluster bootstrap for selected-model contrasts. Each interval resamples held-out scaffold groups jointly across five paired refits.}
\label{tab:scaffold-bootstrap}
\tblsetup
\begin{lrbox}{\AppendixTableBox}
\begin{tabular}{@{}lcc@{}}
\toprule
\rowcolor{csHeader}
\tblhdr{Contrast} & \tblhdr{Evaluation} & \tblhdr{$\Delta$ Global PCC\\{[95\% CI]}} \\
\midrule
\rowcolor{csHeader}\multicolumn{3}{@{}l}{\textbf{BBBC036}} \\
CellScientist $-$ control-only & Audit & \tblstack{-0.0002}{[-0.0020, +0.0015]} \\
CellScientist $-$ HarmonyCell & Audit & \tblstack{-0.0014}{[-0.0032, +0.0003]} \\
score-selected $-$ $h_0$ & Replication & \tblstack{+0.0344}{[+0.0195, +0.0514]} \\
\addlinespace[4pt]
\rowcolor{csHeader}\multicolumn{3}{@{}l}{\textbf{BBBC047}} \\
CellScientist $-$ control-only & Audit & \tblstack{+0.0016}{[+0.0001, +0.0032]} \\
CellScientist $-$ HarmonyCell & Audit & \tblstack{+0.0052}{[+0.0034, +0.0069]} \\
score-selected $-$ $h_0$ & Replication & \tblstack{+0.0366}{[+0.0308, +0.0424]} \\
\bottomrule
\end{tabular}
\end{lrbox}
\AppendixTableOutput{generated_scaffold_bootstrap-1}
\end{table}

For the central falsification-guided study, effects are first computed from all 32 fixed within-model input permutations. The conditional interval resamples held-out Murcko scaffolds while holding the ten selected models fixed; the hierarchical sensitivity interval jointly resamples selected models and scaffolds. Models and scaffolds are the resampling units, whereas the 32 permutations remain fixed repeated measurements within a model.

% Appendix presentation derived from ../generated_d3_scaffold_bootstrap.tex
\begin{table}[!htbp]
\centering
\caption{Scaffold-aware uncertainty for falsification-guided discovery. The conditional interval resamples held-out Murcko scaffolds jointly across all ten models; the hierarchical interval resamples both models and scaffolds. Compound PCC and loss effects are the context-averaged factorial contrasts $E_{\rm cmpd}$ and $E^{\mathcal L}_{\rm cmpd}$. All 32 permutations remain fixed.}
\label{tab:d3-scaffold-bootstrap}
\tblsetup
\begin{lrbox}{\AppendixTableBox}
\begin{tabular}{@{}lrcc@{}}
\toprule
\rowcolor{csHeader}
\tblhdr{Quantity} & \tblhdr{Estimate} & \tblhdr{Scaffold\\95\% CI} & \tblhdr{Model$\times$scaffold\\95\% CI} \\
\midrule
\multicolumn{4}{@{}l}{\textbf{BBBC036 / Fold 4}} \\
Full $-$ anchor PCC & +0.0046 & [-0.0042, +0.0165] & [-0.0043, +0.0164] \\
Compound PCC drop & +0.0091 & [+0.0001, +0.0211] & [$+1.78\times 10^{-5}$, +0.0213] \\
Compound target-loss gain & +0.00039 & [+0.00002, +0.00089] & [+0.00002, +0.00090] \\
\addlinespace[3pt]
\multicolumn{4}{@{}l}{\textbf{BBBC036 / Fold 5}} \\
Full $-$ anchor PCC & -0.0020 & [-0.0104, +0.0052] & [-0.0102, +0.0053] \\
Compound PCC drop & +0.0022 & [-0.0050, +0.0090] & [-0.0051, +0.0092] \\
Compound target-loss gain & +0.00009 & [-0.00022, +0.00038] & [-0.00022, +0.00039] \\
\addlinespace[3pt]
\multicolumn{4}{@{}l}{\textbf{BBBC047 / Fold 4}} \\
Full $-$ anchor PCC & +0.0037 & [+0.0018, +0.0060] & [+0.0018, +0.0060] \\
Compound PCC drop & +0.0053 & [+0.0033, +0.0078] & [+0.0033, +0.0078] \\
Compound target-loss gain & +0.00018 & [+0.00011, +0.00026] & [+0.00011, +0.00026] \\
\addlinespace[3pt]
\multicolumn{4}{@{}l}{\textbf{BBBC047 / Fold 5}} \\
Full $-$ anchor PCC & +0.0028 & [+0.0003, +0.0061] & [+0.0002, +0.0061] \\
Compound PCC drop & +0.0048 & [+0.0024, +0.0079] & [+0.0023, +0.0079] \\
Compound target-loss gain & +0.00017 & [+0.00009, +0.00026] & [+0.00009, +0.00026] \\
\bottomrule
\end{tabular}
\end{lrbox}
\AppendixTableOutput{generated_d3_scaffold_bootstrap-1}
\end{table}

The larger BBBC047 cohort retains positive hierarchical intervals for all three quantities on both held-out folds: Full$-$control-only PCC, compound-shuffle PCC drop, and compound target-loss gain. On BBBC036, Fold-4 compound and target-loss intervals remain positive, whereas Fold-5 intervals include zero. Scaffold resampling therefore shows which compound-family effects persist without selecting models or permutations again.

% Appendix presentation derived from ../generated_policy_mapping.tex
\begin{table}[!htbp]
\centering
\caption{Search-method comparison under a common executable model interface. Every method starts from the same $h_0$, receives the same Fold-3 feedback, evaluates ten candidates including $h_0$, and uses the same data, folds, evaluator, training budget, seed rule, repair limit, and model-selection rule.}
\label{tab:policy-mapping}
\tblsetup
\begin{tabularx}{\linewidth}{@{}L{59pt}L{93pt}XL{71pt}@{}}
\toprule
\rowcolor{csHeader}
\tblhdr{Method} & \tblhdr{Information used} & \tblhdr{Revision rule} & \tblhdr{Main difference} \\
\midrule

CellScientist & Incumbent; compact four-record modeling history; diagnostic agenda. & Feedback-conditioned, source-local revision of architecture, fusion, readout, objective, optimizer, or scheduler. & Diagnostic history; incumbent-conditioned revision. \\
\addlinespace[3pt]
AIDE & Incumbent; compact four-record score history. & Greedy source-local code improvement under the shared model interface. & Generic improvement; no CellScientist diagnostic agenda. \\
\addlinespace[3pt]
CellForge & Incumbent; shared Fold-3 feedback. & One-call synthesis of modeling and critic perspectives, followed by one constrained implementation. & Committee framing; no extra proposals or shared CellScientist candidates. \\
\addlinespace[3pt]
HarmonyCell & Independent two-level UCB state; shared Fold-3 feedback. & Hierarchical action selection, followed by constrained leaf implementation. & Tree-directed sequencing; no CellScientist history or candidates. \\
\bottomrule
\end{tabularx}
\end{table}

\paragraph{Statistical reporting and multiplicity.} Trajectory-level Best@10 Global PCC is the primary discovery outcome, with frontier AUC as the co-primary search-efficiency summary. Each contrast pairs the shared trajectory seed and reports its effect estimate, two-sided paired interval, and complete ten-trajectory distribution. Holm adjustment is applied to the three pre-specified policy contrasts within each task and outcome (Table~\ref{tab:holm-discovery}). Response blocks, response-sensitive metrics, cost records, code-check outcomes, and input-shuffle effects retain their own estimands and statistical units rather than being pooled into one omnibus claim. Input permutations are repeated transformations within a trained refit and are never treated as independent biological replicates.

% Appendix presentation derived from ../generated_holm_discovery.tex
\begin{table}[!htbp]
\centering
\caption{Multiplicity reporting for the pre-specified agentic-discovery comparisons. Holm adjustment is applied separately within each task and outcome across the three method contrasts. Values derive from paired $t$ tests over ten shared trajectory seeds.}
\label{tab:holm-discovery}
\tblsetup
\begin{lrbox}{\AppendixTableBox}
\begin{tabular}{@{}lrr@{}}
\toprule
\rowcolor{csHeader}
\tblhdr{Comparator} & \tblhdr{Best@10\\$p_{\mathrm{Holm}}$} & \tblhdr{Frontier AUC\\$p_{\mathrm{Holm}}$} \\
\midrule
\rowcolor{csHeader}\multicolumn{3}{@{}l}{\textbf{BBBC036}} \\
AIDE & $1.9\times10^{-3}$ & $1.231\times10^{-1}$ \\
CellForge & $1\times10^{-4}$ & $2.9\times10^{-3}$ \\
HarmonyCell & $8.06\times10^{-2}$ & $1.962\times10^{-1}$ \\
\addlinespace[4pt]
\rowcolor{csHeader}\multicolumn{3}{@{}l}{\textbf{BBBC047}} \\
AIDE & $4.9\times10^{-3}$ & $9.5\times10^{-3}$ \\
CellForge & $1.81\times10^{-2}$ & $2.17\times10^{-2}$ \\
HarmonyCell & $7.56\times10^{-1}$ & $1.293\times10^{-1}$ \\
\bottomrule
\end{tabular}
\end{lrbox}
\AppendixTableOutput{generated_holm_discovery-1}
\end{table}

\FloatBarrier
\section{Search trajectories, reliability, and cost}
\label{app:receipts}

Table~\ref{tab:main-discovery} summarizes the competitive discovery setting, and Tables~\ref{tab:all10-quality} and~\ref{tab:all10-utility} give the complete trajectory and selected-predictor comparisons. Fixed-model references include the pre-tuned RealMLP \citep{holzmuller2024better}, TabM's parameter-efficient ensemble \citep{gorishniy2025tabm}, and a joint-output adaptation of TabR \citep{gorishniy2024tabr}. Table~\ref{tab:all10-operations} reports valid candidates, failures, repairs, success by budget, model-provider calls, tokens, GPU training time, and trajectory wall time; failed slots and provider attempts remain charged. The RBDE score provides a joint efficiency summary alongside its separate quality and cost columns.

% Appendix presentation derived from ../generated_main_discovery_strip.tex
\begin{table}[!htbp]
\centering
\caption{\textbf{Competitive discovery under a matched ten-slot budget.} Ten-trajectory means $\pm$ 95\% $t$-interval half-widths. Best@$k$: best Global PCC by slot $k$; AUC: mean best-so-far PCC; Valid: executable-slot fraction. Per-search calls and tokens include all slots. Complete predictor and cost results: Appendix~\ref{app:receipts}.}
\label{tab:main-discovery}
\tblsetup
\renewcommand{\arraystretch}{1.0}
\setlength{\tabcolsep}{1pt}
\begin{lrbox}{\AppendixTableBox}
\begin{tabular}{@{}lrrrrrrr@{}}
\toprule
\rowcolor{csHeader}
\textbf{Search} & \tblhdr{Best@3\\$\uparrow$} & \tblhdr{Best@5\\$\uparrow$} & \tblhdr{Best@10\\$\uparrow$} & \tblhdr{AUC $\uparrow$} & \tblhdr{Valid $\uparrow$} & \tblhdr{Calls} & \tblhdr{Tokens\\(K)} \\
\midrule
\rowcolor{csHeader}\multicolumn{8}{@{}l}{\textbf{BBBC036}} \\
 \discoverymethod & \tblpm{\textbf{0.2833}}{0.0075} & \tblpm{\textbf{0.2856}}{0.0052} & \tblpm{\textbf{0.2924}}{0.0037} & \tblpm{\textbf{0.2845}}{0.0046} & \tblpm{0.990}{0.023} & \tblpm{10.50}{1.59} & \tblpm{51.7}{6.4} \\
\textsc{AIDE} & \tblpm{0.2765}{0.0036} & \tblpm{0.2804}{0.0033} & \tblpm{0.2840}{0.0030} & \tblpm{0.2791}{0.0023} & \tblpm{1.000}{0.000} & \tblpm{11.10}{0.98} & \tblpm{41.8}{3.8} \\
\textsc{CellForge} & \tblpm{0.2708}{0.0016} & \tblpm{0.2757}{0.0025} & \tblpm{0.2772}{0.0027} & \tblpm{0.2739}{0.0015} & \tblpm{0.910}{0.092} & \tblpm{16.30}{2.39} & \tblpm{44.2}{7.9} \\
\textsc{HarmonyCell} & \tblpm{\secondbest{0.2803}}{0.0037} & \tblpm{\secondbest{0.2845}}{0.0045} & \tblpm{\secondbest{0.2877}}{0.0038} & \tblpm{\secondbest{0.2820}}{0.0036} & \tblpm{0.950}{0.051} & \tblpm{11.80}{1.46} & \tblpm{38.8}{4.7} \\
\midrule
\rowcolor{csHeader}\multicolumn{8}{@{}l}{\textbf{BBBC047}} \\
 \discoverymethod & \tblpm{\textbf{0.3052}}{0.0127} & \tblpm{\textbf{0.3084}}{0.0127} & \tblpm{\textbf{0.3097}}{0.0119} & \tblpm{\textbf{0.3058}}{0.0111} & \tblpm{0.940}{0.077} & \tblpm{12.50}{3.20} & \tblpm{59.2}{10.4} \\
\textsc{AIDE} & \tblpm{0.2875}{0.0015} & \tblpm{0.2882}{0.0014} & \tblpm{0.2883}{0.0012} & \tblpm{0.2878}{0.0013} & \tblpm{0.990}{0.023} & \tblpm{12.30}{1.17} & \tblpm{45.2}{4.4} \\
\textsc{CellForge} & \tblpm{0.2891}{0.0015} & \tblpm{0.2901}{0.0010} & \tblpm{0.2917}{0.0011} & \tblpm{0.2899}{0.0008} & \tblpm{0.940}{0.090} & \tblpm{12.90}{2.75} & \tblpm{36.1}{7.4} \\
\textsc{HarmonyCell} & \tblpm{\secondbest{0.2918}}{0.0043} & \tblpm{\secondbest{0.2965}}{0.0035} & \tblpm{\secondbest{0.3070}}{0.0083} & \tblpm{\secondbest{0.2964}}{0.0034} & \tblpm{0.950}{0.051} & \tblpm{13.20}{1.99} & \tblpm{41.6}{5.8} \\
\bottomrule
\end{tabular}
\end{lrbox}
\AppendixTableOutput{generated_main_discovery_strip-1}
\tblnote{Bold/underline: best/second distinct displayed predictive mean within each task or fold. Ties share a rank; interval bounds and diagnostic/resource columns are unranked.}
\end{table}

% Appendix presentation derived from ../generated_all10_discovery_quality.tex
\begin{table}[!htbp]
\centering
\caption{\textbf{CellScientist supplies a competitive discovery frontier.} Each cell summarizes ten independent trajectories with ten evaluated-candidate slots including $h_0$; values are mean $\pm$ 95\% $t$-interval half-width. AIDE, CellForge, and HarmonyCell denote the common-executor policy instantiations defined in Section~\ref{sec:experiments}.}
\label{tab:all10-quality}
\tblsetup
\setlength{\tabcolsep}{3pt}
\begin{lrbox}{\AppendixTableBox}
\begin{tabular}{@{}lrrrrrrrr@{}}
\toprule
\rowcolor{csHeader}
\tblhdr{Method} & \tblhdr{Best@3\\PCC $\uparrow$} & \tblhdr{Best@5\\$\uparrow$} & \tblhdr{Best@10\\$\uparrow$} & \tblhdr{Best MSE\\$\downarrow$} & \tblhdr{CP PCC\\$\uparrow$} & \tblhdr{L1000 PCC\\$\uparrow$} & \tblhdr{Frontier\\AUC $\uparrow$} & \tblhdr{Lift AUC\\$\uparrow$} \\
\midrule
\rowcolor{csHeader}\multicolumn{9}{@{}l}{\textbf{BBBC036}} \\
 \discoverymethod & \tblstack{\textbf{0.2833}}{$\pm$ 0.0075} & \tblstack{\textbf{0.2856}}{$\pm$ 0.0052} & \tblstack{\textbf{0.2924}}{$\pm$ 0.0037} & \tblstack{\textbf{0.0607}}{$\pm$ 0.0001} & \tblstack{\textbf{0.2824}}{$\pm$ 0.0015} & \tblstack{\textbf{0.2965}}{$\pm$ 0.0052} & \tblstack{\textbf{0.2845}}{$\pm$ 0.0046} & \tblstack{\textbf{0.0172}}{$\pm$ 0.0046} \\
\textsc{AIDE} & \tblstack{0.2765}{$\pm$ 0.0036} & \tblstack{0.2804}{$\pm$ 0.0033} & \tblstack{0.2840}{$\pm$ 0.0030} & \tblstack{0.0612}{$\pm$ 0.0001} & \tblstack{0.2762}{$\pm$ 0.0028} & \tblstack{0.2875}{$\pm$ 0.0034} & \tblstack{0.2791}{$\pm$ 0.0023} & \tblstack{0.0118}{$\pm$ 0.0028} \\
\textsc{CellForge} & \tblstack{0.2708}{$\pm$ 0.0016} & \tblstack{0.2757}{$\pm$ 0.0025} & \tblstack{0.2772}{$\pm$ 0.0027} & \tblstack{0.0617}{$\pm$ 0.0003} & \tblstack{0.2687}{$\pm$ 0.0038} & \tblstack{0.2810}{$\pm$ 0.0032} & \tblstack{0.2739}{$\pm$ 0.0015} & \tblstack{0.0066}{$\pm$ 0.0020} \\
\textsc{HarmonyCell} & \tblstack{\secondbest{0.2803}}{$\pm$ 0.0037} & \tblstack{\secondbest{0.2845}}{$\pm$ 0.0045} & \tblstack{\secondbest{0.2877}}{$\pm$ 0.0038} & \tblstack{\secondbest{0.0609}}{$\pm$ 0.0002} & \tblstack{\secondbest{0.2821}}{$\pm$ 0.0034} & \tblstack{\secondbest{0.2905}}{$\pm$ 0.0045} & \tblstack{\secondbest{0.2820}}{$\pm$ 0.0036} & \tblstack{\secondbest{0.0147}}{$\pm$ 0.0038} \\
\addlinespace[4pt]
\rowcolor{csHeader}\multicolumn{9}{@{}l}{\textbf{BBBC047}} \\
 \discoverymethod & \tblstack{\textbf{0.3052}}{$\pm$ 0.0127} & \tblstack{\textbf{0.3084}}{$\pm$ 0.0127} & \tblstack{\textbf{0.3097}}{$\pm$ 0.0119} & \tblstack{\textbf{0.0451}}{$\pm$ 0.0004} & \tblstack{\secondbest{0.3351}}{$\pm$ 0.0048} & \tblstack{\textbf{0.2806}}{$\pm$ 0.0212} & \tblstack{\textbf{0.3058}}{$\pm$ 0.0111} & \tblstack{\textbf{0.0201}}{$\pm$ 0.0108} \\
\textsc{AIDE} & \tblstack{0.2875}{$\pm$ 0.0015} & \tblstack{0.2882}{$\pm$ 0.0014} & \tblstack{0.2883}{$\pm$ 0.0012} & \tblstack{0.0458}{$\pm$ 0.0000} & \tblstack{0.3269}{$\pm$ 0.0028} & \tblstack{0.2433}{$\pm$ 0.0025} & \tblstack{0.2878}{$\pm$ 0.0013} & \tblstack{0.0021}{$\pm$ 0.0014} \\
\textsc{CellForge} & \tblstack{0.2891}{$\pm$ 0.0015} & \tblstack{0.2901}{$\pm$ 0.0010} & \tblstack{0.2917}{$\pm$ 0.0011} & \tblstack{0.0457}{$\pm$ 0.0001} & \tblstack{0.3315}{$\pm$ 0.0040} & \tblstack{0.2449}{$\pm$ 0.0049} & \tblstack{0.2899}{$\pm$ 0.0008} & \tblstack{0.0042}{$\pm$ 0.0016} \\
\textsc{HarmonyCell} & \tblstack{\secondbest{0.2918}}{$\pm$ 0.0043} & \tblstack{\secondbest{0.2965}}{$\pm$ 0.0035} & \tblstack{\secondbest{0.3070}}{$\pm$ 0.0083} & \tblstack{\secondbest{0.0452}}{$\pm$ 0.0003} & \tblstack{\textbf{0.3373}}{$\pm$ 0.0041} & \tblstack{\secondbest{0.2724}}{$\pm$ 0.0151} & \tblstack{\secondbest{0.2964}}{$\pm$ 0.0034} & \tblstack{\secondbest{0.0107}}{$\pm$ 0.0039} \\
\bottomrule
\end{tabular}
\end{lrbox}
\AppendixTableOutput{generated_all10_discovery_quality-1}
\tblnote{Bold/underline: best/second distinct displayed predictive mean within each task or fold. Ties share a rank; interval bounds and diagnostic/resource columns are unranked.}
\end{table}

% Appendix presentation derived from ../generated_all10_frozen_utility.tex
\begin{table}[!htbp]
\centering
\caption{\textbf{Selected sources remain competitive after five-seed refitting.} All models use one joint multi-output predictor and the same input, target, partition, preprocessing, and evaluator contract; Fold-3-selected sources are evaluated on Fold 4. Values are mean $\pm$ 95\% $t$-interval half-width.}
\label{tab:all10-utility}
\tblsetup
\setlength{\tabcolsep}{2.5pt}
\begin{lrbox}{\AppendixTableBox}
\begin{tabular}{@{}lrrrrrrr@{}}
\toprule
\rowcolor{csHeader}
\tblhdr{Model} & \tblhdr{Global PCC\\$\uparrow$} & \tblhdr{MSE\\$\downarrow$} & \tblhdr{CP PCC\\$\uparrow$} & \tblhdr{L1000 PCC\\$\uparrow$} & \tblhdr{Parameters} & \tblhdr{Train\\(s)} & \tblhdr{Selected\\epoch} \\
\midrule
\rowcolor{csHeader}\multicolumn{8}{@{}l}{\textbf{BBBC036}} \\
 \discoverymethod & \tblstack{\secondbest{0.3101}}{$\pm$ 0.0023} & \tblstack{\textbf{0.0586}}{$\pm$ 0.0001} & \tblstack{0.3107}{$\pm$ 0.0025} & \tblstack{\secondbest{0.3098}}{$\pm$ 0.0032} & 4.00M & \tblstack{7.2}{$\pm$ 1.2} & \tblstack{21.8}{$\pm$ 17.0} \\
\textsc{AIDE} & \tblstack{0.2984}{$\pm$ 0.0091} & \tblstack{0.0593}{$\pm$ 0.0006} & \tblstack{0.3085}{$\pm$ 0.0043} & \tblstack{0.2950}{$\pm$ 0.0124} & 4.42M & \tblstack{6.3}{$\pm$ 1.2} & \tblstack{4.8}{$\pm$ 1.0} \\
\textsc{CellForge} & \tblstack{0.2942}{$\pm$ 0.0043} & \tblstack{0.0593}{$\pm$ 0.0002} & \tblstack{0.3052}{$\pm$ 0.0048} & \tblstack{0.2902}{$\pm$ 0.0056} & 3.82M & \tblstack{6.2}{$\pm$ 1.3} & \tblstack{2.0}{$\pm$ 0.0} \\
\textsc{HarmonyCell} & \tblstack{\textbf{0.3116}}{$\pm$ 0.0013} & \tblstack{\textbf{0.0586}}{$\pm$ 0.0001} & \tblstack{0.3117}{$\pm$ 0.0024} & \tblstack{\textbf{0.3115}}{$\pm$ 0.0020} & 2.06M & \tblstack{6.7}{$\pm$ 1.4} & \tblstack{17.2}{$\pm$ 12.5} \\
$h_0$ & \tblstack{0.2894}{$\pm$ 0.0081} & \tblstack{0.0597}{$\pm$ 0.0003} & \tblstack{0.2984}{$\pm$ 0.0048} & \tblstack{0.2860}{$\pm$ 0.0098} & 1.14M & \tblstack{5.9}{$\pm$ 1.2} & \tblstack{3.2}{$\pm$ 0.6} \\
Standard MLP & \tblstack{0.2908}{$\pm$ 0.0114} & \tblstack{0.0594}{$\pm$ 0.0004} & \tblstack{0.3058}{$\pm$ 0.0089} & \tblstack{0.2845}{$\pm$ 0.0125} & 2.42M & \tblstack{5.8}{$\pm$ 1.2} & \tblstack{1.8}{$\pm$ 0.6} \\
RealMLP & \tblstack{0.2983}{$\pm$ 0.0202} & \tblstack{0.0592}{$\pm$ 0.0007} & \tblstack{\secondbest{0.3170}}{$\pm$ 0.0129} & \tblstack{0.2902}{$\pm$ 0.0260} & 3.47M & \tblstack{5.9}{$\pm$ 0.7} & -- \\
TabM & \tblstack{0.3053}{$\pm$ 0.0039} & \tblstack{\secondbest{0.0589}}{$\pm$ 0.0001} & \tblstack{\textbf{0.3175}}{$\pm$ 0.0063} & \tblstack{0.3004}{$\pm$ 0.0044} & 2.37M & \tblstack{11.4}{$\pm$ 1.7} & \tblstack{2.8}{$\pm$ 1.0} \\
Joint TabR & \tblstack{0.2732}{$\pm$ 0.0033} & \tblstack{0.0604}{$\pm$ 0.0001} & \tblstack{0.2835}{$\pm$ 0.0036} & \tblstack{0.2687}{$\pm$ 0.0044} & 2.17M & \tblstack{2.6}{$\pm$ 0.3} & \tblstack{7.6}{$\pm$ 1.1} \\
Ridge & \tblstack{0.0777}{$\pm$ 0.0000} & \tblstack{0.2048}{$\pm$ 0.0000} & \tblstack{0.0812}{$\pm$ 0.0000} & \tblstack{0.0763}{$\pm$ 0.0000} & 4.14M & \tblstack{1.5}{$\pm$ 0.4} & -- \\
\addlinespace[4pt]
\rowcolor{csHeader}\multicolumn{8}{@{}l}{\textbf{BBBC047}} \\
 \discoverymethod & \tblstack{\textbf{0.3033}}{$\pm$ 0.0011} & \tblstack{\textbf{0.0457}}{$\pm$ 0.0000} & \tblstack{0.3254}{$\pm$ 0.0010} & \tblstack{\textbf{0.2787}}{$\pm$ 0.0015} & 1.78M & \tblstack{24.3}{$\pm$ 3.7} & \tblstack{36.6}{$\pm$ 10.5} \\
\textsc{AIDE} & \tblstack{0.2696}{$\pm$ 0.0047} & \tblstack{0.0468}{$\pm$ 0.0002} & \tblstack{0.3167}{$\pm$ 0.0062} & \tblstack{0.2125}{$\pm$ 0.0101} & 1.20M & \tblstack{9.5}{$\pm$ 0.7} & \tblstack{2.0}{$\pm$ 0.9} \\
\textsc{CellForge} & \tblstack{0.2740}{$\pm$ 0.0035} & \tblstack{0.0467}{$\pm$ 0.0001} & \tblstack{0.3178}{$\pm$ 0.0061} & \tblstack{0.2216}{$\pm$ 0.0060} & 2.53M & \tblstack{9.6}{$\pm$ 0.8} & \tblstack{1.8}{$\pm$ 1.0} \\
\textsc{HarmonyCell} & \tblstack{\secondbest{0.3021}}{$\pm$ 0.0098} & \tblstack{\secondbest{0.0458}}{$\pm$ 0.0003} & \tblstack{\secondbest{0.3265}}{$\pm$ 0.0016} & \tblstack{\secondbest{0.2743}}{$\pm$ 0.0196} & 2.28M & \tblstack{26.0}{$\pm$ 9.2} & \tblstack{41.8}{$\pm$ 27.1} \\
$h_0$ & \tblstack{0.2696}{$\pm$ 0.0047} & \tblstack{0.0468}{$\pm$ 0.0002} & \tblstack{0.3167}{$\pm$ 0.0062} & \tblstack{0.2125}{$\pm$ 0.0101} & 1.20M & \tblstack{9.3}{$\pm$ 0.9} & \tblstack{2.0}{$\pm$ 0.9} \\
Standard MLP & \tblstack{0.2696}{$\pm$ 0.0059} & \tblstack{0.0468}{$\pm$ 0.0002} & \tblstack{0.3167}{$\pm$ 0.0062} & \tblstack{0.2127}{$\pm$ 0.0067} & 2.53M & \tblstack{9.6}{$\pm$ 0.9} & \tblstack{1.2}{$\pm$ 0.6} \\
RealMLP & \tblstack{0.2831}{$\pm$ 0.0042} & \tblstack{0.0463}{$\pm$ 0.0001} & \tblstack{0.3259}{$\pm$ 0.0038} & \tblstack{0.2326}{$\pm$ 0.0049} & 3.62M & \tblstack{13.7}{$\pm$ 0.8} & -- \\
TabM & \tblstack{0.2852}{$\pm$ 0.0033} & \tblstack{0.0463}{$\pm$ 0.0001} & \tblstack{\textbf{0.3282}}{$\pm$ 0.0023} & \tblstack{0.2350}{$\pm$ 0.0105} & 2.51M & \tblstack{16.7}{$\pm$ 2.1} & \tblstack{2.6}{$\pm$ 0.7} \\
Joint TabR & \tblstack{0.2533}{$\pm$ 0.0049} & \tblstack{0.0475}{$\pm$ 0.0002} & \tblstack{0.2951}{$\pm$ 0.0050} & \tblstack{0.2027}{$\pm$ 0.0066} & 2.25M & \tblstack{10.8}{$\pm$ 1.1} & \tblstack{4.6}{$\pm$ 0.7} \\
Ridge & \tblstack{0.0492}{$\pm$ 0.0000} & \tblstack{0.3880}{$\pm$ 0.0000} & \tblstack{0.0547}{$\pm$ 0.0000} & \tblstack{0.0438}{$\pm$ 0.0000} & 4.62M & \tblstack{1.3}{$\pm$ 0.2} & -- \\
\bottomrule
\end{tabular}
\end{lrbox}
\AppendixTableOutput{generated_all10_frozen_utility-1}
\tblnote{Bold/underline: best/second distinct displayed predictive mean within each task or fold. Ties share a rank; interval bounds and diagnostic/resource columns are unranked.}
\end{table}

% Appendix presentation derived from ../generated_discovery_configuration.tex
\begin{table}[!htbp]
\centering
\caption{Model-discovery configuration. All agentic discovery calls covered by this configuration use the recorded DeepSeek V4 Flash model identifier through an OpenAI-compatible chat-completions interface. No fallback is enabled. Each request explicitly sets the model, messages, temperature, and maximum output length; other sampling fields use provider defaults. For reproducibility, prompt-construction code, search history, request/response SHA-256 hashes, returned code or design, response model, and token usage are logged for every call.}
\label{tab:discovery-configuration}
\tblsetup
\begin{lrbox}{\AppendixTableBox}
\begin{tabular}{@{}llrr@{}}
\toprule
\rowcolor{csHeader}
\tblhdr{Formulation} & \tblhdr{Generated artifact} & \tblhdr{Max output} & \tblhdr{Slots} \\
\midrule
Prediction-score policies & Executable source & 9000 & 10 \\
Path-constrained & Structured design card & 2400 & 10 \\
Falsification-guided & Semantic design card & 1800 & 10 \\
\bottomrule
\end{tabular}
\end{lrbox}
\AppendixTableOutput{generated_discovery_configuration-1}
\tblnote{Shared settings: model \texttt{deepseek-v4-flash}; temperature 0.2; reasoning effort not set; fallback disabled. Max output is the token limit per request.}
\end{table}

% Appendix presentation derived from ../generated_all10_discovery_operations.tex
\begin{table}[!htbp]
\centering
\caption{Complete agentic-discovery execution and cost receipts. Values are mean $\pm$ 95\% $t$-interval half-width across ten trajectories. Success@$k$ counts trajectories (out of ten) whose best Fold3 Global PCC by candidate slot $k$ strictly exceeds that of the initial $h_0$; the candidate-slot budget includes $h_0$. RBDE is the registered valid-candidate rate times the non-negative 95\% lower confidence bound of Frontier-Lift AUC per ten minutes of total trajectory wall time. Tokens are thousands and exclude only trajectories without a complete provider-usage receipt; all other fields have $n=10$.}
\label{tab:all10-operations}
\tblsetup
\noindent\textbf{A. Execution quality and budget efficiency}\par\vspace{2pt}
\begin{lrbox}{\AppendixTableBox}
\begin{tabular}{@{}lrrrrc@{}}
\toprule
\rowcolor{csHeader}
\tblhdr{Method} & \tblhdr{RBDE\\$\uparrow$} & \tblhdr{Valid rate\\$\uparrow$} & \tblhdr{Failed slots\\$\downarrow$} & \tblhdr{Repairs\\$\downarrow$} & \tblhdr{Success@3/5/10\\$\uparrow$} \\
\midrule
\rowcolor{csHeader}\multicolumn{6}{@{}l}{\textbf{BBBC036}} \\
 \discoverymethod & 0.0104 & \tblinlinepm{0.990}{$\pm$ 0.023} & \tblinlinepm{0.10}{$\pm$ 0.23} & \tblinlinepm{1.50}{$\pm$ 1.59} & 8/10/10 \\
\textsc{AIDE} & 0.0096 & \tblinlinepm{1.000}{$\pm$ 0.000} & \tblinlinepm{0.00}{$\pm$ 0.00} & \tblinlinepm{2.10}{$\pm$ 0.98} & 9/10/10 \\
\textsc{CellForge} & 0.0044 & \tblinlinepm{0.910}{$\pm$ 0.092} & \tblinlinepm{0.90}{$\pm$ 0.92} & \tblinlinepm{7.30}{$\pm$ 2.39} & 8/10/10 \\
\textsc{HarmonyCell} & 0.0102 & \tblinlinepm{0.950}{$\pm$ 0.051} & \tblinlinepm{0.50}{$\pm$ 0.51} & \tblinlinepm{2.80}{$\pm$ 1.46} & 9/10/10 \\
\addlinespace[4pt]
\rowcolor{csHeader}\multicolumn{6}{@{}l}{\textbf{BBBC047}} \\
 \discoverymethod & 0.0064 & \tblinlinepm{0.940}{$\pm$ 0.077} & \tblinlinepm{0.60}{$\pm$ 0.77} & \tblinlinepm{3.50}{$\pm$ 3.20} & 7/8/9 \\
\textsc{AIDE} & 0.0008 & \tblinlinepm{0.990}{$\pm$ 0.023} & \tblinlinepm{0.10}{$\pm$ 0.23} & \tblinlinepm{3.30}{$\pm$ 1.17} & 5/8/9 \\
\textsc{CellForge} & 0.0027 & \tblinlinepm{0.940}{$\pm$ 0.090} & \tblinlinepm{0.60}{$\pm$ 0.90} & \tblinlinepm{3.90}{$\pm$ 2.75} & 8/9/9 \\
\textsc{HarmonyCell} & 0.0057 & \tblinlinepm{0.950}{$\pm$ 0.051} & \tblinlinepm{0.50}{$\pm$ 0.51} & \tblinlinepm{4.20}{$\pm$ 1.99} & 8/10/10 \\
\bottomrule
\end{tabular}
\end{lrbox}
\AppendixTableOutput{generated_all10_discovery_operations-1}
\par\vspace{6pt}\noindent\textbf{B. Language-model and compute costs}\par\vspace{2pt}
\begin{lrbox}{\AppendixTableBox}
\begin{tabular}{@{}lrrrr@{}}
\toprule
\rowcolor{csHeader}
\tblhdr{Method} & \tblhdr{Calls} & \tblhdr{Tokens (K)} & \tblhdr{GPU training (s)} & \tblhdr{Wall time (s)} \\
\midrule
\rowcolor{csHeader}\multicolumn{5}{@{}l}{\textbf{BBBC036}} \\
 \discoverymethod & \tblinlinepm{10.50}{$\pm$ 1.59} & \tblinlinepm{51.7}{$\pm$ 6.4} & \tblinlinepm{19.8}{$\pm$ 2.2} & \tblinlinepm{716.3}{$\pm$ 101.3} \\
\textsc{AIDE} & \tblinlinepm{11.10}{$\pm$ 0.98} & \tblinlinepm{41.8}{$\pm$ 3.8} & \tblinlinepm{16.6}{$\pm$ 2.5} & \tblinlinepm{559.0}{$\pm$ 212.1} \\
\textsc{CellForge} & \tblinlinepm{16.30}{$\pm$ 2.39} & \tblinlinepm{44.2}{$\pm$ 7.9} & \tblinlinepm{15.7}{$\pm$ 1.4} & \tblinlinepm{578.0}{$\pm$ 86.4} \\
\textsc{HarmonyCell} & \tblinlinepm{11.80}{$\pm$ 1.46} & \tblinlinepm{38.8}{$\pm$ 4.7} & \tblinlinepm{16.8}{$\pm$ 1.1} & \tblinlinepm{610.8}{$\pm$ 158.2} \\
\addlinespace[4pt]
\rowcolor{csHeader}\multicolumn{5}{@{}l}{\textbf{BBBC047}} \\
 \discoverymethod & \tblinlinepm{12.50}{$\pm$ 3.20} & \tblinlinepm{59.2}{$\pm$ 10.4} & \tblinlinepm{111.7}{$\pm$ 21.3} & \tblinlinepm{821.9}{$\pm$ 116.0} \\
\textsc{AIDE} & \tblinlinepm{12.30}{$\pm$ 1.17} & \tblinlinepm{45.2}{$\pm$ 4.4} & \tblinlinepm{60.3}{$\pm$ 3.9} & \tblinlinepm{533.0}{$\pm$ 152.3} \\
\textsc{CellForge} & \tblinlinepm{12.90}{$\pm$ 2.75} & \tblinlinepm{36.1}{$\pm$ 7.4} & \tblinlinepm{56.7}{$\pm$ 5.8} & \tblinlinepm{533.0}{$\pm$ 83.6} \\
\textsc{HarmonyCell} & \tblinlinepm{13.20}{$\pm$ 1.99} & \tblinlinepm{41.6}{$\pm$ 5.8} & \tblinlinepm{72.9}{$\pm$ 11.0} & \tblinlinepm{680.5}{$\pm$ 228.0} \\
\bottomrule
\end{tabular}
\end{lrbox}
\AppendixTableOutput{generated_all10_discovery_operations-2}
\end{table}

% Appendix presentation derived from ../generated_all10_response_metrics.tex
\begin{table}[!htbp]
\centering
\caption{Diagnostic response-block metrics for the frozen joint predictors on Fold 4. Global PCC is the registered joint selection endpoint; CP and L1000 metrics localize response behavior by readout block. Top-$K$ response dimensions are registered from Fold-1--2 response magnitude. Values are mean $\pm$ 95\% $t$-interval half-width over five paired seeds.}
\label{tab:all10-response}
\tblsetup
\setlength{\tabcolsep}{2.5pt}
\begin{lrbox}{\AppendixTableBox}
\begin{tabular}{@{}lrrrrrrrr@{}}
\toprule
\rowcolor{csHeader}
 & \multicolumn{4}{c}{\cellcolor{csCP}\textbf{CP response block}} & \multicolumn{4}{c}{\cellcolor{csL1000}\textbf{L1000 response block}} \\
\cmidrule(lr){2-5}\cmidrule(lr){6-9}
\rowcolor{csHeader}
\tblhdr{Model} & \tblhdr{PCC\\@20\\$\uparrow$} & \tblhdr{PCC\\@50\\$\uparrow$} & \tblhdr{RMSE\\@20\\$\downarrow$} & \tblhdr{RMSE\\@50\\$\downarrow$} & \tblhdr{PCC\\@20\\$\uparrow$} & \tblhdr{PCC\\@50\\$\uparrow$} & \tblhdr{RMSE\\@20\\$\downarrow$} & \tblhdr{RMSE\\@50\\$\downarrow$} \\
\midrule
\rowcolor{csHeader}\multicolumn{9}{@{}l}{\textbf{BBBC036}} \\
 \discoverymethod & \tblstack{0.4620}{$\pm$ 0.0035} & \tblstack{0.4454}{$\pm$ 0.0033} & \tblstack{0.2411}{$\pm$ 0.0005} & \tblstack{0.2385}{$\pm$ 0.0004} & \tblstack{0.2828}{$\pm$ 0.0051} & \tblstack{0.3167}{$\pm$ 0.0043} & \tblstack{0.5135}{$\pm$ 0.0008} & \tblstack{0.4497}{$\pm$ 0.0007} \\
\textsc{AIDE} & \tblstack{0.4636}{$\pm$ 0.0024} & \tblstack{0.4445}{$\pm$ 0.0025} & \tblstack{0.2408}{$\pm$ 0.0006} & \tblstack{0.2386}{$\pm$ 0.0004} & \tblstack{0.2611}{$\pm$ 0.0200} & \tblstack{0.3005}{$\pm$ 0.0160} & \tblstack{0.5190}{$\pm$ 0.0054} & \tblstack{0.4540}{$\pm$ 0.0048} \\
\textsc{CellForge} & \tblstack{0.4607}{$\pm$ 0.0050} & \tblstack{0.4430}{$\pm$ 0.0034} & \tblstack{0.2412}{$\pm$ 0.0006} & \tblstack{0.2388}{$\pm$ 0.0005} & \tblstack{0.2598}{$\pm$ 0.0081} & \tblstack{0.2966}{$\pm$ 0.0099} & \tblstack{0.5176}{$\pm$ 0.0011} & \tblstack{0.4531}{$\pm$ 0.0013} \\
\textsc{HarmonyCell} & \tblstack{0.4633}{$\pm$ 0.0023} & \tblstack{0.4464}{$\pm$ 0.0024} & \tblstack{0.2409}{$\pm$ 0.0005} & \tblstack{0.2383}{$\pm$ 0.0004} & \tblstack{0.2823}{$\pm$ 0.0040} & \tblstack{0.3170}{$\pm$ 0.0031} & \tblstack{0.5137}{$\pm$ 0.0006} & \tblstack{0.4497}{$\pm$ 0.0004} \\
$h_0$ & \tblstack{0.4526}{$\pm$ 0.0034} & \tblstack{0.4357}{$\pm$ 0.0032} & \tblstack{0.2426}{$\pm$ 0.0005} & \tblstack{0.2399}{$\pm$ 0.0005} & \tblstack{0.2555}{$\pm$ 0.0079} & \tblstack{0.2983}{$\pm$ 0.0034} & \tblstack{0.5210}{$\pm$ 0.0056} & \tblstack{0.4545}{$\pm$ 0.0031} \\
Standard MLP & \tblstack{0.4608}{$\pm$ 0.0077} & \tblstack{0.4404}{$\pm$ 0.0070} & \tblstack{0.2412}{$\pm$ 0.0011} & \tblstack{0.2391}{$\pm$ 0.0009} & \tblstack{0.2473}{$\pm$ 0.0048} & \tblstack{0.2923}{$\pm$ 0.0111} & \tblstack{0.5197}{$\pm$ 0.0006} & \tblstack{0.4536}{$\pm$ 0.0017} \\
RealMLP & \tblstack{0.4668}{$\pm$ 0.0081} & \tblstack{0.4497}{$\pm$ 0.0080} & \tblstack{0.2402}{$\pm$ 0.0013} & \tblstack{0.2378}{$\pm$ 0.0011} & \tblstack{0.2476}{$\pm$ 0.0332} & \tblstack{0.2945}{$\pm$ 0.0320} & \tblstack{0.5200}{$\pm$ 0.0053} & \tblstack{0.4540}{$\pm$ 0.0047} \\
TabM & \tblstack{0.4640}{$\pm$ 0.0023} & \tblstack{0.4477}{$\pm$ 0.0021} & \tblstack{0.2408}{$\pm$ 0.0005} & \tblstack{0.2382}{$\pm$ 0.0003} & \tblstack{0.2603}{$\pm$ 0.0079} & \tblstack{0.3066}{$\pm$ 0.0033} & \tblstack{0.5180}{$\pm$ 0.0031} & \tblstack{0.4518}{$\pm$ 0.0016} \\
Joint TabR & \tblstack{0.4433}{$\pm$ 0.0041} & \tblstack{0.4235}{$\pm$ 0.0045} & \tblstack{0.2437}{$\pm$ 0.0005} & \tblstack{0.2414}{$\pm$ 0.0006} & \tblstack{0.2389}{$\pm$ 0.0124} & \tblstack{0.2812}{$\pm$ 0.0120} & \tblstack{0.5231}{$\pm$ 0.0024} & \tblstack{0.4567}{$\pm$ 0.0020} \\
Ridge & \tblstack{0.1172}{$\pm$ 0.0000} & \tblstack{0.1153}{$\pm$ 0.0000} & \tblstack{0.4678}{$\pm$ 0.0000} & \tblstack{0.4603}{$\pm$ 0.0000} & \tblstack{0.0807}{$\pm$ 0.0000} & \tblstack{0.0903}{$\pm$ 0.0000} & \tblstack{0.9852}{$\pm$ 0.0000} & \tblstack{0.8401}{$\pm$ 0.0000} \\
\addlinespace[4pt]
\rowcolor{csHeader}\multicolumn{9}{@{}l}{\textbf{BBBC047}} \\
 \discoverymethod & \tblstack{0.4003}{$\pm$ 0.0028} & \tblstack{0.4061}{$\pm$ 0.0018} & \tblstack{0.2689}{$\pm$ 0.0002} & \tblstack{0.2547}{$\pm$ 0.0002} & \tblstack{0.3262}{$\pm$ 0.0025} & \tblstack{0.3154}{$\pm$ 0.0019} & \tblstack{0.3703}{$\pm$ 0.0004} & \tblstack{0.3250}{$\pm$ 0.0003} \\
\textsc{AIDE} & \tblstack{0.4127}{$\pm$ 0.0124} & \tblstack{0.4078}{$\pm$ 0.0083} & \tblstack{0.2676}{$\pm$ 0.0022} & \tblstack{0.2549}{$\pm$ 0.0015} & \tblstack{0.2577}{$\pm$ 0.0090} & \tblstack{0.2486}{$\pm$ 0.0060} & \tblstack{0.3789}{$\pm$ 0.0009} & \tblstack{0.3323}{$\pm$ 0.0006} \\
\textsc{CellForge} & \tblstack{0.4117}{$\pm$ 0.0095} & \tblstack{0.4069}{$\pm$ 0.0032} & \tblstack{0.2678}{$\pm$ 0.0010} & \tblstack{0.2550}{$\pm$ 0.0007} & \tblstack{0.2684}{$\pm$ 0.0051} & \tblstack{0.2583}{$\pm$ 0.0041} & \tblstack{0.3776}{$\pm$ 0.0009} & \tblstack{0.3312}{$\pm$ 0.0007} \\
\textsc{HarmonyCell} & \tblstack{0.4029}{$\pm$ 0.0049} & \tblstack{0.4079}{$\pm$ 0.0026} & \tblstack{0.2686}{$\pm$ 0.0009} & \tblstack{0.2545}{$\pm$ 0.0004} & \tblstack{0.3234}{$\pm$ 0.0156} & \tblstack{0.3130}{$\pm$ 0.0148} & \tblstack{0.3705}{$\pm$ 0.0021} & \tblstack{0.3252}{$\pm$ 0.0016} \\
$h_0$ & \tblstack{0.4127}{$\pm$ 0.0124} & \tblstack{0.4078}{$\pm$ 0.0083} & \tblstack{0.2676}{$\pm$ 0.0022} & \tblstack{0.2549}{$\pm$ 0.0015} & \tblstack{0.2577}{$\pm$ 0.0090} & \tblstack{0.2486}{$\pm$ 0.0060} & \tblstack{0.3789}{$\pm$ 0.0009} & \tblstack{0.3323}{$\pm$ 0.0006} \\
Standard MLP & \tblstack{0.4125}{$\pm$ 0.0157} & \tblstack{0.4083}{$\pm$ 0.0089} & \tblstack{0.2678}{$\pm$ 0.0023} & \tblstack{0.2548}{$\pm$ 0.0013} & \tblstack{0.2626}{$\pm$ 0.0135} & \tblstack{0.2531}{$\pm$ 0.0123} & \tblstack{0.3782}{$\pm$ 0.0018} & \tblstack{0.3316}{$\pm$ 0.0015} \\
RealMLP & \tblstack{0.4142}{$\pm$ 0.0072} & \tblstack{0.4137}{$\pm$ 0.0043} & \tblstack{0.2667}{$\pm$ 0.0010} & \tblstack{0.2536}{$\pm$ 0.0005} & \tblstack{0.2883}{$\pm$ 0.0034} & \tblstack{0.2798}{$\pm$ 0.0028} & \tblstack{0.3750}{$\pm$ 0.0004} & \tblstack{0.3287}{$\pm$ 0.0003} \\
TabM & \tblstack{0.4152}{$\pm$ 0.0107} & \tblstack{0.4142}{$\pm$ 0.0050} & \tblstack{0.2666}{$\pm$ 0.0015} & \tblstack{0.2536}{$\pm$ 0.0007} & \tblstack{0.2784}{$\pm$ 0.0051} & \tblstack{0.2685}{$\pm$ 0.0067} & \tblstack{0.3763}{$\pm$ 0.0010} & \tblstack{0.3300}{$\pm$ 0.0008} \\
Joint TabR & \tblstack{0.3751}{$\pm$ 0.0078} & \tblstack{0.3810}{$\pm$ 0.0061} & \tblstack{0.2757}{$\pm$ 0.0007} & \tblstack{0.2599}{$\pm$ 0.0006} & \tblstack{0.2482}{$\pm$ 0.0101} & \tblstack{0.2409}{$\pm$ 0.0084} & \tblstack{0.3806}{$\pm$ 0.0012} & \tblstack{0.3336}{$\pm$ 0.0009} \\
Ridge & \tblstack{0.0833}{$\pm$ 0.0000} & \tblstack{0.0808}{$\pm$ 0.0000} & \tblstack{0.7616}{$\pm$ 0.0000} & \tblstack{0.7251}{$\pm$ 0.0000} & \tblstack{0.0583}{$\pm$ 0.0000} & \tblstack{0.0583}{$\pm$ 0.0000} & \tblstack{1.1990}{$\pm$ 0.0000} & \tblstack{1.0275}{$\pm$ 0.0000} \\
\bottomrule
\end{tabular}
\end{lrbox}
\AppendixTableOutput{generated_all10_response_metrics-1}
\end{table}

% Appendix presentation derived from ../generated_d2_process_evidence.tex
\begin{table}[!htbp]
\centering
\caption{\textbf{Deterministic compilation eliminates candidate-code execution failures in the evaluated path-constrained runs.} PCC and resource values are trajectory-level means $\pm$ 95\% interval half-width. Valid covers all candidate slots, repairs are candidate-code repair attempts per trajectory, and Linked is the number of candidates connected to the declared input in code. Provider retries remain included in calls and tokens.}
\label{tab:d2-process}
\tblsetup
\setlength{\tabcolsep}{3pt}
\begin{lrbox}{\AppendixTableBox}
\begin{tabular}{lrrrrrrr}
\toprule
\rowcolor{csHeader}
Task & \tblhdr{Score\\PCC} & \tblhdr{Path\\PCC} & $\Delta$ & \tblhdr{Tokens\\(K)} & Valid & Repairs & Linked \\
\midrule
BBBC036 & \tblstack{0.2924}{$\pm$ 0.0037} & \tblstack{0.2925}{$\pm$ 0.0019} & +0.0001 & \tblstack{21.9}{$\pm$ 1.4} & 100.0\% & \tblstack{0.00}{$\pm$ 0.00} & 100/100 \\
BBBC047 & \tblstack{0.3097}{$\pm$ 0.0119} & \tblstack{0.3008}{$\pm$ 0.0012} & -0.0090 & \tblstack{21.7}{$\pm$ 1.2} & 100.0\% & \tblstack{0.00}{$\pm$ 0.00} & 100/100 \\
\bottomrule
\end{tabular}
\end{lrbox}
\AppendixTableOutput{generated_d2_process_evidence-1}
\end{table}

\FloatBarrier
\section{Input-use claims in combinatorial genetic-perturbation prediction}
\label{app:norman-transfer}

% Appendix presentation derived from ../generated_main_norman_strip.tex
\begin{table}[!htbp]
\centering
\caption{\textbf{Registered pair-input dependence on unseen genetic combinations.} Five refits per predictor; all pass the input-use test. Anchored predictors replace the pair-input bundle; GEARS uses its native pair-identity replacement. Drops are reported without cross-interface ranking. Complete comparisons: Appendix~\ref{app:norman-transfer}.}
\label{tab:main-norman}
\tblsetup
\renewcommand{\arraystretch}{1.0}
\setlength{\tabcolsep}{2.5pt}
\begin{lrbox}{\AppendixTableBox}
\begin{tabular}{@{}L{63.40pt}L{86.19pt}R{104.02pt}R{78.26pt}C{40.62pt}@{}}
\toprule
\rowcolor{csHeader}
\textbf{Boundary} & \textbf{Predictor} & \tblhdr{Global PCC $\uparrow$} & \tblhdr{Input $\Delta$PCC $\uparrow$} & \tblhdr{Support} \\
\midrule
\multirow{3}{*}{Fold 4 audit} & GEARS & \tblci{0.6819}{0.6714, 0.6924} & 0.3029 & 5/5 \\
 & Initial model ($h_0$) & \tblci{0.8960}{0.8909, 0.9011} & 0.5425 & 5/5 \\
\cellcolor{white} & \discoverymethod & \tblci{0.9020}{0.8988, 0.9051} & 0.5578 & 5/5 \\
\addlinespace[2pt]
\multirow{3}{*}{Fold 5 replication} & GEARS & \tblci{0.6676}{0.6568, 0.6784} & 0.3599 & 5/5 \\
 & Initial model ($h_0$) & \tblci{0.9051}{0.9023, 0.9079} & 0.6193 & 5/5 \\
\cellcolor{white} & \discoverymethod & \tblci{0.9096}{0.9051, 0.9142} & 0.6280 & 5/5 \\
\bottomrule
\end{tabular}
\end{lrbox}
\AppendixTableOutput{generated_main_norman_strip-1}
\end{table}

The transfer task is reconstructed from the count layer of the Norman K562 CRISPRa Perturb-seq experiment (GEO GSE133344) \citep{norman2019genetic}. Its source matrix contains 91,205 cells and 5,045 genes, including 7,353 control cells. Per-cell counts are library-size normalized to 10,000 and log-transformed, after which canonical condition means are centered by the control mean. The 284 source labels resolve to 237 canonical conditions: 105 observed singles, 131 doubles, and one control. The common discovery interface receives a pair-input bundle containing a 105-dimensional multi-hot perturbation identity and its matching mean single-perturbation anchor, and returns the full 5,045-gene response without external pathway features or cell-type covariates. GEARS retains its task-native graph inputs as described below.

The five-fold assignment is constructed without expression responses and balances pair count, source-cell count, and perturbation-gene incidence. All singles and controls remain fit-only with the Fold-1/2 doubles; Fold 3 supplies discovery feedback, and Folds 4 and 5 are successive held-out evaluation folds. Table~\ref{tab:norman-fold-statistics} lists every held-out double-perturbation block. The comparison predicts unseen double combinations whose constituent single-gene responses are available during fitting.

Each held-out boundary contains 26 unseen double-perturbation conditions. Before Fold 4 is opened, we construct 32 distinct complete derangements of the 26 row positions; each is a one-to-one permutation with no fixed point and is reused positionally on Fold 5. For each source condition, the registered intervention replaces both coordinates of the pair-input bundle (its 105-dimensional pair-identity vector and matching mean single-perturbation anchor) with the donor pair's values while retaining the source response target. Replacing both coordinates prevents a residual predictor from retaining pair information through the anchor; consequently, the test measures dependence on the complete bundle rather than attributing the effect to the multi-hot coordinate alone. The maps use neither responses nor model outputs and do not require donor pairs to be gene-disjoint from source pairs.

For refit $j$ and map $r$, let $L_j$ be joint-output MSE under the correct pair-input bundle and $L^{\mathrm{shuf}}_{j,r}$ the MSE after the bundle shuffle. We define $g_{j,r}=L^{\mathrm{shuf}}_{j,r}-L_j$ and $\tilde g_{j,r}=g_{j,r}/\max(L_j,10^{-12})$. A refit is qualified under the pair-input-use criterion when $32^{-1}\sum_r\tilde g_{j,r}>0.001$ and $Q_{0.05}(\{g_{j,r}\}_{r=1}^{32})>0$. It is unsupported when the first quantity lies within $[-0.001,0.001]$ and inconclusive otherwise. The Global-\pcc drop is reported as a continuous effect but does not enter this categorical rule. The 32 maps are repeated interventions within a refit; the five refits are the statistical units for model-performance intervals and support counts.

GEARS is evaluated through the shared task boundary while retaining its original multigene-perturbation formulation \citep{roohani2024gears}: GO and coexpression graphs, native GNN/decoder, optimization objective, and validation-selected checkpoint. It shares count normalization, response target, fold roles, and held-out metrics with the other predictors. Table~\ref{tab:norman-gears-protocol} records the architecture, graph scope, optimizer, and checkpoint rule used in the comparison.

% Appendix presentation derived from ../generated_norman_protocol_statistics.tex
\begin{table}[!htbp]
\centering
\caption{\textbf{Norman combinatorial-perturbation fold inventory.} Source cells are counted before condition-level aggregation. All 105 observed single perturbations and the control condition remain fit-only; the table reports the held-out double-perturbation partition.}
\label{tab:norman-fold-statistics}
\tblsetup
\setlength{\tabcolsep}{3pt}
\begin{lrbox}{\AppendixTableBox}
\begin{tabular}{clrrr}
\toprule
\rowcolor{csHeader}
Fold & Role & \tblhdr{Double\\perturbations} & \tblhdr{Source\\cells} & \tblhdr{Component\\genes} \\
\midrule
1 & Fit & 26 & 7,301 & 40 \\
2 & Fit & 26 & 7,849 & 41 \\
3 & Discovery/selection & 27 & 6,817 & 42 \\
4 & Independent audit & 26 & 6,541 & 40 \\
5 & Final replication & 26 & 6,937 & 38 \\
\bottomrule
\end{tabular}
\end{lrbox}
\AppendixTableOutput{generated_norman_protocol_statistics-1}
\end{table}

\begin{table}[!htbp]
\centering
\caption{\textbf{GEARS alignment for the Norman transfer.} The original GEARS architecture and graph construction are retained, while normalization, response target, data-split roles, and held-out evaluator match the common 5,045-gene task.}
\label{tab:norman-gears-protocol}
\tblsetup
\setlength{\tabcolsep}{3pt}
\begin{tabularx}{\linewidth}{L{77pt}X}
\toprule
\rowcolor{csHeader}
Protocol item & Setting \\
\midrule
\textbf{Input and output} & 105 observed perturbation genes; one joint 5,045-gene response \\
\textbf{Partition roles} & Folds 1--2 fit; Fold 3 selects checkpoints; Folds 4/5 are audit/replication \\
\textbf{Development graph} & Fit and Fold-3 conditions only; 102/105 genes GO-annotated; three isolated genes receive self-loops; 2,043 GO edges \\
\textbf{Architecture} & Hidden size 64; one GO GNN layer; one gene GNN layer; decoder hidden size 16; 20 GO and 20 coexpression neighbors; coexpression threshold 0.4 \\
\textbf{Optimization} & 20 epochs; batch 128; Adam, learning rate $10^{-3}$, weight decay $5\times10^{-4}$; StepLR$(1,0.5)$; five seeds (41--45) \\
\textbf{Checkpoint rule} & Minimum Fold-3 DEG-20 validation MSE; frozen before either held-out boundary is opened \\
\textbf{Input-use test} & 32 complete pair-identity derangements fixed without response values; pass requires mean relative MSE increase $>0.001$ and positive 5th-percentile raw MSE increase \\
\bottomrule
\end{tabularx}
\end{table}

Across ten trajectories, all 77 nonduplicate CellScientist candidates pass source, shape, and training checks; the 23 duplicate design cards still consume their candidate slots, and no Fold-4 or Fold-5 response enters proposal or selection. The trajectories account for 154 provider calls, 403,762 tokens, 17.8 H100 fit-minutes, and 50.5 minutes of end-to-end time. Held-out input-use testing is deterministic and uses zero model-provider calls.

% Appendix presentation derived from ../generated_norman_rna_transfer_appendix.tex
\begin{table}[!htbp]
\centering
\caption{\textbf{Search statistics for the Norman transfer.} Results use ten independent trajectories; the candidate budget includes $h_0$.}
\label{tab:norman-discovery-receipt}
\tblsetup
\setlength{\tabcolsep}{3pt}
\begin{lrbox}{\AppendixTableBox}
\begin{tabular}{rrrrrr}
\multicolumn{6}{@{}l}{\textbf{A. Search performance}} \\
\addlinespace[2pt]
\toprule
\rowcolor{csHeader}
$B$ & Best@3 & Best@5 & Best@10 & Frontier AUC & $\Delta h_0$ [95\% CI] \\
\midrule
10 & 0.9192 & 0.9201 & 0.9210 & 0.9194 & \tblstack{+0.0062}{[+0.0036, +0.0087]} \\
\bottomrule
\end{tabular}
\end{lrbox}
\AppendixTableOutput{generated_norman_rna_transfer_appendix-1}
\par\vspace{6pt}
\begin{lrbox}{\AppendixTableBox}
\begin{tabular}{rrrr}
\multicolumn{4}{@{}l}{\textbf{B. Completion and resources}} \\
\addlinespace[2pt]
\toprule
\rowcolor{csHeader}
Positive & Executed & Calls & Tokens \\
\midrule
10/10 & 77/100 & 154 & 403,762 \\
\bottomrule
\end{tabular}
\end{lrbox}
\AppendixTableOutput{generated_norman_rna_transfer_appendix-2}
\end{table}

\begin{table}[!htbp]
\centering
\caption{\textbf{Complete held-out Norman response metrics.} Neural entries summarize five refits; deterministic references are evaluated once. All predictors return the same 5,045-gene joint response. Input-use support is the number of refits passing the predictor's registered perturbation-pair input test; deterministic references report their single result. Predictors with an observed-single anchor are tested by replacing the complete pair-input bundle, while GEARS retains its task-native pair-identity intervention.}
\label{tab:norman-complete-metrics}
\tblsetup
\setlength{\tabcolsep}{3pt}
% presentation-panels: 5
\setlength{\tabcolsep}{2.5pt}
\begin{lrbox}{\AppendixTableBox}
\begin{tabular}{@{}lrrrrrrrr@{}}
\toprule
\rowcolor{csHeader}
Predictor & \tblhdr{Global\\PCC $\uparrow$} & MSE $\downarrow$ & \tblhdr{Condition\\PCC $\uparrow$} & \tblhdr{Rank corr.\\$\uparrow$} & \tblhdr{DEG-20\\PCC $\uparrow$} & \tblhdr{DEG-50\\PCC $\uparrow$} & \tblhdr{Full-state\\PCC $\uparrow$} & \tblhdr{Input-use\\support} \\
\midrule

\rowcolor{csHeader}\multicolumn{9}{@{}l}{\textbf{Fold 4 audit}} \\
\addlinespace[1pt]
Matching mean & 0.8810 & 0.0059 & 0.8966 & 0.4902 & 0.9702 & 0.9634 & 0.9915 & supported \\
Additive singles & 0.8810 & 0.0046 & 0.8966 & 0.4902 & 0.9702 & 0.9634 & 0.9932 & supported \\
Ridge & 0.8989 & 0.0033 & 0.8965 & 0.4584 & 0.9724 & 0.9665 & 0.9951 & supported \\
GEARS & 0.6819 & 0.0122 & 0.6826 & 0.2512 & 0.9078 & 0.8915 & 0.9843 & 5/5 \\
$h_0$ & 0.8960 & 0.0037 & 0.8999 & 0.4288 & 0.9664 & 0.9618 & 0.9945 & 5/5 \\
CellScientist & 0.9020 & 0.0039 & 0.9098 & 0.4345 & 0.9750 & 0.9697 & 0.9943 & 5/5 \\
\midrule
\rowcolor{csHeader}\multicolumn{9}{@{}l}{\textbf{Fold 5 replication}} \\
\addlinespace[1pt]
Matching mean & 0.8982 & 0.0043 & 0.8799 & 0.4681 & 0.9614 & 0.9469 & 0.9938 & supported \\
Additive singles & 0.8982 & 0.0033 & 0.8799 & 0.4681 & 0.9614 & 0.9469 & 0.9952 & supported \\
Ridge & 0.8959 & 0.0029 & 0.8813 & 0.4375 & 0.9672 & 0.9546 & 0.9957 & supported \\
GEARS & 0.6676 & 0.0115 & 0.6156 & 0.2267 & 0.8021 & 0.8096 & 0.9853 & 5/5 \\
$h_0$ & 0.9051 & 0.0027 & 0.8844 & 0.4065 & 0.9542 & 0.9474 & 0.9961 & 5/5 \\
CellScientist & 0.9096 & 0.0028 & 0.8944 & 0.4071 & 0.9650 & 0.9554 & 0.9960 & 5/5 \\
\bottomrule
\end{tabular}
\end{lrbox}
\AppendixTableOutput{generated_norman_rna_transfer_appendix-3}
\end{table}

The matched transfer analysis gives CellScientist, AIDE, CellForge, and HarmonyCell the same initial model, design language, Fold-3 feedback, physical budget, and model-selection rule; each policy generates its own candidates. CellScientist has the highest mean Best@10 and frontier AUC. Its paired Best@10 effect is $+0.0029$ versus AIDE (95\% CI $[-0.0003,+0.0061]$), $+0.0058$ versus CellForge ($[+0.0037,+0.0080]$), and $+0.0006$ versus HarmonyCell ($[-0.0005,+0.0016]$). After selection, all sources are refit under five paired seeds and audited without provider calls. Global-PCC ordering varies across folds, while all four selected models pass the pair-input-bundle test in 5/5 refits; CellScientist leads condition-, DEG-20-, and DEG-50-PCC at Fold 5.

% Appendix presentation derived from ../generated_norman_agent_policy_appendix.tex
\begin{table}[!htbp]
\centering
\caption{\textbf{Matched agentic model discovery on the Norman transfer.} Each method receives the same initial model, trainer, Fold-3 feedback, ten trajectories, and ten candidate evaluations. AIDE, CellForge, and HarmonyCell are implemented within the same runnable model space. Arrows mark the preferred direction. Executed slots count candidate proposals that trained; duplicate design cards consume slots but are not executable failures.}
\label{tab:norman-agent-discovery}
\tblsetup
\setlength{\tabcolsep}{3pt}
% presentation-panels: 6
\setlength{\tabcolsep}{2.5pt}
\begin{lrbox}{\AppendixTableBox}
\begin{tabular}{@{}lrrrrrrrrr@{}}
\toprule
\rowcolor{csHeader}
Policy & Best@3 $\uparrow$ & Best@5 $\uparrow$ & Best@10 $\uparrow$ & \tblhdr{Frontier\\AUC $\uparrow$} & $\Delta h_0$ [95\% CI] $\uparrow$ & \tblhdr{Positive\\$\uparrow$} & \tblhdr{Executed\\slots $\uparrow$} & Calls $\downarrow$ & Tokens $\downarrow$ \\
\midrule

CellScientist & \textbf{0.9192} & \secondbest{0.9201} & \textbf{0.9210} & \textbf{0.9194} & \tblstack{\textbf{+0.0062}}{[+0.0036,+0.0087]} & 10/10 & 77/100 & 154 & 403,762 \\
AIDE & 0.9164 & 0.9168 & 0.9181 & 0.9170 & \tblstack{+0.0033}{[+0.0011,+0.0054]} & 9/10 & 91/100 & 120 & 335,610 \\
CellForge & 0.9151 & 0.9152 & 0.9152 & 0.9151 & \tblstack{+0.0003}{[$-$0.0004,+0.0010]} & 1/10 & 97/100 & 108 & 309,742 \\
HarmonyCell & \secondbest{0.9191} & \textbf{0.9203} & \secondbest{0.9204} & \secondbest{0.9193} & \tblstack{\secondbest{+0.0056}}{[+0.0028,+0.0084]} & 9/10 & 88/100 & 117 & 328,705 \\
\bottomrule
\end{tabular}
\end{lrbox}
\AppendixTableOutput{generated_norman_agent_policy_appendix-1}
\tblnote{Bold/underline: best/second distinct displayed predictive mean within each task or fold. Ties share a rank; interval bounds and diagnostic/resource columns are unranked.}
\end{table}

\begin{table}[!htbp]
\centering
\caption{\textbf{Models selected on Fold 3 before audit.} One model per search method is chosen by the common rule: highest Fold-3 Global PCC, then lower MSE and fewer parameters.}
\label{tab:norman-agent-endpoints}
\tblsetup
\setlength{\tabcolsep}{3pt}
\begin{lrbox}{\AppendixTableBox}
\begin{tabular}{llrrrr}
\multicolumn{6}{@{}l}{\textbf{A. Selected architecture}} \\
\addlinespace[2pt]
\toprule
\rowcolor{csHeader}
Policy & Architecture & Width & Depth & Dropout & \tblhdr{PCC--MSE\\weight} \\
\midrule
 CellScientist & Gated MLP & 512 & 4 & 0.1 & 0.3 \\
AIDE & Gated MLP & 512 & 4 & 0.2 & 0.2 \\
CellForge & Residual MLP & 256 & 3 & 0.1 & 0.1 \\
HarmonyCell & Residual MLP & 384 & 4 & 0.2 & 0.1 \\
\bottomrule
\end{tabular}
\end{lrbox}
\AppendixTableOutput{generated_norman_agent_policy_appendix-2}
\par\vspace{6pt}
\begin{lrbox}{\AppendixTableBox}
\begin{tabular}{llrr}
\multicolumn{4}{@{}l}{\textbf{B. Optimization and selection}} \\
\addlinespace[2pt]
\toprule
\rowcolor{csHeader}
Policy & \tblhdr{Optimizer /\\learning rate} & Parameters & Fold-3 PCC \\
\midrule
 CellScientist & AdamW / 0.001 & 5,850,037 & 0.9237 \\
AIDE & AdamW / 0.0005 & 5,850,037 & 0.9234 \\
CellForge & AdamW / 0.001 & 1,850,549 & 0.9195 \\
HarmonyCell & AdamW / 0.001 & 3,758,261 & 0.9247 \\
\bottomrule
\end{tabular}
\end{lrbox}
\AppendixTableOutput{generated_norman_agent_policy_appendix-3}
\end{table}

\begin{table}[!htbp]
\centering
\caption{\textbf{Matched input-use test and final replication of selected models.} Neural models are refit with five paired seeds. Brackets are 95\% intervals over those seeds; bundle drop averages 32 fixed interventions that jointly replace the pair identity and matching observed-single anchor. All sources are selected before Fold 4 and are refit at Fold 5 without source reselection.}
\label{tab:norman-agent-heldout}
\tblsetup
\setlength{\tabcolsep}{3pt}
\begin{lrbox}{\AppendixTableBox}
\begin{tabular}{lrrrrrrc}
\toprule
\rowcolor{csHeader}
Policy & \tblhdr{Global PCC $\uparrow$\\{[95\% CI]}} & MSE $\downarrow$ & \tblhdr{Condition\\PCC $\uparrow$} & \tblhdr{DEG-20\\PCC $\uparrow$} & \tblhdr{DEG-50\\PCC $\uparrow$} & \tblhdr{Bundle\\drop} & \tblhdr{Bundle-use\\support} \\
\midrule
\rowcolor{csHeader}\multicolumn{8}{@{}l}{\textbf{Fold 4 audit}} \\
\addlinespace[1pt]
 CellScientist & \tblstack{\textbf{0.9020}}{[0.8988,0.9051]} & 0.00388 & \textbf{0.9098} & \textbf{0.9750} & \textbf{0.9697} & 0.5578 & 5/5 \\
AIDE & \tblstack{\secondbest{0.8981}}{[0.8920,0.9042]} & 0.00407 & \secondbest{0.9038} & 0.9682 & 0.9637 & 0.5458 & 5/5 \\
CellForge & \tblstack{0.8960}{[0.8909,0.9011]} & \secondbest{0.00372} & 0.8999 & 0.9664 & 0.9618 & 0.5425 & 5/5 \\
HarmonyCell & \tblstack{0.8960}{[0.8920,0.9000]} & \textbf{0.00369} & 0.9019 & \secondbest{0.9693} & \secondbest{0.9641} & 0.5506 & 5/5 \\
\midrule
\rowcolor{csHeader}\multicolumn{8}{@{}l}{\textbf{Fold 5 replication}} \\
\addlinespace[1pt]
 CellScientist & \tblstack{\secondbest{0.9096}}{[0.9051,0.9142]} & 0.00277 & \textbf{0.8944} & \textbf{0.9650} & \textbf{0.9554} & 0.6280 & 5/5 \\
AIDE & \tblstack{\textbf{0.9125}}{[0.9088,0.9161]} & 0.00285 & \secondbest{0.8912} & 0.9567 & 0.9488 & 0.6270 & 5/5 \\
CellForge & \tblstack{0.9051}{[0.9023,0.9079]} & \secondbest{0.00267} & 0.8844 & 0.9542 & 0.9474 & 0.6193 & 5/5 \\
HarmonyCell & \tblstack{0.9088}{[0.9029,0.9148]} & \textbf{0.00255} & 0.8867 & \secondbest{0.9597} & \secondbest{0.9505} & 0.6294 & 5/5 \\
\bottomrule
\end{tabular}
\end{lrbox}
\AppendixTableOutput{generated_norman_agent_policy_appendix-4}
\tblnote{Bold/underline: best/second distinct displayed predictive mean within each task or fold. Ties share a rank; interval bounds and diagnostic/resource columns are unranked.}
\end{table}

\begin{table}[!htbp]
\centering
\caption{\textbf{Paired CellScientist contrasts in the Norman held-out study.} Effects are CellScientist minus the comparator over five paired fitting seeds. Global-PCC wins count seeds with a positive effect.}
\label{tab:norman-agent-contrasts}
\tblsetup
\setlength{\tabcolsep}{3pt}
\begin{lrbox}{\AppendixTableBox}
\begin{tabular}{lrrr}
\toprule
\rowcolor{csHeader}
Comparator & \tblhdr{$\Delta$ Global PCC\\{[95\% CI]}} & PCC wins & \tblhdr{$\Delta$ bundle drop\\{[95\% CI]}} \\
\midrule
\rowcolor{csHeader}\multicolumn{4}{@{}l}{\textbf{Fold 4 audit}} \\
\addlinespace[1pt]
AIDE & \tblstack{+0.0039}{[$-$0.0044,+0.0122]} & 4/5 & \tblstack{+0.0121}{[$-$0.0004,+0.0245]} \\
CellForge & \tblstack{+0.0060}{[$-$0.0008,+0.0128]} & 4/5 & \tblstack{+0.0154}{[+0.0030,+0.0277]} \\
HarmonyCell & \tblstack{+0.0060}{[$-$0.0009,+0.0129]} & 4/5 & \tblstack{+0.0072}{[$-$0.0061,+0.0205]} \\
\midrule
\rowcolor{csHeader}\multicolumn{4}{@{}l}{\textbf{Fold 5 replication}} \\
\addlinespace[1pt]
AIDE & \tblstack{$-$0.0028}{[$-$0.0103,+0.0046]} & 2/5 & \tblstack{+0.0010}{[$-$0.0059,+0.0078]} \\
CellForge & \tblstack{+0.0045}{[$-$0.0004,+0.0094]} & 5/5 & \tblstack{+0.0087}{[+0.0037,+0.0137]} \\
HarmonyCell & \tblstack{+0.0008}{[$-$0.0056,+0.0072]} & 4/5 & \tblstack{$-$0.0015}{[$-$0.0131,+0.0101]} \\
\bottomrule
\end{tabular}
\end{lrbox}
\AppendixTableOutput{generated_norman_agent_policy_appendix-5}
\end{table}

\FloatBarrier
\section{Independent-acquisition test of small-molecule claim generalization}
\label{app:external-transfer}

After the final falsification-guided procedure is fixed, LINCS--Pilot1 provides an out-of-development-cohort test. It uses an observed control profile, compound identity, and dose to predict a joint 2,719-dimensional morphology--transcription response across 4,155 matched conditions and 1,108 Murcko groups. The ten trajectory-selected design instances are then fixed for transfer to \texttt{cpg0004/LKCP/2017\_12\_05\_Batch2}, an independently acquired Cell Painting release \citep{weisbart2024gallery}. Its 134 files contain 11,904 rows; 8,958 treatment and 2,946 control rows yield 1,728 retained conditions and 64 compound-held-out groups. Response- and fold-blind interface quality control removes 110 non-finite or extreme release features from 1,781 common morphology columns, retaining a 1,671-dimensional same-plate control profile and response.

The scientific question and model designs precede LKCP transfer. The final dataset adapter is fixed after interface quality control and before LKCP training or held-out-fold access. The transferred designs, input-permutation construction, map seeds, threshold-construction rule, and decision rule are registered and hash-bound at that point. Each design is then refit under five paired seeds with no further source edit or model reselection. Compound permutations hold dose fixed; dose permutations use only observed doses of the same compound, require at least a $0.1$ log-dose contrast, and retain the same eligibility rule. Across $2$ cohorts $\times$ $2$ folds $\times$ $10$ designs $\times$ $5$ seeds, the analysis contains 200 cohort--fold--design--seed refit boundaries. Each boundary reuses the 32 registered ordered compound--dose permutation pairs, yielding 6,400 matched within-refit permutation units; these maps are repeated measurements rather than independent statistical units. For $U_{11}$ with both inputs correct, $U_{01}$ with compound shuffled, $U_{10}$ with dose shuffled, and $U_{00}$ with both shuffled, the two-player allocation is $\phi_{\mathrm{chem}}=\tfrac12[(U_{11}-U_{01})+(U_{10}-U_{00})]$ and $\phi_{\mathrm{dose}}=\tfrac12[(U_{11}-U_{10})+(U_{01}-U_{00})]$. These allocate the contrast $U_{11}-U_{00}$; adding the residual $U_{00}-U_{\mathrm{anchor}}$ gives Full$-$control-only to numerical precision. The component allocations thus refer to the fixed replacement baseline, while Full$-$control-only measures the separately fitted predictors' performance difference.

% Appendix presentation derived from ../generated_coordinate_decomposition.tex
\begin{table}[!htbp]
\centering
\caption{\textbf{Compound identity and dose yield distinct replacement effects.} LINCS evaluates the final small-molecule procedure, and LKCP transfers the ten trajectory-selected LINCS design instances to an independent acquisition. These instances contain six unique configurations and retain repeated selections as part of the trajectory-level policy distribution. Compound identity and dose are two-player Shapley allocations of the Global-PCC change from both inputs replaced to both correct. Adding the both-shuffled residual gives Full$-$control-only, the separate full-model versus control-only comparison. The interaction is reported separately because it is already allocated across the two Shapley terms. Intervals use the ten endpoint instances after averaging five paired refits within each instance.}
\label{tab:coordinate-decomposition}
\tblsetup
\setlength{\tabcolsep}{3pt}
\begin{lrbox}{\AppendixTableBox}
\begin{tabular}{L{128pt}R{118pt}R{118pt}}
\multicolumn{3}{@{}l}{\textbf{A. LINCS--Pilot1}} \\
\addlinespace[2pt]
\toprule
\rowcolor{csHeader}
Contrast & Fold 4 [95\% CI] & Fold 5 [95\% CI] \\
\midrule
Full$-$control-only & \tblstack{+0.0293}{[+0.0281, +0.0305]} & \tblstack{+0.0233}{[+0.0226, +0.0241]} \\
Compound identity & \tblstack{+0.0175}{[+0.0164, +0.0186]} & \tblstack{+0.0126}{[+0.0117, +0.0134]} \\
Dose & \tblstack{+0.0486}{[+0.0473, +0.0498]} & \tblstack{+0.0467}{[+0.0456, +0.0478]} \\
Both-shuffled residual & \tblstack{-0.0368}{[-0.0378, -0.0358]} & \tblstack{-0.0359}{[-0.0370, -0.0349]} \\
Interaction & \tblstack{+0.0050}{[+0.0033, +0.0067]} & \tblstack{+0.0042}{[+0.0027, +0.0058]} \\
\bottomrule
\end{tabular}
\end{lrbox}
\AppendixTableOutput{generated_coordinate_decomposition-1}
\par\vspace{6pt}
\begin{lrbox}{\AppendixTableBox}
\begin{tabular}{L{128pt}R{118pt}R{118pt}}
\multicolumn{3}{@{}l}{\textbf{B. LKCP Batch 2}} \\
\addlinespace[2pt]
\toprule
\rowcolor{csHeader}
Contrast & Fold 4 [95\% CI] & Fold 5 [95\% CI] \\
\midrule
Full$-$control-only & \tblstack{+0.0104}{[+0.0103, +0.0106]} & \tblstack{+0.0211}{[+0.0205, +0.0217]} \\
Compound identity & \tblstack{-0.00007000}{[-0.00028998, +0.00014998]} & \tblstack{+0.0033}{[+0.0027, +0.0039]} \\
Dose & \tblstack{+0.0237}{[+0.0228, +0.0247]} & \tblstack{+0.0341}{[+0.0331, +0.0350]} \\
Both-shuffled residual & \tblstack{-0.0132}{[-0.0140, -0.0124]} & \tblstack{-0.0163}{[-0.0170, -0.0157]} \\
Interaction & \tblstack{+0.00004049}{[-0.00001503, +0.00009601]} & \tblstack{+0.0002}{[+0.0001, +0.0003]} \\
\bottomrule
\end{tabular}
\end{lrbox}
\AppendixTableOutput{generated_coordinate_decomposition-2}
\end{table}

\begin{table}[!htbp]
\centering
\caption{\textbf{Dose exceeds the reference-relative criterion consistently; compound status varies by cohort and fold.} The criterion compares each refit's lower effect quantile with its fold-specific reference threshold. Counts summarize stability relative to that reference scale, separately from the sign of the mean contribution. The mean-effect column is the mean of the 32 map-level coordinate effects used in this threshold test and is distinct from the two-player Shapley allocations in the preceding table. The 50 refits at each boundary are nested within ten trajectory-selected endpoint instances; the endpoint instance is the unit for continuous-effect intervals.}
\label{tab:coordinate-threshold}
\tblsetup
\setlength{\tabcolsep}{3pt}
\begin{lrbox}{\AppendixTableBox}
\begin{tabular}{lrrrrrr}
\toprule
\rowcolor{csHeader}
Cohort & Fold & \tblhdr{Mean\\effect} & \tblhdr{Lower-effect\\quantile} & Threshold & \tblhdr{Reference\\margin} & \tblhdr{Refits\\above} \\
\midrule
\multicolumn{7}{@{}l}{\textbf{Chemical identity}} \\
\addlinespace[1pt]
LINCS--Pilot1 & 4 & 0.0200 & 0.0139 & 0.0112 & +0.0028 & 35/50 \\
LINCS--Pilot1 & 5 & 0.0147 & 0.0090 & 0.0100 & -0.0010 & 10/50 \\
LKCP Batch 2 & 4 & 0.00003872 & -0.0016 & 0.0303 & -0.0319 & 0/50 \\
LKCP Batch 2 & 5 & 0.0032 & 0.0013 & 0.0307 & -0.0294 & 0/50 \\
\midrule
\multicolumn{7}{@{}l}{\textbf{Dose}} \\
\addlinespace[1pt]
LINCS--Pilot1 & 4 & 0.0510 & 0.0467 & 0.0068 & +0.0398 & 50/50 \\
LINCS--Pilot1 & 5 & 0.0488 & 0.0452 & 0.0059 & +0.0393 & 50/50 \\
LKCP Batch 2 & 4 & 0.0238 & 0.0205 & 0.0060 & +0.0145 & 50/50 \\
LKCP Batch 2 & 5 & 0.0342 & 0.0267 & 0.0125 & +0.0142 & 50/50 \\
\bottomrule
\end{tabular}
\end{lrbox}
\AppendixTableOutput{generated_coordinate_decomposition-3}
\end{table}

\FloatBarrier
\section{Full selected-model audit panel and calibration}
\label{app:audit}

The full Fold-4 panel contains the Fold-3-selected source from each of the four agents together with three distinct diagnostic sources: control-only, compound-only, and the shared $h_0$/simple-concatenation source shown under both reference roles. Predictive estimates use a separate five-seed refit panel from Table~\ref{tab:all10-utility}; model source or design, data fingerprint, population, and every non-seed fitting setting are identical. Each within-policy source is fixed before Fold 4; the pre-specified across-policy score rule then uses mean Fold-4 Global \pcc to choose the single prediction-score source carried to Fold 5, which includes its paired control-only comparator. Controlled fixtures check recognized input consumption, the one-key attention contradiction, missing declarations, abstention, and known input dependence. A fixed compound-strength sweep characterizes threshold crossing in those fixtures; Appendix~\ref{app:factorial-definition} explains the criterion's reference-relative interpretation.

\begin{figure}[!htbp]
    \centering
    \input{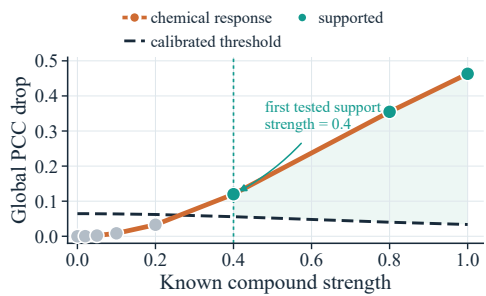}
    \caption{\textbf{Controlled checks of input-use measurements.} Null, control-only, perturbation-only, and gated-interaction fixtures produce distinct effects; the strength sweep locates threshold crossing in these fixtures. Source fixtures in Table~\ref{tab:checker-calibration} check implemented, implementation-contradicted, and unresolved source statuses.}
    \label{fig:audit-calibration}
\end{figure}

% Appendix presentation derived from ../generated_a1_fold4_audit.tex
\begin{table}[!htbp]
\centering
\caption{Post-search Fold-4 code and behavioral audit, held out from discovery. Global, CP, and L1000 PCC describe each selected model, while pre-specified input permutations test whether the trained predictor uses compound identity. A separate paired five-seed panel refits every model under the same training and evaluation setup used in Table~\ref{tab:all10-utility}. Values are mean $\pm$ 95\% $t$-interval half-width over five refits. Each seed-level mean aggregates 32 fixed permutations, and the compound-identity test uses the pre-specified threshold rule. Q/U/I gives the qualified/unsupported/inconclusive seed counts; aggregate status applies the registered five-seed rule, with an inconclusive aggregate when no category reaches four of five refits. The across-policy prediction-score selection for Fold 5 is performed only after this panel is complete.}
\label{tab:a1-fold4}
\tblsetup
\setlength{\tabcolsep}{3pt}
\begin{lrbox}{\AppendixTableBox}
\begin{tabular}{lrrrr}
\multicolumn{5}{@{}l}{\textbf{A. Predictive performance}} \\
\addlinespace[2pt]
\toprule
\rowcolor{csHeader}
Model & Global PCC & MSE & CP PCC & L1000 PCC \\
\midrule
\rowcolor{csHeader}\multicolumn{5}{@{}l}{\textbf{BBBC036}} \\
\addlinespace[1pt]
 \discoverymethod & \tblpm{0.3096}{0.0029} & \tblpm{0.0586}{0.0001} & \tblpm{0.3113}{0.0014} & \tblpm{0.3089}{0.0041} \\
\textsc{AIDE} & \tblpm{0.2970}{0.0075} & \tblpm{0.0593}{0.0004} & \tblpm{0.3052}{0.0031} & \tblpm{0.2944}{0.0102} \\
\textsc{CellForge} & \tblpm{0.2960}{0.0036} & \tblpm{0.0592}{0.0002} & \tblpm{0.3056}{0.0043} & \tblpm{0.2925}{0.0037} \\
\textsc{HarmonyCell} & \tblpm{0.3110}{0.0016} & \tblpm{0.0586}{0.0001} & \tblpm{0.3109}{0.0020} & \tblpm{0.3110}{0.0017} \\
$h_0$ & \tblpm{0.2879}{0.0040} & \tblpm{0.0598}{0.0004} & \tblpm{0.2981}{0.0016} & \tblpm{0.2839}{0.0056} \\
Control-only & \tblpm{0.3098}{0.0024} & \tblpm{0.0586}{0.0001} & \tblpm{0.3094}{0.0013} & \tblpm{0.3099}{0.0028} \\
Chemical-only & \tblpm{0.2239}{0.0002} & \tblpm{0.0616}{0.0000} & \tblpm{0.2998}{0.0003} & \tblpm{0.1832}{0.0002} \\
Simple concat & \tblpm{0.2879}{0.0040} & \tblpm{0.0598}{0.0004} & \tblpm{0.2981}{0.0016} & \tblpm{0.2839}{0.0056} \\
\midrule
\rowcolor{csHeader}\multicolumn{5}{@{}l}{\textbf{BBBC047}} \\
\addlinespace[1pt]
 \discoverymethod & \tblpm{0.3035}{0.0008} & \tblpm{0.0457}{0.0000} & \tblpm{0.3260}{0.0016} & \tblpm{0.2784}{0.0017} \\
\textsc{AIDE} & \tblpm{0.2697}{0.0048} & \tblpm{0.0469}{0.0002} & \tblpm{0.3148}{0.0047} & \tblpm{0.2152}{0.0116} \\
\textsc{CellForge} & \tblpm{0.2727}{0.0019} & \tblpm{0.0467}{0.0001} & \tblpm{0.3175}{0.0035} & \tblpm{0.2190}{0.0059} \\
\textsc{HarmonyCell} & \tblpm{0.2983}{0.0110} & \tblpm{0.0459}{0.0003} & \tblpm{0.3271}{0.0030} & \tblpm{0.2654}{0.0231} \\
$h_0$ & \tblpm{0.2697}{0.0048} & \tblpm{0.0469}{0.0002} & \tblpm{0.3148}{0.0047} & \tblpm{0.2152}{0.0116} \\
Control-only & \tblpm{0.3019}{0.0015} & \tblpm{0.0458}{0.0001} & \tblpm{0.3223}{0.0011} & \tblpm{0.2789}{0.0023} \\
Chemical-only & \tblpm{0.2159}{0.0012} & \tblpm{0.0480}{0.0000} & \tblpm{0.2935}{0.0022} & \tblpm{0.0924}{0.0028} \\
Simple concat & \tblpm{0.2697}{0.0048} & \tblpm{0.0469}{0.0002} & \tblpm{0.3148}{0.0047} & \tblpm{0.2152}{0.0116} \\
\bottomrule
\end{tabular}
\end{lrbox}
\AppendixTableOutput{generated_a1_fold4_audit-1}
\par\vspace{6pt}
\begin{lrbox}{\AppendixTableBox}
\begin{tabular}{lrrrcc}
\multicolumn{6}{@{}l}{\textbf{B. Input-use evidence}} \\
\addlinespace[2pt]
\toprule
\rowcolor{csHeader}
Model & $E_{\rm chem}$ & $E_{\rm control}$ & $I_{\rm chem,control}$ & Q/U/I & Status \\
\midrule
\rowcolor{csHeader}\multicolumn{6}{@{}l}{\textbf{BBBC036}} \\
\addlinespace[1pt]
 \discoverymethod & \tblpm{0.0000}{0.0000} & \tblpm{0.1622}{0.0095} & \tblpm{0.0000}{0.0000} & 0/5/0 & Unsup. \\
\textsc{AIDE} & \tblpm{0.0029}{0.0017} & \tblpm{0.1577}{0.0118} & \tblpm{-0.0021}{0.0016} & 0/5/0 & Unsup. \\
\textsc{CellForge} & \tblpm{0.0089}{0.0052} & \tblpm{0.1464}{0.0063} & \tblpm{-0.0000}{0.0007} & 0/5/0 & Unsup. \\
\textsc{HarmonyCell} & \tblpm{0.0000}{0.0000} & \tblpm{0.1693}{0.0033} & \tblpm{-0.0000}{0.0000} & 0/5/0 & Unsup. \\
$h_0$ & \tblpm{0.0193}{0.0087} & \tblpm{0.1359}{0.0057} & \tblpm{-0.0003}{0.0007} & 1/2/2 & Inc. \\
Control-only & \tblpm{0.0000}{0.0000} & \tblpm{0.1719}{0.0067} & \tblpm{0.0000}{0.0000} & -- & N/C \\
Chemical-only & \tblpm{0.0011}{0.0002} & \tblpm{0.0000}{0.0000} & \tblpm{0.0000}{0.0000} & 0/0/5 & Inc. \\
Simple concat & \tblpm{0.0193}{0.0087} & \tblpm{0.1359}{0.0057} & \tblpm{-0.0003}{0.0007} & 1/2/2 & Inc. \\
\midrule
\rowcolor{csHeader}\multicolumn{6}{@{}l}{\textbf{BBBC047}} \\
\addlinespace[1pt]
 \discoverymethod & \tblpm{0.0000}{0.0000} & \tblpm{0.1932}{0.0044} & \tblpm{0.0000}{0.0000} & 0/5/0 & Unsup. \\
\textsc{AIDE} & \tblpm{0.0118}{0.0052} & \tblpm{0.1119}{0.0121} & \tblpm{0.0013}{0.0009} & 0/1/4 & Inc. \\
\textsc{CellForge} & \tblpm{0.0100}{0.0045} & \tblpm{0.1114}{0.0059} & \tblpm{0.0012}{0.0008} & 0/2/3 & Inc. \\
\textsc{HarmonyCell} & \tblpm{0.0006}{0.0010} & \tblpm{0.1866}{0.0225} & \tblpm{0.0009}{0.0015} & 0/5/0 & Unsup. \\
$h_0$ & \tblpm{0.0118}{0.0052} & \tblpm{0.1119}{0.0121} & \tblpm{0.0013}{0.0009} & 0/1/4 & Inc. \\
Control-only & \tblpm{0.0000}{0.0000} & \tblpm{0.2030}{0.0054} & \tblpm{0.0000}{0.0000} & -- & N/C \\
Chemical-only & \tblpm{0.0111}{0.0010} & \tblpm{0.0000}{0.0000} & \tblpm{0.0000}{0.0000} & 5/0/0 & Sup. \\
Simple concat & \tblpm{0.0118}{0.0052} & \tblpm{0.1119}{0.0121} & \tblpm{0.0013}{0.0009} & 0/1/4 & Inc. \\
\bottomrule
\end{tabular}
\end{lrbox}
\AppendixTableOutput{generated_a1_fold4_audit-2}
\tblnote{Aggregate status: Sup., claim-supported; Unsup., unsupported; Inc., inconclusive; N/C, not claimed.}
\end{table}

% Appendix presentation derived from ../generated_a1_fold5_replication.tex
\begin{table}[!htbp]
\centering
\caption{Fold-5 evaluation after automatic model selection on Fold 4. Prediction-score selection carries forward the source with the highest mean Fold-4 Global PCC; none of the four agent-selected sources meets the pre-specified aggregate compound-use criterion on either task. Values are mean $\pm$ 95\% $t$-interval half-width over five paired Fold-5 refits of the frozen source under the same seed schedule. Q/U/I gives the qualified/unsupported/inconclusive seed counts; aggregate status applies the registered five-seed rule, with an inconclusive aggregate when no category reaches four of five refits. ``Status replicated'' records whether this aggregate status agrees between Fold 4 and Fold 5.}
\label{tab:a1-fold5}
\tblsetup
\setlength{\tabcolsep}{3pt}
\begin{lrbox}{\AppendixTableBox}
\begin{tabular}{llrrrr}
\multicolumn{6}{@{}l}{\textbf{A. Predictive performance}} \\
\addlinespace[2pt]
\toprule
\rowcolor{csHeader}
Policy & Model & \tblhdr{Global\\PCC $\uparrow$} & MSE $\downarrow$ & \tblhdr{CP\\PCC $\uparrow$} & \tblhdr{L1000\\PCC $\uparrow$} \\
\midrule
\rowcolor{csHeader}\multicolumn{6}{@{}l}{\textbf{BBBC036}} \\
\addlinespace[1pt]
$h_0$ reference & $h_0$ & \tblstack{0.2588}{$\pm$ 0.0059} & \tblstack{0.0608}{$\pm$ 0.0004} & \tblstack{0.2596}{$\pm$ 0.0100} & \tblstack{0.2586}{$\pm$ 0.0094} \\
 Score-selected & \textsc{HarmonyCell} & \tblstack{0.2932}{$\pm$ 0.0020} & \tblstack{0.0590}{$\pm$ 0.0001} & \tblstack{0.2909}{$\pm$ 0.0018} & \tblstack{0.2942}{$\pm$ 0.0023} \\
\midrule
\rowcolor{csHeader}\multicolumn{6}{@{}l}{\textbf{BBBC047}} \\
\addlinespace[1pt]
$h_0$ reference & $h_0$ & \tblstack{0.2786}{$\pm$ 0.0031} & \tblstack{0.0465}{$\pm$ 0.0003} & \tblstack{0.3220}{$\pm$ 0.0091} & \tblstack{0.2243}{$\pm$ 0.0105} \\
 Score-selected & \discoverymethod & \tblstack{0.3153}{$\pm$ 0.0014} & \tblstack{0.0452}{$\pm$ 0.0000} & \tblstack{0.3393}{$\pm$ 0.0007} & \tblstack{0.2872}{$\pm$ 0.0025} \\
\bottomrule
\end{tabular}
\end{lrbox}
\AppendixTableOutput{generated_a1_fold5_replication-1}
\par\vspace{6pt}
\begin{lrbox}{\AppendixTableBox}
\begin{tabular}{lrrrccc}
\multicolumn{7}{@{}l}{\textbf{B. Input-use evidence and replication}} \\
\addlinespace[2pt]
\toprule
\rowcolor{csHeader}
Model & $E_{\rm chem}$ & $E_{\rm control}$ & $I_{\rm chem,control}$ & Q/U/I & Status & \tblhdr{Status\\replicated?} \\
\midrule
\rowcolor{csHeader}\multicolumn{7}{@{}l}{\textbf{BBBC036}} \\
\addlinespace[1pt]
$h_0$ & \tblstack{0.0047}{$\pm$ 0.0031} & \tblstack{0.1202}{$\pm$ 0.0085} & \tblstack{0.0042}{$\pm$ 0.0011} & 0/5/0 & Unsup. & no \\
 \textsc{HarmonyCell} & \tblstack{0.0000}{$\pm$ 0.0000} & \tblstack{0.1516}{$\pm$ 0.0035} & \tblstack{-0.0000}{$\pm$ 0.0000} & 0/5/0 & Unsup. & yes \\
\midrule
\rowcolor{csHeader}\multicolumn{7}{@{}l}{\textbf{BBBC047}} \\
\addlinespace[1pt]
$h_0$ & \tblstack{0.0096}{$\pm$ 0.0037} & \tblstack{0.1199}{$\pm$ 0.0109} & \tblstack{0.0012}{$\pm$ 0.0007} & 0/1/4 & Inc. & yes \\
 \discoverymethod & \tblstack{0.0000}{$\pm$ 0.0000} & \tblstack{0.2004}{$\pm$ 0.0061} & \tblstack{0.0000}{$\pm$ 0.0000} & 0/5/0 & Unsup. & yes \\
\bottomrule
\end{tabular}
\end{lrbox}
\AppendixTableOutput{generated_a1_fold5_replication-2}
\tblnote{Score-selected denotes prediction-score selection. Aggregate status: Unsup., unsupported; Inc., inconclusive.}
\end{table}

% Appendix presentation derived from ../generated_control_only_replication.tex
\begin{table}[!htbp]
\centering
\caption{Fold-5 comparison with a control-only reference. The score-selected model and control-only predictor are refit with the same five seeds and evaluated without reselection. Values are mean $\pm$ 95\% $t$-interval half-width; differences are paired across seeds.}
\label{tab:control-only-final}
\tblsetup
\setlength{\tabcolsep}{3pt}
\begin{lrbox}{\AppendixTableBox}
\begin{tabular}{llrrr}
\toprule
\rowcolor{csHeader}
Task & Score-selected model & \tblhdr{Selected-model\\PCC $\uparrow$} & \tblhdr{Control-only\\PCC $\uparrow$} & \tblhdr{Paired $\Delta$\\{[95\% CI]}} \\
\midrule
BBBC036 & \textsc{HarmonyCell} & \tblstack{0.2932}{$\pm$ 0.0020} & \tblstack{0.2921}{$\pm$ 0.0017} & \tblstack{+0.0011}{[$-$0.0008, +0.0030]} \\
 BBBC047 & \discoverymethod & \tblstack{0.3153}{$\pm$ 0.0014} & \tblstack{0.3142}{$\pm$ 0.0012} & \tblstack{+0.0011}{[$-$0.0005, +0.0027]} \\
\bottomrule
\end{tabular}
\end{lrbox}
\AppendixTableOutput{generated_control_only_replication-1}
\end{table}

% Appendix presentation derived from ../generated_control_adjusted_residual.tex
\begin{table}[!htbp]
\centering
\caption{Control-adjusted Fold-4 response evaluation. Target and prediction are residualized by the same paired control-only prediction $g$ before Global PCC is computed. Chemical effect is the mean residual-PCC reduction after a target-blind chemical shuffle. Control-only self-comparison has exactly zero increment, so residual PCC and its shuffle contrast are undefined ($\na$). MSE is unchanged: $\operatorname{MSE}(f-g,y-g)=\operatorname{MSE}(f,y)$; the displayed MSE remains on the original response scale. Residual PCC is mean [95\% $t$ interval] over five paired fitting seeds.}
\label{tab:control-adjusted-residual}
\tblsetup
\begin{lrbox}{\AppendixTableBox}
\begin{tabular}{@{}L{99pt}R{126pt}R{62pt}R{91pt}@{}}
\toprule
\rowcolor{csHeader}\multicolumn{4}{l}{\textbf{A. BBBC036}} \\
Model & \tblhdr{Residual PCC $\uparrow$\\Mean [95\% CI]} & \tblhdr{MSE\\$\downarrow$} & \tblhdr{Pert. effect\\$\uparrow$} \\
\midrule
$h_0$ & \tblstack{0.0474}{[0.0371, 0.0577]} & 0.0598 & 0.0381 \\
 \discoverymethod & \tblstack{0.0294}{[0.0052, 0.0535]} & 0.0586 & 0.0000 \\
\textsc{AIDE} & \tblstack{0.0301}{[0.0236, 0.0366]} & 0.0593 & 0.0050 \\
\textsc{CellForge} & \tblstack{0.0336}{[0.0234, 0.0438]} & 0.0592 & 0.0223 \\
\textsc{HarmonyCell} & \tblstack{0.0341}{[0.0226, 0.0455]} & 0.0586 & 0.0000 \\
Control-only & Zero increment & 0.0586 & \na \\
Chemical-only & \tblstack{0.0052}{[-0.0044, 0.0148]} & 0.0616 & 0.0012 \\
Simple concat & \tblstack{0.0474}{[0.0371, 0.0577]} & 0.0598 & 0.0381 \\
\addlinespace[5pt]
\rowcolor{csHeader}\multicolumn{4}{l}{\textbf{B. BBBC047}} \\
Model & \tblhdr{Residual PCC $\uparrow$\\Mean [95\% CI]} & \tblhdr{MSE\\$\downarrow$} & \tblhdr{Pert. effect\\$\uparrow$} \\
\midrule
$h_0$ & \tblstack{0.0558}{[0.0510, 0.0606]} & 0.0469 & 0.0205 \\
 \discoverymethod & \tblstack{0.0538}{[0.0422, 0.0654]} & 0.0457 & 0.0000 \\
\textsc{AIDE} & \tblstack{0.0558}{[0.0510, 0.0606]} & 0.0469 & 0.0205 \\
\textsc{CellForge} & \tblstack{0.0553}{[0.0510, 0.0596]} & 0.0467 & 0.0174 \\
\textsc{HarmonyCell} & \tblstack{0.0567}{[0.0470, 0.0663]} & 0.0459 & 0.0020 \\
Control-only & Zero increment & 0.0458 & \na \\
Chemical-only & \tblstack{0.0457}{[0.0382, 0.0532]} & 0.0480 & 0.0104 \\
Simple concat & \tblstack{0.0558}{[0.0510, 0.0606]} & 0.0469 & 0.0205 \\
\bottomrule
\end{tabular}
\end{lrbox}
\AppendixTableOutput{generated_control_adjusted_residual-1}
\end{table}

% Appendix presentation derived from ../generated_checker_calibration.tex
\begin{table}[!htbp]
\centering
\caption{Calibration of the fixed audit rule. \textbf{Top:} controlled code examples test whether source inspection identifies known implementation errors. \textbf{Bottom:} synthetic responses with known input dependence test the held-out permutation rule. These fixtures verify the fixed adjudication rule independently of the biological-model results.}
\label{tab:checker-calibration}
\tblsetup
\begin{tabularx}{\textwidth}{@{}L{92pt}L{135pt}X@{}}
\toprule
\rowcolor{csHeader}
\multicolumn{3}{c}{\textbf{A. Code inspection}} \\
Test case & Expected code pattern & Decision \\
\midrule
Compound input used & Direct compound input path & Implemented \\
Control input used & Direct control input path & Implemented \\
Gated interaction & Joint compound--control path & Implemented \\
Singleton attention & Query cancellation in the cited path & Implementation-contradicted \\
Absent source symbol & Manifest anchor missing from source & Implementation-contradicted \\
Dynamic function & Input use cannot be resolved statically & Unresolved \\
\midrule
\rowcolor{csHeader}
\multicolumn{3}{c}{\textbf{B. Held-out input test}} \\
Fixture & Supported input-use claims & Other tested claims \\
\midrule
Null & None & Chemical, control, interaction unsupported \\
Chemical positive & Chemical & Control and interaction unsupported \\
Control positive & Control & Chemical and interaction unsupported \\
Interaction positive & Chemical, control, interaction & \na \\
\bottomrule
\end{tabularx}
\end{table}

\FloatBarrier
\subsection{Known-truth measurement and uncertainty checks}
\label{app:measurement-validity}
% Appendix presentation derived from ../generated_measurement_validity_appendix.tex
% Generated by tools/render_validation_evidence.py; do not edit by hand.
\paragraph{Registered independent-dataset experiment.}
The fixed-context simulator uses $f(z,c)=(az,c)$ and $Y=(bz+e,c)$. Matched replacement draws an independent normal $z'$ while fixing $c,Y$. The loss is the \emph{sum} of squared errors over both channels, so the population paired loss increment is $\tau=2ab$; the mean squared Euclidean prediction distance is $\eta=2a^2$, not RMS. Groups receive equal weight and rows equal weight within group. The experiment evaluates equal-group-weighted population contrasts; real-data row-weighted resampling is reported separately.
Four classes are invariant $(a,b)=(0,.2)$, sensitive without benefit $(.2,0)$, beneficial $(.2,.2)$, and harmful $(-.2,.2)$. Their loss truths are $0,0,.08,-.08$ and distance truths $0,.08,.08,.08$.

The frozen $2^4=16$ scenarios cross $G\in\{8,32\}$ independent source groups, balanced size 8 or independently sampled sizes $\{2,4,8,32\}$ with equal probabilities, within-group correlation $\rho\in\{0,.6\}$, and noise SD $\sigma\in\{.25,1\}$. Each scenario has 400 independent datasets (6400 total). Maps 32/128 are nested on the same datasets, not independent cohorts. Group-percentile intervals use 999 bootstrap draws; group-$t$ uses $G-1$ degrees of freedom. Source/donor, bootstrap and context seeds are 2026091401, 2026091402 and 2026091403.

\paragraph{Conditional and population objectives.}
Map-only intervals target the donor-integrated contrast of one fixed dataset; group-aware intervals target the population expectation over independent groups. Their coverage is therefore reported against both truths. Positive evidence means a strictly positive lower bound of a two-sided 95\% CI: the positive one-tail nominal Type-I error is $.025$, not $.05$. Power is positive-direction evidence for benefit and negative-direction evidence for harm. Prediction distance does not test target benefit. Invariant data and intervals are identically zero and are reported as a deterministic check, not ``100\% calibrated'' stochastic coverage.

\begin{table}[!htbp]
\centering
\caption{Calibration ranges across all 16 scenarios under the two nested donor-map budgets.}
\label{tab:measurement-validity-ranges-32}
\label{tab:measurement-validity-ranges}
\tblsetup
\begin{lrbox}{\AppendixTableBox}
\begin{tabular}{@{}L{76pt}L{75pt}R{70pt}R{70pt}R{70pt}@{}}
\toprule
\rowcolor{csHeader}
Endpoint / class & CI method & \tblhdr{Population\\coverage} & \tblhdr{Conditional\\coverage} & \tblhdr{Directed\\evidence rate} \\
\midrule
\rowcolor{csHeader}\multicolumn{5}{@{}l}{\textbf{32 donor maps}} \\
Null loss & Map only & $[0.1500,\,0.3000]$ & $[0.9275,\,0.9675]$ & $[0.2850,\,0.5000]$ \\
Null loss & Group $t$ & $[0.9200,\,0.9775]$ & $[1.0000,\,1.0000]$ & $[0.0150,\,0.0725]$ \\
Null loss & Group bootstrap & $[0.8175,\,0.9400]$ & $[1.0000,\,1.0000]$ & $[0.0350,\,0.1475]$ \\
Beneficial loss & Map only & $[0.1425,\,0.3350]$ & $[0.9250,\,0.9650]$ & $[0.7100,\,1.0000]$ \\
Beneficial loss & Group $t$ & $[0.9100,\,0.9650]$ & $[0.9975,\,1.0000]$ & $[0.1200,\,1.0000]$ \\
Beneficial loss & Group bootstrap & $[0.8350,\,0.9425]$ & $[0.9975,\,1.0000]$ & $[0.2650,\,1.0000]$ \\
Harmful loss & Map only & $[0.1050,\,0.2750]$ & $[0.9375,\,0.9650]$ & $[0.6775,\,1.0000]$ \\
Harmful loss & Group $t$ & $[0.8625,\,0.9725]$ & $[1.0000,\,1.0000]$ & $[0.0550,\,1.0000]$ \\
Harmful loss & Group bootstrap & $[0.8300,\,0.9400]$ & $[1.0000,\,1.0000]$ & $[0.1900,\,1.0000]$ \\
Squared distance & Map only & $[0.2175,\,0.5250]$ & $[0.9400,\,0.9700]$ & $[1.0000,\,1.0000]$ \\
Squared distance & Group $t$ & $[0.8375,\,0.9525]$ & $[1.0000,\,1.0000]$ & $[0.9975,\,1.0000]$ \\
Squared distance & Group bootstrap & $[0.7800,\,0.9375]$ & $[0.9975,\,1.0000]$ & $[1.0000,\,1.0000]$ \\
\addlinespace[3pt]
\rowcolor{csHeader}\multicolumn{5}{@{}l}{\textbf{128 donor maps}} \\
Null loss & Map only & $[0.0675,\,0.1625]$ & $[0.9225,\,0.9725]$ & $[0.3600,\,0.5375]$ \\
Null loss & Group $t$ & $[0.9125,\,0.9775]$ & $[1.0000,\,1.0000]$ & $[0.0125,\,0.0775]$ \\
Null loss & Group bootstrap & $[0.8200,\,0.9350]$ & $[1.0000,\,1.0000]$ & $[0.0350,\,0.1450]$ \\
Beneficial loss & Map only & $[0.0750,\,0.1750]$ & $[0.9175,\,0.9650]$ & $[0.7675,\,1.0000]$ \\
Beneficial loss & Group $t$ & $[0.9075,\,0.9725]$ & $[1.0000,\,1.0000]$ & $[0.1200,\,1.0000]$ \\
Beneficial loss & Group bootstrap & $[0.8400,\,0.9450]$ & $[1.0000,\,1.0000]$ & $[0.2475,\,1.0000]$ \\
Harmful loss & Map only & $[0.0400,\,0.1450]$ & $[0.9125,\,0.9600]$ & $[0.7025,\,1.0000]$ \\
Harmful loss & Group $t$ & $[0.8600,\,0.9700]$ & $[1.0000,\,1.0000]$ & $[0.0500,\,1.0000]$ \\
Harmful loss & Group bootstrap & $[0.8200,\,0.9425]$ & $[1.0000,\,1.0000]$ & $[0.1875,\,1.0000]$ \\
Squared distance & Map only & $[0.0925,\,0.2900]$ & $[0.9350,\,0.9600]$ & $[1.0000,\,1.0000]$ \\
Squared distance & Group $t$ & $[0.8425,\,0.9500]$ & $[1.0000,\,1.0000]$ & $[0.9975,\,1.0000]$ \\
Squared distance & Group bootstrap & $[0.8025,\,0.9350]$ & $[1.0000,\,1.0000]$ & $[1.0000,\,1.0000]$ \\
\bottomrule
\end{tabular}
\end{lrbox}
\AppendixTableOutput{generated_measurement_validity_appendix-1}

\tblnote{Each rate uses 400 independent datasets per scenario; ranges are minima/maxima over scenarios, not pooled coverage. The last column is positive Type I for null loss (nominal .025), correctly directed power for beneficial/harmful loss, and positive-distance evidence for distance. Conditional and population targets differ. Invariant cases are excluded from stochastic coverage.}
\end{table}

\begin{table}[!htbp]
\centering
\caption{All 16 null-loss scenarios at 128 maps: population coverage / positive Type I (each cell, top / bottom).}
\label{tab:measurement-validity-scenarios}
\tblsetup
\begin{lrbox}{\AppendixTableBox}
\begin{tabular}{@{}L{75pt}R{48pt}R{48pt}R{54pt}R{47pt}R{93pt}@{}}
\toprule
\rowcolor{csHeader}
$G$/size/$\rho$/$\sigma$ & Map only & Group $t$ & \tblhdr{Group\\bootstrap} & \tblhdr{$t$ Type-I\\MCSE} & \tblhdr{$t$ Type-I\\Wilson 95\% CI} \\
\midrule
8/B/0/0.25 & \makecell[r]{0.1275\\0.4600} & \makecell[r]{0.9500\\0.0400} & \makecell[r]{0.8675\\0.0950} & $0.0098$ & $[0.0248,\,0.0640]$ \\
8/B/0/1 & \makecell[r]{0.1175\\0.4925} & \makecell[r]{0.9550\\0.0325} & \makecell[r]{0.8950\\0.0700} & $0.0089$ & $[0.0191,\,0.0548]$ \\
8/B/0.6/0.25 & \makecell[r]{0.0875\\0.5075} & \makecell[r]{0.9325\\0.0600} & \makecell[r]{0.8200\\0.1450} & $0.0119$ & $[0.0406,\,0.0877]$ \\
8/B/0.6/1 & \makecell[r]{0.0675\\0.4925} & \makecell[r]{0.9650\\0.0225} & \makecell[r]{0.8525\\0.0850} & $0.0074$ & $[0.0119,\,0.0422]$ \\
8/I/0/0.25 & \makecell[r]{0.1300\\0.4675} & \makecell[r]{0.9325\\0.0650} & \makecell[r]{0.8500\\0.1275} & $0.0123$ & $[0.0447,\,0.0935]$ \\
8/I/0/1 & \makecell[r]{0.1550\\0.3600} & \makecell[r]{0.9775\\0.0125} & \makecell[r]{0.8825\\0.0650} & $0.0056$ & $[0.0054,\,0.0289]$ \\
8/I/0.6/0.25 & \makecell[r]{0.0725\\0.5375} & \makecell[r]{0.9350\\0.0650} & \makecell[r]{0.8400\\0.1375} & $0.0123$ & $[0.0447,\,0.0935]$ \\
8/I/0.6/1 & \makecell[r]{0.0825\\0.4625} & \makecell[r]{0.9600\\0.0250} & \makecell[r]{0.8300\\0.1000} & $0.0078$ & $[0.0136,\,0.0454]$ \\
32/B/0/0.25 & \makecell[r]{0.1125\\0.4275} & \makecell[r]{0.9325\\0.0300} & \makecell[r]{0.9225\\0.0375} & $0.0085$ & $[0.0172,\,0.0517]$ \\
32/B/0/1 & \makecell[r]{0.1350\\0.4500} & \makecell[r]{0.9625\\0.0225} & \makecell[r]{0.9350\\0.0350} & $0.0074$ & $[0.0119,\,0.0422]$ \\
32/B/0.6/0.25 & \makecell[r]{0.0850\\0.4650} & \makecell[r]{0.9325\\0.0575} & \makecell[r]{0.9175\\0.0675} & $0.0116$ & $[0.0386,\,0.0848]$ \\
32/B/0.6/1 & \makecell[r]{0.0875\\0.4600} & \makecell[r]{0.9600\\0.0250} & \makecell[r]{0.9350\\0.0350} & $0.0078$ & $[0.0136,\,0.0454]$ \\
32/I/0/0.25 & \makecell[r]{0.1450\\0.3975} & \makecell[r]{0.9450\\0.0425} & \makecell[r]{0.9225\\0.0500} & $0.0101$ & $[0.0267,\,0.0670]$ \\
32/I/0/1 & \makecell[r]{0.1625\\0.4175} & \makecell[r]{0.9450\\0.0375} & \makecell[r]{0.9225\\0.0425} & $0.0095$ & $[0.0229,\,0.0609]$ \\
32/I/0.6/0.25 & \makecell[r]{0.0975\\0.4900} & \makecell[r]{0.9125\\0.0775} & \makecell[r]{0.8925\\0.0850} & $0.0134$ & $[0.0551,\,0.1079]$ \\
32/I/0.6/1 & \makecell[r]{0.0750\\0.4450} & \makecell[r]{0.9375\\0.0450} & \makecell[r]{0.9025\\0.0625} & $0.0104$ & $[0.0287,\,0.0700]$ \\
\bottomrule
\end{tabular}
\end{lrbox}
\AppendixTableOutput{generated_measurement_validity_appendix-2}
\tblnote{B/I: balanced/imbalanced. Type-I nominal .025. MCSE is $\sqrt{p(1-p)/400}$ (worst possible .025); Wilson intervals are pointwise Monte Carlo intervals, not simultaneous bounds over 16 scenarios.}
\end{table}

\begin{table}[!htbp]
\centering
\caption{Estimator bias and Monte Carlo uncertainty across the 16 scenarios.}
\label{tab:measurement-validity-bias}
\tblsetup
\begin{lrbox}{\AppendixTableBox}
\begin{tabular}{@{}L{78pt}C{29pt}R{35pt}R{112pt}R{112pt}@{}}
\toprule
\rowcolor{csHeader}
Endpoint / class & Maps & Truth & Bias range & Bias MCSE range \\
\midrule
Null loss & 32 & $0$ & $[-8.41\!\times\!10^{-3},\,4.95\!\times\!10^{-3}]$ & $[3.84\!\times\!10^{-4},\,5.21\!\times\!10^{-3}]$ \\
Null loss & 128 & $0$ & $[-8.42\!\times\!10^{-3},\,5.48\!\times\!10^{-3}]$ & $[3.82\!\times\!10^{-4},\,5.14\!\times\!10^{-3}]$ \\
Beneficial loss & 32 & $0.08$ & $[-7.32\!\times\!10^{-3},\,4.58\!\times\!10^{-3}]$ & $[3.50\!\times\!10^{-4},\,5.14\!\times\!10^{-3}]$ \\
Beneficial loss & 128 & $0.08$ & $[-7.47\!\times\!10^{-3},\,5.06\!\times\!10^{-3}]$ & $[3.47\!\times\!10^{-4},\,5.06\!\times\!10^{-3}]$ \\
Harmful loss & 32 & $-0.08$ & $[-4.17\!\times\!10^{-3},\,6.55\!\times\!10^{-3}]$ & $[5.70\!\times\!10^{-4},\,5.48\!\times\!10^{-3}]$ \\
Harmful loss & 128 & $-0.08$ & $[-4.68\!\times\!10^{-3},\,6.51\!\times\!10^{-3}]$ & $[5.65\!\times\!10^{-4},\,5.40\!\times\!10^{-3}]$ \\
Squared distance & 32 & $0.08$ & $[-4.19\!\times\!10^{-4},\,7.09\!\times\!10^{-4}]$ & $[1.69\!\times\!10^{-4},\,7.60\!\times\!10^{-4}]$ \\
Squared distance & 128 & $0.08$ & $[-3.08\!\times\!10^{-4},\,5.67\!\times\!10^{-4}]$ & $[1.68\!\times\!10^{-4},\,7.49\!\times\!10^{-4}]$ \\
\bottomrule
\end{tabular}
\end{lrbox}
\AppendixTableOutput{generated_measurement_validity_appendix-3}
\tblnote{All CI methods share the same point estimator. Bias MCSE is the empirical SD of estimation error divided by $\sqrt{400}$; full-precision per-scenario bias, RMSE, coverage and Wilson intervals are in the portable CSV.}
\end{table}

\begin{table}[!htbp]
\centering
\caption{Historical reference-relative gate replay at 128 maps with unrelated context variation.}
\label{tab:measurement-validity-reference}
\tblsetup
\begin{lrbox}{\AppendixTableBox}
\begin{tabular}{@{}L{91pt}C{40pt}R{66pt}R{66pt}R{101pt}@{}}
\toprule
\rowcolor{csHeader}
Class & Amplitude & Mean effect & Mean $Q$ & Support-rate range \\
\midrule
Invariant & $0$ & $0$ & $0$ & $[0,\,0]$ \\
Invariant & $0.2$ & $0$ & $0.08792$ & $[0,\,0]$ \\
Invariant & $2.0$ & $0$ & $8.79230$ & $[0,\,0]$ \\
Sensitive/no benefit & $0$ & $-0.00067$ & $0.07744$ & $[0,\,0.007]$ \\
Sensitive/no benefit & $0.2$ & $-0.00067$ & $0.12034$ & $[0,\,0.003]$ \\
Sensitive/no benefit & $2.0$ & $-0.00067$ & $8.79274$ & $[0,\,0]$ \\
Beneficial & $0$ & $0.07946$ & $0.08226$ & $[0,\,1.000]$ \\
Beneficial & $0.2$ & $0.07946$ & $0.12260$ & $[0,\,0.110]$ \\
Beneficial & $2.0$ & $0.07946$ & $8.79279$ & $[0,\,0]$ \\
Harmful & $0$ & $-0.07961$ & $0.08218$ & $[0,\,0]$ \\
Harmful & $0.2$ & $-0.07961$ & $0.12276$ & $[0,\,0]$ \\
Harmful & $2.0$ & $-0.07961$ & $8.79248$ & $[0,\,0]$ \\
\bottomrule
\end{tabular}
\end{lrbox}
\AppendixTableOutput{generated_measurement_validity_appendix-4}
\tblnote{Means average all 16 scenarios; support ranges remain scenario-specific. This replays the historical formula on negative squared-loss scores, not historical PCC. Changing orthogonal context changes its reference threshold while keeping the chemical contrast fixed. Non-support is not a false negative for the different proposition $\tau>0$; both map budgets and all original statuses remain in the portable CSV.}
\end{table}

\paragraph{Limits retained in interpretation.}
The group-aware methods still under-cover in some scenarios; the bootstrap is particularly weak with few groups, and positive null declarations can exceed the nominal one-tail rate. Independent normal donors, independent groups, and correct context matching are explicit simulation assumptions, not demonstrations of real biological conditional exchangeability or support. Finite reused donor pools and dependent biological groups require additional dependence handling. More maps reduce conditional Monte Carlo error but do not create more biological replicates.

\FloatBarrier
\section{Complete input-intervention results}
\label{app:intervention-evidence}

The complete intervention matrix reports search performance, held-out predictive performance, compound-identity effect, control-profile effect, target-loss gain, and input-use status in separate columns. Rows summarize the full pre-specified model sets, and $\na$ marks contrasts that cannot be identified under the protocol. Response-block decompositions are diagnostic views of the same selected models and fitted checkpoints. Figure~\ref{fig:bbbc-input-audit} follows the compound and control effects of the falsification-guided checkpoints across both held-out folds.

% Appendix presentation derived from ../generated_evidence_intervention_matrix.tex
\begin{table}[!htbp]
\centering
\caption{Prediction and input-use evidence across three model-search settings. The prediction-score source is chosen by the Fold-4 score rule and summarized across five paired refits; constrained settings retain all ten trajectory endpoints. Values are mean $\pm$ 95\% interval half-width. Factorial effects $E_{\rm cmpd}$ and $E_{\rm ctrl}$ average PCC drops over the other input's two states; $E^{\mathcal L}_{\rm cmpd}$ is the corresponding loss effect. Q/U/I reports contribution relative to the registered reference variation.}
\label{tab:evidence-matrix}
\tblsetup
\begin{lrbox}{\AppendixTableBox}
\begin{tabular}{@{}L{60pt}L{110pt}L{136pt}R{72pt}@{}}
\toprule
\rowcolor{csHeader}
\multicolumn{4}{l}{\textbf{A. Selection record}} \\
\rowcolor{csHeader}
Task & Search setting & Selected source & Fold-3 PCC \\
\midrule
BBBC036 & Prediction-score & HarmonyCell & 0.2985 \\
BBBC036 & Path-constrained & 10-endpoint cohort & \tblstack{0.2925}{$\pm$ 0.0019} \\
BBBC036 & Falsification-guided & 10-endpoint cohort & \tblstack{0.3006}{$\pm$ 0.0006} \\
BBBC047 & Prediction-score & CellScientist & 0.3231 \\
BBBC047 & Path-constrained & 10-endpoint cohort & \tblstack{0.3008}{$\pm$ 0.0012} \\
BBBC047 & Falsification-guided & 10-endpoint cohort & \tblstack{0.3238}{$\pm$ 0.0002} \\
\bottomrule
\end{tabular}
\end{lrbox}
\AppendixTableOutput{generated_evidence_intervention_matrix-1}
\par\vspace{5pt}
\tblsetup
\begin{lrbox}{\AppendixTableBox}
\begin{tabular}{@{}L{105pt}R{50pt}R{50pt}R{57pt}R{60pt}R{44pt}@{}}
\toprule
\rowcolor{csHeader}
\multicolumn{6}{l}{\textbf{B. Fold 4: held-out input-use evidence}} \\
\rowcolor{csHeader}
Search setting & PCC & $E_{\rm cmpd}$ & $E^{\mathcal L}_{\rm cmpd}$ & $E_{\rm ctrl}$ & Q/U/I \\
\midrule
\rowcolor{csHeader}\multicolumn{6}{l}{\textbf{BBBC036}} \\
Prediction-score & \tblstack{0.3110}{$\pm$ 0.0016} & \tblstack{0.0000}{$\pm$ 0.0000} & \tblstack{0.00000}{$\pm$ 0.00000} & \tblstack{0.1693}{$\pm$ 0.0033} & 0/5/0 \\
Path-constrained & \tblstack{0.3034}{$\pm$ 0.0029} & \tblstack{0.0007}{$\pm$ 0.0006} & \tblstack{0.00003}{$\pm$ 0.00003} & \tblstack{0.1592}{$\pm$ 0.0050} & 0/10/0 \\
Falsification-guided & \tblstack{0.3167}{$\pm$ 0.0013} & \tblstack{0.0091}{$\pm$ 0.0010} & \tblstack{0.00039}{$\pm$ 0.00004} & \tblstack{0.1716}{$\pm$ 0.0042} & 0/10/0 \\
\midrule
\rowcolor{csHeader}\multicolumn{6}{l}{\textbf{BBBC047}} \\
Prediction-score & \tblstack{0.3035}{$\pm$ 0.0008} & \tblstack{0.0000}{$\pm$ 0.0000} & \tblstack{0.00000}{$\pm$ 0.00000} & \tblstack{0.1932}{$\pm$ 0.0044} & 0/5/0 \\
Path-constrained & \tblstack{0.2822}{$\pm$ 0.0017} & \tblstack{0.0083}{$\pm$ 0.0016} & \tblstack{0.00024}{$\pm$ 0.00005} & \tblstack{0.1388}{$\pm$ 0.0052} & 0/6/4 \\
Falsification-guided & \tblstack{0.3060}{$\pm$ 0.0004} & \tblstack{0.0053}{$\pm$ 0.0002} & \tblstack{0.00018}{$\pm$ 0.00001} & \tblstack{0.1988}{$\pm$ 0.0027} & 0/10/0 \\
\bottomrule
\end{tabular}
\end{lrbox}
\AppendixTableOutput{generated_evidence_intervention_matrix-2}
\par\vspace{5pt}
\tblsetup
\begin{lrbox}{\AppendixTableBox}
\begin{tabular}{@{}L{105pt}R{50pt}R{50pt}R{57pt}R{60pt}R{44pt}@{}}
\toprule
\rowcolor{csHeader}
\multicolumn{6}{l}{\textbf{C. Fold 5: held-out input-use evidence}} \\
\rowcolor{csHeader}
Search setting & PCC & $E_{\rm cmpd}$ & $E^{\mathcal L}_{\rm cmpd}$ & $E_{\rm ctrl}$ & Q/U/I \\
\midrule
\rowcolor{csHeader}\multicolumn{6}{l}{\textbf{BBBC036}} \\
Prediction-score & \tblstack{0.2932}{$\pm$ 0.0020} & \tblstack{0.0000}{$\pm$ 0.0000} & \tblstack{0.00000}{$\pm$ 0.00000} & \tblstack{0.1516}{$\pm$ 0.0035} & 0/5/0 \\
Path-constrained & \tblstack{0.2886}{$\pm$ 0.0018} & \tblstack{0.0039}{$\pm$ 0.0022} & \tblstack{0.00016}{$\pm$ 0.00009} & \tblstack{0.1423}{$\pm$ 0.0050} & 0/10/0 \\
Falsification-guided & \tblstack{0.2897}{$\pm$ 0.0008} & \tblstack{0.0022}{$\pm$ 0.0012} & \tblstack{0.00009}{$\pm$ 0.00005} & \tblstack{0.1515}{$\pm$ 0.0037} & 0/10/0 \\
\midrule
\rowcolor{csHeader}\multicolumn{6}{l}{\textbf{BBBC047}} \\
Prediction-score & \tblstack{0.3153}{$\pm$ 0.0014} & \tblstack{0.0000}{$\pm$ 0.0000} & \tblstack{0.00000}{$\pm$ 0.00000} & \tblstack{0.2004}{$\pm$ 0.0061} & 0/5/0 \\
Path-constrained & \tblstack{0.2922}{$\pm$ 0.0018} & \tblstack{0.0074}{$\pm$ 0.0017} & \tblstack{0.00022}{$\pm$ 0.00005} & \tblstack{0.1477}{$\pm$ 0.0057} & 0/7/3 \\
Falsification-guided & \tblstack{0.3173}{$\pm$ 0.0004} & \tblstack{0.0048}{$\pm$ 0.0004} & \tblstack{0.00017}{$\pm$ 0.00001} & \tblstack{0.2051}{$\pm$ 0.0024} & 0/10/0 \\
\bottomrule
\end{tabular}
\end{lrbox}
\AppendixTableOutput{generated_evidence_intervention_matrix-3}
\end{table}

% Appendix presentation derived from ../generated_d3_anchor_evidence.tex
\begin{table}[!htbp]
\centering
\caption{\textbf{Falsification-guided models increase average held-out compound contribution, while predictive increments vary by task and fold.} $\Delta$ compares the full and control-only predictors on the same fold. Factorial effects $E_{\rm cmpd}$ and $E_{\rm ctrl}$ average PCC drops over the other input's two states; $E^{\mathcal L}_{\rm cmpd}$ averages the corresponding loss increases. Q/U/I summarizes per-model contribution relative to the registered reference variation.}
\label{tab:d3-anchor}
\tblsetup
\begin{lrbox}{\AppendixTableBox}
\begin{tabular}{@{}L{60pt}C{31pt}R{86pt}R{80pt}R{115pt}@{}}
\toprule
\rowcolor{csHeader}
\multicolumn{5}{l}{\textbf{A. Predictive performance}} \\
Task & Fold & Control-only PCC & Full PCC & $\Delta$ Full$-$Control \\
\midrule
BBBC036 & Fold 4 & \tblstack{0.3121}{$\pm$ 0.0011} & \tblstack{0.3167}{$\pm$ 0.0013} & \tblstack{0.0046}{$\pm$ 0.0006} \\
BBBC036 & Fold 5 & \tblstack{0.2917}{$\pm$ 0.0006} & \tblstack{0.2897}{$\pm$ 0.0008} & \tblstack{-0.0020}{$\pm$ 0.0009} \\
BBBC047 & Fold 4 & \tblstack{0.3023}{$\pm$ 0.0004} & \tblstack{0.3060}{$\pm$ 0.0004} & \tblstack{0.0037}{$\pm$ 0.0003} \\
BBBC047 & Fold 5 & \tblstack{0.3145}{$\pm$ 0.0005} & \tblstack{0.3173}{$\pm$ 0.0004} & \tblstack{0.0028}{$\pm$ 0.0004} \\
\bottomrule
\end{tabular}
\end{lrbox}
\AppendixTableOutput{generated_d3_anchor_evidence-1}
\par\vspace{5pt}
\tblsetup
\begin{lrbox}{\AppendixTableBox}
\begin{tabular}{@{}L{60pt}C{31pt}R{68pt}R{80pt}R{67pt}R{60pt}@{}}
\toprule
\rowcolor{csHeader}
\multicolumn{6}{l}{\textbf{B. Input-use evidence}} \\
Task & Fold & $E_{\rm cmpd}$ & $E^{\mathcal L}_{\rm cmpd}$ & $E_{\rm ctrl}$ & Model Q/U/I \\
\midrule
BBBC036 & Fold 4 & \tblstack{0.0091}{$\pm$ 0.0010} & \tblstack{0.00039}{$\pm$ 0.00004} & \tblstack{0.1716}{$\pm$ 0.0042} & 0/10/0 \\
BBBC036 & Fold 5 & \tblstack{0.0022}{$\pm$ 0.0012} & \tblstack{0.00009}{$\pm$ 0.00005} & \tblstack{0.1515}{$\pm$ 0.0037} & 0/10/0 \\
BBBC047 & Fold 4 & \tblstack{0.0053}{$\pm$ 0.0002} & \tblstack{0.00018}{$\pm$ 0.00001} & \tblstack{0.1988}{$\pm$ 0.0027} & 0/10/0 \\
BBBC047 & Fold 5 & \tblstack{0.0048}{$\pm$ 0.0004} & \tblstack{0.00017}{$\pm$ 0.00001} & \tblstack{0.2051}{$\pm$ 0.0024} & 0/10/0 \\
\bottomrule
\end{tabular}
\end{lrbox}
\AppendixTableOutput{generated_d3_anchor_evidence-2}
\end{table}

\begin{figure}[!htbp]
    \centering
    \input{figures/bbbc_panels}
    \caption{\textbf{Compound sensitivity and control dependence across the BBBC cohorts.} Each cohort contributes ten trajectory-selected checkpoints, evaluated unchanged on both folds. (a,b) Checkpoint-level factorial effects $E_{\rm cmpd}$ and $E_{\rm ctrl}$; diamonds and bars give ten-endpoint means and 95\% $t$-intervals. (c,d) Lower effect quantile minus the checkpoint-local threshold; lines connect the same checkpoint across folds. Inputs use distinct vertical scales. Dose contrasts are non-identifiable under the registered eligibility rule (Table~\ref{tab:all-input-status}). Table~\ref{tab:d3-anchor} gives predictive comparisons.}
    \label{fig:bbbc-input-audit}
\end{figure}

% Appendix presentation derived from ../generated_d3_residual_frontier.tex
\begin{table}[!htbp]
\centering
\caption{\textbf{Predictive performance and perturbation evidence define a measurable frontier.} Falsification-guided entries summarize ten independently selected models; fixed residual learners summarize five paired refits. Global PCC, compound-shuffle PCC drop, and compound-shuffle loss increase are evaluated on the same responses, data splits, and fixed input permutations.}
\label{tab:d3-residual-frontier}
\tblsetup
\begin{lrbox}{\AppendixTableBox}
\begin{tabular}{@{}L{103pt}R{82pt}R{93pt}R{100pt}@{}}
\toprule
\rowcolor{csHeader}
Predictor & \tblhdr{Global PCC\\{[95\% CI]}} & \tblhdr{Chemical PCC drop\\{[95\% CI]}} & \tblhdr{Target-loss gain\\{[95\% CI]}} \\
\midrule
\rowcolor{csHeader}\multicolumn{4}{l}{\textbf{BBBC036 / Fold 4}} \\
Falsification-guided & \tblstack{0.3167}{[0.3154, 0.3181]} & \tblstack{0.0091}{[0.0081, 0.0101]} & \tblstack{0.00039}{[0.00034, 0.00043]} \\
Residual MLP & \tblstack{0.3056}{[0.3027, 0.3085]} & \tblstack{0.0036}{[0.0028, 0.0044]} & \tblstack{0.00015}{[0.00012, 0.00019]} \\
Joint Ridge & \tblstack{0.0892}{[0.0867, 0.0917]} & \tblstack{0.0153}{[0.0147, 0.0159]} & \tblstack{0.00299}{[0.00290, 0.00308]} \\
\midrule
\rowcolor{csHeader}\multicolumn{4}{l}{\textbf{BBBC036 / Fold 5}} \\
Falsification-guided & \tblstack{0.2897}{[0.2890, 0.2905]} & \tblstack{0.0022}{[0.0010, 0.0034]} & \tblstack{0.00009}{[0.00004, 0.00014]} \\
Residual MLP & \tblstack{0.2861}{[0.2829, 0.2892]} & \tblstack{0.0007}{[0.0001, 0.0013]} & \tblstack{0.00003}{[0.00000, 0.00005]} \\
Joint Ridge & \tblstack{0.1114}{[0.1092, 0.1136]} & \tblstack{0.0336}{[0.0328, 0.0344]} & \tblstack{0.00569}{[0.00557, 0.00582]} \\
\midrule
\rowcolor{csHeader}\multicolumn{4}{l}{\textbf{BBBC047 / Fold 4}} \\
Falsification-guided & \tblstack{0.3060}{[0.3056, 0.3064]} & \tblstack{0.0053}{[0.0051, 0.0055]} & \tblstack{0.00018}{[0.00017, 0.00019]} \\
Residual MLP & \tblstack{0.3031}{[0.3007, 0.3055]} & \tblstack{0.0119}{[0.0094, 0.0145]} & \tblstack{0.00042}{[0.00033, 0.00050]} \\
Joint Ridge & \tblstack{0.0774}{[0.0765, 0.0783]} & \tblstack{0.0082}{[0.0079, 0.0085]} & \tblstack{0.00420}{[0.00416, 0.00423]} \\
\midrule
\rowcolor{csHeader}\multicolumn{4}{l}{\textbf{BBBC047 / Fold 5}} \\
Falsification-guided & \tblstack{0.3173}{[0.3169, 0.3177]} & \tblstack{0.0048}{[0.0044, 0.0052]} & \tblstack{0.00017}{[0.00015, 0.00018]} \\
Residual MLP & \tblstack{0.3154}{[0.3135, 0.3174]} & \tblstack{0.0113}{[0.0095, 0.0132]} & \tblstack{0.00040}{[0.00033, 0.00047]} \\
Joint Ridge & \tblstack{0.0704}{[0.0693, 0.0715]} & \tblstack{0.0052}{[0.0050, 0.0053]} & \tblstack{0.00394}{[0.00393, 0.00395]} \\
\bottomrule
\end{tabular}
\end{lrbox}
\AppendixTableOutput{generated_d3_residual_frontier-1}
\end{table}

% Appendix presentation derived from ../generated_modality_decomposition.tex
\begin{table}[!htbp]
\centering
\caption{Response-block decomposition of the pre-specified $2\times2$ input audit. Perturbation and context columns are factorial PCC effects $E_{\rm cmpd}$ and $E_{\rm ctrl}$; interaction is their score nonadditivity. The blocks locate input contribution within the joint response. Values are mean $\pm$ 95\% interval half-width across selected models.}
\label{tab:modality-decomposition}
\tblsetup
\begin{lrbox}{\AppendixTableBox}
\begin{tabular}{@{}L{36pt}R{54pt}R{54pt}R{54pt}R{54pt}R{54pt}R{54pt}@{}}
\toprule
\rowcolor{csHeader}
Fold & \multicolumn{2}{c}{Perturbation} & \multicolumn{2}{c}{Context} & \multicolumn{2}{c}{Interaction} \\
\rowcolor{csHeader}
 & CP & L1000 & CP & L1000 & CP & L1000 \\
\midrule
\rowcolor{csHeader}\multicolumn{7}{l}{\textbf{Path-constrained / BBBC036}} \\
F4 & \tblstack{0.0012}{$\pm$ 0.0009} & \tblstack{0.0004}{$\pm$ 0.0005} & \tblstack{0.0250}{$\pm$ 0.0023} & \tblstack{0.2164}{$\pm$ 0.0057} & \tblstack{0.0003}{$\pm$ 0.0004} & \tblstack{-0.0003}{$\pm$ 0.0006} \\
F5 & \tblstack{0.0008}{$\pm$ 0.0004} & \tblstack{0.0053}{$\pm$ 0.0031} & \tblstack{0.0290}{$\pm$ 0.0024} & \tblstack{0.1935}{$\pm$ 0.0056} & \tblstack{0.0005}{$\pm$ 0.0002} & \tblstack{0.0013}{$\pm$ 0.0008} \\
\midrule
\rowcolor{csHeader}\multicolumn{7}{l}{\textbf{Path-constrained / BBBC047}} \\
F4 & \tblstack{0.0130}{$\pm$ 0.0026} & \tblstack{0.0020}{$\pm$ 0.0002} & \tblstack{0.0911}{$\pm$ 0.0037} & \tblstack{0.2178}{$\pm$ 0.0054} & \tblstack{0.0026}{$\pm$ 0.0012} & \tblstack{0.0007}{$\pm$ 0.0002} \\
F5 & \tblstack{0.0118}{$\pm$ 0.0026} & \tblstack{0.0013}{$\pm$ 0.0005} & \tblstack{0.1005}{$\pm$ 0.0054} & \tblstack{0.2251}{$\pm$ 0.0054} & \tblstack{0.0011}{$\pm$ 0.0012} & \tblstack{-0.0001}{$\pm$ 0.0003} \\
\midrule
\rowcolor{csHeader}\multicolumn{7}{l}{\textbf{Falsification-guided / BBBC036}} \\
F4 & \tblstack{0.0141}{$\pm$ 0.0023} & \tblstack{0.0069}{$\pm$ 0.0007} & \tblstack{0.0248}{$\pm$ 0.0015} & \tblstack{0.2363}{$\pm$ 0.0041} & \tblstack{0.0000}{$\pm$ 0.0000} & \tblstack{0.0001}{$\pm$ 0.0000} \\
F5 & \tblstack{-0.0014}{$\pm$ 0.0016} & \tblstack{0.0039}{$\pm$ 0.0013} & \tblstack{0.0284}{$\pm$ 0.0016} & \tblstack{0.2088}{$\pm$ 0.0035} & \tblstack{0.0001}{$\pm$ 0.0000} & \tblstack{0.0002}{$\pm$ 0.0001} \\
\midrule
\rowcolor{csHeader}\multicolumn{7}{l}{\textbf{Falsification-guided / BBBC047}} \\
F4 & \tblstack{0.0088}{$\pm$ 0.0003} & \tblstack{0.0014}{$\pm$ 0.0001} & \tblstack{0.1338}{$\pm$ 0.0031} & \tblstack{0.2759}{$\pm$ 0.0013} & \tblstack{0.0002}{$\pm$ 0.0001} & \tblstack{0.0001}{$\pm$ 0.0000} \\
F5 & \tblstack{0.0079}{$\pm$ 0.0007} & \tblstack{0.0013}{$\pm$ 0.0002} & \tblstack{0.1421}{$\pm$ 0.0025} & \tblstack{0.2806}{$\pm$ 0.0014} & \tblstack{-0.0000}{$\pm$ 0.0001} & \tblstack{0.0001}{$\pm$ 0.0000} \\
\bottomrule
\end{tabular}
\end{lrbox}
\AppendixTableOutput{generated_modality_decomposition-1}
\end{table}

\FloatBarrier
% Appendix presentation derived from ../generated_all_input_claim_status.tex
\begin{table}[!htbp]
\centering
\caption{Input-use outcomes for the two constrained search settings. Counts are qualified/unsupported/inconclusive/not-identifiable (Q/U/I/N) under the reference-variation criterion across ten selected models. Path-constrained dose eligibility accepts any observed dose change within the released BRD compound identity; falsification-guided eligibility additionally requires a $0.1$ log-dose contrast and at least eight eligible compounds. Identity validation gives 0/4,380 and 14/3,878 eligible BBBC047 rows on Folds 4 and 5, below the 64-row and 10\% requirements. $\na$ with N=10 denotes a non-identifiable contrast, not missing execution.}
\label{tab:all-input-status}
\tblsetup
\begin{lrbox}{\AppendixTableBox}
\begin{tabular}{@{}L{56pt}C{19pt}R{34pt}R{58pt}R{55pt}R{50pt}R{88pt}@{}}
\toprule
\rowcolor{csHeader}
Task & Fold & $\Delta_{\rm dose}$ & Compound & Control & Dose & \tblhdr{Compound$\times$\\control} \\
\midrule
\rowcolor{csHeader}\multicolumn{7}{l}{\textbf{Path-constrained}} \\
BBBC036 & 4 & \na & 0/10/0/0 & 10/0/0/0 & 0/0/0/10 & 0/10/0/0 \\
BBBC036 & 5 & \na & 0/10/0/0 & 10/0/0/0 & 0/0/0/10 & 0/10/0/0 \\
BBBC047 & 4 & \na & 0/6/4/0 & 10/0/0/0 & 0/0/0/10 & 0/10/0/0 \\
BBBC047 & 5 & \na & 0/7/3/0 & 10/0/0/0 & 0/0/0/10 & 0/10/0/0 \\
\addlinespace[2pt]
\rowcolor{csHeader}\multicolumn{7}{l}{\textbf{Falsification-guided}} \\
BBBC036 & 4 & \na & 0/10/0/0 & 10/0/0/0 & 0/0/0/10 & 0/10/0/0 \\
BBBC036 & 5 & \na & 0/10/0/0 & 10/0/0/0 & 0/0/0/10 & 0/10/0/0 \\
BBBC047 & 4 & \na & 0/10/0/0 & 10/0/0/0 & 0/0/0/10 & 0/10/0/0 \\
BBBC047 & 5 & \na & 0/10/0/0 & 10/0/0/0 & 0/0/0/10 & 0/10/0/0 \\
\bottomrule
\end{tabular}
\end{lrbox}
\AppendixTableOutput{generated_all_input_claim_status-1}
\end{table}

\FloatBarrier
\section{All-candidate source checking}
\label{app:coverage}

Table~\ref{tab:source-coverage} applies the deterministic checker to every retained prediction-score candidate and separates execution failure from checker abstention. Resolution coverage is the fraction of checked candidates whose cited source locations are all resolved. For candidates without a structured manifest declaring input use, the checker reports findings supported by pre-specified source locations. The one-key attention certificate is a code-level finding; complete-model replacement tests provide the corresponding behavioral evidence.

% Appendix presentation derived from ../generated_source_checker_coverage.tex
\begin{table}[!htbp]
\centering
\caption{Deterministic code-checker results over all prediction-score candidate slots. Resolved candidates have every registered source location linked to input-consuming code or a specific contradiction; abstention leaves at least one location unresolved. Coverage reports resolved/checked candidates and checked/total slots, rather than classification accuracy. A singleton-attention certificate concerns the cited pathway.}
\label{tab:source-coverage}
\tblsetup
\begin{lrbox}{\AppendixTableBox}
\begin{tabular}{@{}L{78pt}R{33.75pt}R{33.75pt}R{33.75pt}R{33.75pt}R{33.75pt}R{33.75pt}R{33.75pt}R{33.75pt}@{}}
\toprule
\rowcolor{csHeader}
& \multicolumn{3}{c}{Candidate slots} & \multicolumn{3}{c}{Code checker} & \multicolumn{2}{c}{Coverage} \\
\rowcolor{csHeader}
Policy & Total & Exec. & Checked & Resolved & Singleton & Abstain & \tblhdr{Resolved/\\checked} & \tblhdr{Checked/\\slots} \\
\midrule
\rowcolor{csHeader}\multicolumn{9}{l}{\textbf{BBBC036}} \\
CellScientist & 100 & 99 & 99 & 60 & 15 & 39 & 60.6\% & 99.0\% \\
AIDE & 100 & 100 & 100 & 100 & 0 & 0 & 100.0\% & 100.0\% \\
CellForge & 100 & 91 & 91 & 91 & 0 & 0 & 100.0\% & 91.0\% \\
HarmonyCell & 100 & 95 & 95 & 94 & 0 & 1 & 98.9\% & 95.0\% \\
\addlinespace[2pt]
\rowcolor{csHeader}\multicolumn{9}{l}{\textbf{BBBC047}} \\
CellScientist & 100 & 94 & 94 & 48 & 17 & 46 & 51.1\% & 94.0\% \\
AIDE & 100 & 99 & 99 & 99 & 0 & 0 & 100.0\% & 99.0\% \\
CellForge & 100 & 94 & 94 & 94 & 0 & 0 & 100.0\% & 94.0\% \\
HarmonyCell & 100 & 95 & 95 & 94 & 0 & 1 & 98.9\% & 95.0\% \\
\bottomrule
\end{tabular}
\end{lrbox}
\AppendixTableOutput{generated_source_checker_coverage-1}
\end{table}

% Appendix presentation derived from ../generated_formulation_source_coverage.tex
\begin{table}[!htbp]
\centering
\caption{\textbf{Deterministic compilation links each declared input use to code.} ``All statements resolved'' means that the checker or compiler finds supporting code or a specific contradiction for every emitted static statement. After selection, the independent behavioral audit tests the registered input-use claim; singleton-attention findings remain narrow code-level certificates.}
\label{tab:formulation-coverage}
\tblsetup
\begin{lrbox}{\AppendixTableBox}
\begin{tabular}{@{}L{102pt}R{36pt}R{36pt}R{36pt}R{36pt}R{36pt}R{36pt}R{36pt}@{}}
\toprule
\rowcolor{csHeader}
Search & Slots & Exec. & Resolved & Singleton & Abstain & Resolve \% & Exec. \% \\
\midrule
\rowcolor{csHeader}\multicolumn{8}{l}{\textbf{BBBC036}} \\
Prediction-score & 100 & 99 & 60 & 15 & 39 & 60.6\% & 99.0\% \\
Path-constrained & 100 & 100 & 100 & 0 & 0 & 100.0\% & 100.0\% \\
\addlinespace[2pt]
\rowcolor{csHeader}\multicolumn{8}{l}{\textbf{BBBC047}} \\
Prediction-score & 100 & 94 & 48 & 17 & 46 & 51.1\% & 94.0\% \\
Path-constrained & 100 & 100 & 100 & 0 & 0 & 100.0\% & 100.0\% \\
\bottomrule
\end{tabular}
\end{lrbox}
\AppendixTableOutput{generated_formulation_source_coverage-1}
\end{table}

% Appendix presentation derived from ../generated_checker_taxonomy.tex
\begin{table}[!htbp]
\centering
\caption{\textbf{Checker abstention and source-interface inspection.} A deterministic hash selects sources within each category for inspection of metadata anchors and exposed fusion interfaces. The taxonomy describes static-analysis coverage; complete-model input tests provide behavioral evidence.}
\label{tab:checker-taxonomy}
\tblsetup
\begin{tabularx}{\textwidth}{@{}L{93pt}R{30pt}R{39pt}X@{}}
\toprule
\rowcolor{csHeader}
Type & Sources & Inspected & Deterministic disposition \\
\midrule
Missing fusion source location & 75 & 12 & No cited fusion symbol is available; the checker returns unresolved and the behavioral input tests remain applicable. \\
Partial input scope & 11 & 11 & The cited fusion exposes only part of the registered input-use claim; the unchecked role receives an unresolved source status. \\
Generic unbound aliases & 1 & 1 & No declared binding identifies the generic argument names; the checker assigns an unresolved source status. \\
\bottomrule
\end{tabularx}
\end{table}

A deterministic hash selects model sources within each checker category for source-interface inspection. The inspection is restricted to pre-specified source locations and public fusion interfaces; fixed behavioral replacements remain applicable across all code-check categories.

A linked sample of 48 real candidate sources extends testing beyond the one-key attention case. Of these, 47 change their predictions after compound replacement at both boundaries; 20 have positive target-loss scaffold intervals at both. An implemented multi-tower source is exactly invariant. Source consumption, fitted dependence, and target benefit thus separate in diverse generated code (Appendix~\ref{app:existing-audit-checks}).

\FloatBarrier
\section{Prediction-space dependence}
\label{app:prediction-space}

The prediction-space analysis measures normalized RMS change and cosine change between $f(c,p,a)$ and predictions obtained after fixed input permutations. Table~\ref{tab:prediction-dependence} reports prediction-change RMS divided by the target RMS after centering each output feature across evaluation rows, with a denominator floor of $10^{-12}$. Both RMS quantities average over rows and features within the reported response block. This complements target-based effects: near-zero distance indicates negligible prediction change on the tested replacements, whereas nonzero distance with near-zero target-loss gain indicates sensitivity with little measured benefit on the observed response. The calculation requires inference only over fixed checkpoints.

% Appendix presentation derived from ../generated_prediction_dependence.tex
\begin{table}[!htbp]
\centering
\caption{Prediction-space dependence under target-blind interventions. Normalized RMS changes quantify whether the fitted outputs change under replacement; target-loss contrasts, reported separately, determine whether that change improves predictive utility.}
\label{tab:prediction-dependence}
\tblsetup
\begin{lrbox}{\AppendixTableBox}
\begin{tabular}{@{}L{94pt}R{64pt}R{64pt}R{76pt}R{74pt}@{}}
\toprule
\rowcolor{csHeader}
Model & \multicolumn{3}{c}{Compound replacement} & Control replacement \\
\rowcolor{csHeader}
 & Global & CP & L1000 & Global \\
\midrule
\rowcolor{csHeader}\multicolumn{5}{l}{\textbf{BBBC036}} \\
CellScientist & \tblstack{0.0000}{$\pm$ 0.0000} & \tblstack{0.0000}{$\pm$ 0.0000} & \tblstack{0.0000}{$\pm$ 0.0000} & \tblstack{0.3112}{$\pm$ 0.0246} \\
Chemical-only & \tblstack{0.0560}{$\pm$ 0.0013} & \tblstack{0.0580}{$\pm$ 0.0014} & \tblstack{0.0552}{$\pm$ 0.0013} & \tblstack{0.0000}{$\pm$ 0.0000} \\
Control-only & \tblstack{0.0000}{$\pm$ 0.0000} & \tblstack{0.0000}{$\pm$ 0.0000} & \tblstack{0.0000}{$\pm$ 0.0000} & \tblstack{0.3390}{$\pm$ 0.0199} \\
\midrule
\rowcolor{csHeader}\multicolumn{5}{l}{\textbf{BBBC047}} \\
CellScientist & \tblstack{0.0000}{$\pm$ 0.0000} & \tblstack{0.0000}{$\pm$ 0.0000} & \tblstack{0.0000}{$\pm$ 0.0000} & \tblstack{0.3593}{$\pm$ 0.0207} \\
Chemical-only & \tblstack{0.1253}{$\pm$ 0.0066} & \tblstack{0.1467}{$\pm$ 0.0093} & \tblstack{0.1025}{$\pm$ 0.0055} & \tblstack{0.0000}{$\pm$ 0.0000} \\
Control-only & \tblstack{0.0000}{$\pm$ 0.0000} & \tblstack{0.0000}{$\pm$ 0.0000} & \tblstack{0.0000}{$\pm$ 0.0000} & \tblstack{0.4011}{$\pm$ 0.0199} \\
\bottomrule
\end{tabular}
\end{lrbox}
\AppendixTableOutput{generated_prediction_dependence-1}
\end{table}

% Appendix presentation derived from ../generated_discovery_casebook.tex
\section{Representative discovery records}
\label{app:casebook}

The main text follows one model from selection through source checking and held-out replication. This section complements that input-use analysis with two preselected trajectory-level run records: a representative successful search and a search containing code repair. Each record preserves the initial hypothesis, decisive score changes, component edits, failures, repair actions, prompt and source hashes, and final selection. The complete ten-slot traces show both successful and failed execution paths.

The two cases are selected using Fold-3 run records only: a lower-median BBBC036 trajectory and the highest-scoring BBBC047 trajectory among those with at least one recorded candidate-code repair. Their records include the initial program, complete candidate sequence, parent hashes, structured decision context, candidate source, execution and repair outcomes, provider request/response hashes, and the versioned prompt constructor that defines the generation input.

\subsection{Representative multimodal discovery trace}

The initial candidate is a reproducible concatenation--MLP joint predictor with joint input fusion, a shared response program, a joint readout, response loss, and AdamW optimization; it obtains Fold-3 Global PCC $0.2668$. The recorded decision context for the retained final revision is: \emph{``Make one final diagnosis-driven revision of the best incumbent; prioritize a complete, executable perturbation-response hypothesis over a hyperparameter-only change.''} The selected multi-head cross-modal-attention candidate uses separate control, compound, and dose encoders, cross-modal attention fusion, a shared response program, joint readout, block-weighted MSE, AdamW, and cosine scheduling, reaching $0.2928$. The trace contains 11 documented provider calls (53,096 reported tokens), each with a request hash, response hash, status, and token record.

\begin{table}[!htbp]
\centering
\caption{Complete candidate record for the preselected representative search trajectory. Selection-fold PCC is shown for every completed candidate; repairs are candidate-code repair attempts. The selected candidate is bold.}
\label{tab:case-typical}
\tblsetup
\begin{tabularx}{\textwidth}{@{}C{20pt}X R{55pt}C{31pt}@{}}
\toprule
\rowcolor{csHeader}
Slot & Recorded revision design & \tblhdr{Global PCC\\$\uparrow$} & Repairs \\
\midrule
$h_0$ & Initial concatenation--MLP joint predictor & 0.2668 & 0 \\
1 & Chemical-query/control-key-value cross-modal attention with dose residual & 0.2903 & 0 \\
2 & FiLM-style chemical-and-dose modulation & 0.2742 & 0 \\
3 & Learned chemical--control gating & 0.2865 & 0 \\
4 & Low-rank bilinear chemical--control fusion & 0.2355 & 0 \\
5 & Multi-tower residual fusion with correlation-balanced MSE & 0.2812 & 0 \\
6 & Conditional low-rank transformation of control representation & 0.2733 & 0 \\
7 & Gated bilinear bottleneck fusion & 0.2682 & 1 \\
8 & Chemical-and-dose hypernetwork fusion & 0.2787 & 1 \\
 9 & \textbf{Multi-head cross-modal attention revision} & \textbf{0.2928} & 0 \\
\bottomrule
\end{tabularx}
\end{table}

\FloatBarrier
\subsection{Repair-containing discovery trace}

The second trace begins from the same initial program at Fold-3 Global PCC $0.2881$. Its structured agenda specifies separate control, compound, and dose encoders; compound-query/control-key-value interaction with a dose residual; a shared response program; joint response prediction; CP/L1000 block-weighted MSE (0.4/0.6); AdamW; and cosine scheduling. Its selected first revision implements this design, resolves an output-shape mismatch by returning a $[\mathrm{batch},\mathrm{target}]$ tensor, and reaches $0.3230$. The complete trace contains 13 provider calls (63,572 reported tokens); a later slot exhausts all three repair attempts and remains recorded as a failed slot.

\begin{table}[!htbp]
\centering
\caption{Complete candidate record for the preselected search trajectory containing a repair failure. The failed eighth slot has no evaluation score because all three allowed repair attempts were exhausted.}
\label{tab:case-repair}
\tblsetup
\begin{tabularx}{\textwidth}{@{}C{20pt}X R{55pt}C{31pt}@{}}
\toprule
\rowcolor{csHeader}
Slot & Recorded revision design & \tblhdr{Global PCC\\$\uparrow$} & Repairs \\
\midrule
$h_0$ & Initial concatenation--MLP joint predictor & 0.2881 & 0 \\
 1 & \textbf{Cross-modal attention, dose residual, block-weighted MSE} & \textbf{0.3230} & 1 \\
2 & FiLM-style gated perturbation modulation & 0.3065 & 0 \\
3 & Factorized response heads with cross-modal gate & 0.2894 & 0 \\
4 & Low-rank bilinear fusion with dose gate & 0.2837 & 0 \\
5 & Multi-tower residual response program & 0.3223 & 0 \\
6 & Low-rank joint readout bottleneck & 0.3165 & 0 \\
7 & Multi-tower residual response program & 0.3227 & 0 \\
8 & Failed candidate-code attempt & -- & 3 \\
9 & Multi-tower residual response program & 0.3217 & 0 \\
\bottomrule
\end{tabularx}
\end{table}

\section{Extended related work}
\label{app:related-work}

\paragraph{Cellular perturbation data and virtual-cell predictors.}
Cell Painting, L1000, Perturb-seq, and sci-Plex measure morphological and transcriptional responses across chemical and genetic interventions \citep{bray2016protocol,caicedo2017profiling,subramanian2017l1000,dixit2016perturbseq,srivatsan2020sciplex}. Matched resources and harmonized collections broaden this coverage \citep{haghighi2022profiles,chandrasekaran2024jump,peidli2024scperturb}. Predictors use latent-state transfer, compositional models, neural transport, genetic graphs, and foundation models \citep{lotfollahi2019scgen,lotfollahi2023cpa,bunne2023cellot,roohani2024gears,cui2024scgpt}; cycleCDR adds cycle-consistency constraints to learn transferable perturbation representations \citep{huang2024cyclecdr}. CIPHER combines unperturbed-cell covariance with the specified perturbation to predict responses, illustrating the predictive value of baseline cellular structure \citep{kuznets2025cipher}. \method asks whether a selected predictor uses the particular perturbation input invoked by its design.

\paragraph{Evaluation of perturbation prediction.}
PerturbBench and PertEval-scFM standardize predictive comparisons across splits and baselines \citep{wu2025perturbench,wenteler2025perteval}. Deep predictors need not outperform linear references, and common metrics can reward systematic variation shared across perturbations \citep{ahlmanneltze2025linear,vinas2026systema}. Other studies show that well-calibrated predictive metrics and informative foundation-model representations reveal gains over simple baselines \citep{miller2025calibrated,hasanaj2025multimodal,cole2026foundation}. In-the-wild evaluation further examines context and perturbation shifts \citep{mao2026virtualcellbenchmark}. Together, these studies establish the importance of metrics, representations, and transfer conditions. \method tests a complementary property: whether registered input-use claims are supported from cited source computation through fitted dependence to target-relevant predictive contribution.

\paragraph{Agentic model discovery.}
Scientific agents span laboratory workflows, program search, and iterative machine-learning experimentation \citep{boiko2023coscientist,romeraparedes2024funsearch,huang2024mlagentbench,li2024automated,lu2026ai,jiang2025aide}. CellScientist, CellForge, and HarmonyCell apply related search procedures to cellular prediction \citep{li2026cellscientist,tang2025cellforge,huang2026harmonycell}. The existing \discoverymethod policy supplies candidate models in these experiments; \method links their registered input-use claims to deterministic tests of source consumption, fitted dependence, and target-relevant predictive contribution.

\paragraph{Reliance, shortcuts, and automated falsification.}
Permutation reliance measures performance changes after input disruption; conditional permutations account for dependence among inputs \citep{fisher2019modelreliance,chamma2023conditional}. Shortcut and underspecification studies show that benchmark success can coexist with unintended or unstable decision rules \citep{damour2022underspecification,geirhos2020shortcut,lapuschkin2019clever,degrave2021shortcuts}. Attribution sanity checks, removal-based evaluation, and behavioral suites test explanations and capabilities under targeted interventions \citep{adebayo2018sanity,hooker2019roar,ribeiro2020checklist}. ConceptSMILE audits concept-explanation reliability through input perturbations and local surrogate modeling \citep{mollapour2026conceptsmile}; POPPER tests free-form hypotheses through agentic sequential falsification with statistical error control \citep{huang2025popper}. \method follows one selected cellular-response model and its registered input-use claim from the cited code to complete fitted predictions and later-data evaluation. This links source implementation, fitted dependence, and target-loss changes under registered replacements, distinguishing a blocked pathway, sensitivity without predictive benefit, and a contribution that recurs on new data.

\section{Factorial contrasts and stability summaries}
\label{app:factorial-definition}
An explicit input route establishes a possible computation; its fitted contribution is tested with the checkpoint and observed targets held fixed. We adapt permutation reliance and behavioral testing to the registered biological inputs \citep{fisher2019modelreliance,ribeiro2020checklist}. For a chemical-response model, the $2\times2$ test combines the correct or shuffled perturbation with the correct or an alternative control profile. Let $m_{pc}$ denote the response score with both correct inputs, $m_{\tilde p c}$ the score after a matched perturbation shuffle, $m_{p\tilde c}$ the score after a control-profile swap, and $m_{\tilde p\tilde c}$ the score after both. For a higher-is-better metric $m$,
\begin{align}
E_{\mathrm{pert}} &= \tfrac12[(m_{pc}-m_{\tilde p c})+(m_{p\tilde c}-m_{\tilde p\tilde c})],\\
E_{\mathrm{context}} &= \tfrac12[(m_{pc}-m_{p\tilde c})+(m_{\tilde p c}-m_{\tilde p\tilde c})],\\
I_{\mathrm{pert,context}} &= m_{pc}-m_{\tilde p c}-m_{p\tilde c}+m_{\tilde p\tilde c}.
\end{align}
These are factorial contrasts in predictive score under the registered replacement distribution: $E_{\mathrm{pert}}$ averages perturbation-replacement effects across the two tested contexts, while $I_{\mathrm{pert,context}}$ measures their nonadditivity in score. For compound inputs we write $E_{\rm cmpd}$ and $E_{\rm ctrl}$ in tables. The loss contrast $E^{\mathcal L}_{\rm cmpd}$ similarly averages the increase after compound replacement over both context states. Each checkpoint uses 32 fixed permutations constructed without response values or model scores. Continuous effects and their intervals are the primary quantitative evidence. For a per-model stability summary, let $\tau$ be the 95th percentile of absolute pairwise differences among the corresponding reference scores. A contribution is \emph{qualified} if its effect's empirical 2.5th percentile exceeds $\tau$, \emph{unsupported} if its 97.5th percentile does not exceed $\tau$, and \emph{inconclusive} otherwise. These labels summarize effect stability relative to map-induced score variation; an unsupported effect may still be positive. The empirical quantiles describe replacement variability within a checkpoint; Appendix~\ref{app:protocol} specifies the reference scores and separate across-model uncertainty.

\paragraph{Interpreting the reference-relative criterion.}
Reference-score variation can include variation induced by the other input. Holding the compound effect fixed while increasing this variation can therefore change the status. An existing synthetic check separates invariant, sensitive-without-benefit, target-relevant, and harmful predictors. All 96 target-relevant instances have positive mean effects, while 32 of 96 exceed the reference-relative criterion, identically with 32, 128, or 512 maps. The continuous measurements recover predictive benefit, whereas the categorical rule asks whether that benefit exceeds the specified reference variation. Its threshold is an operational effect scale. The additional 6,400-dataset study in Appendix~\ref{app:measurement-validity} tests estimator bias, interval coverage, and sensitivity to the reference scale; it distinguishes conditional replacement uncertainty from population uncertainty and retains the observed finite-group undercoverage.

% Appendix presentation derived from ../generated_existing_audit_checks.tex
% Generated by scripts/render_existing_audit_checks.py; edit the renderer, not this file.
% Numerical source: results/existing_audit_checks/source.json.
\section{Observed-support replay and matched reference checks}
\label{app:existing-audit-checks}
\subsection{Single-coordinate replay on observed support}
\label{app:observed-support-replay}
The frozen replay changes one encoded coordinate while holding the source row's other inputs fixed. Eligible donors satisfy the recorded metadata constraints and produce an actual input change. Table~\ref{tab:observed-support-replay} reports the supported populations separately from the original context-averaged factorial summaries. Score-selected BBBC047 remains exactly invariant; the path-constrained and guided panels retain positive compound effects on both boundaries. On LKCP, dose intervals are positive and compound intervals include zero.
For the replay and real-source audit below, 95\% intervals resample source groups while conditioning on the selected checkpoint set, donor pools, and 128 replacement maps; BBBC and LINCS use Murcko source scaffolds, and LKCP uses registered BRD identifier groups. The matched-reference and control-bank intervals instead describe paired training-seed variation at the fixed task split. Individual intervals and positive-interval counts are descriptive and unadjusted.

\begin{table}[!htbp]
\centering
\caption{Observed-support, fixed-context single-coordinate effects. Positive PCC drop and target-loss increase indicate benefit from the original input relative to its replacement. Cells report mean and 95\% source-group interval.}
\label{tab:observed-support-replay}
\tblsetup
\begin{tabular*}{\textwidth}{@{\extracolsep{\fill}}lllrrr@{}}
\toprule
\rowcolor{csHeader}
Panel & Fold & Input & \tblhdr{Supported/all rows\\source groups} & \tblhdr{PCC drop\\$\Delta m_q$} & \tblhdr{Target-loss increase\\$\Delta\mathcal{L}_q$} \\
\midrule
BBBC047: score-selected & 4 & Compound & \tblstack{3341/4380}{838 groups} & \tblstack{$0$}{$[0,0]$} & \tblstack{$0$}{$[0,0]$} \\
BBBC047: score-selected & 5 & Compound & \tblstack{2927/3878}{855 groups} & \tblstack{$0$}{$[0,0]$} & \tblstack{$0$}{$[0,0]$} \\
BBBC047: path-constrained & 4 & Compound & \tblstack{3341/4380}{838 groups} & \tblstack{$0.01025$}{$[0.00654,0.01425]$} & \tblstack{$3.13\!\times\!10^{-4}$}{$[1.95,4.59]\!\times\!10^{-4}$} \\
BBBC047: path-constrained & 5 & Compound & \tblstack{2927/3878}{855 groups} & \tblstack{$0.00654$}{$[0.00236,0.01097]$} & \tblstack{$2.08\!\times\!10^{-4}$}{$[0.75,3.51]\!\times\!10^{-4}$} \\
BBBC047: guided & 4 & Compound & \tblstack{3341/4380}{838 groups} & \tblstack{$0.00534$}{$[0.00336,0.00770]$} & \tblstack{$1.84\!\times\!10^{-4}$}{$[1.14,2.73]\!\times\!10^{-4}$} \\
BBBC047: guided & 5 & Compound & \tblstack{2927/3878}{855 groups} & \tblstack{$0.00404$}{$[0.00179,0.00634]$} & \tblstack{$1.44\!\times\!10^{-4}$}{$[0.64,2.28]\!\times\!10^{-4}$} \\
LINCS: guided & 4 & Compound & \tblstack{843/847}{219 groups} & \tblstack{$0.01996$}{$[0.01276,0.02863]$} & \tblstack{$5.95\!\times\!10^{-3}$}{$[3.71,8.80]\!\times\!10^{-3}$} \\
LINCS: guided & 4 & Dose & \tblstack{847/847}{221 groups} & \tblstack{$0.05096$}{$[0.04419,0.05778]$} & \tblstack{$1.52\!\times\!10^{-2}$}{$[1.28,1.77]\!\times\!10^{-2}$} \\
LINCS: guided & 5 & Compound & \tblstack{864/868}{235 groups} & \tblstack{$0.01446$}{$[0.00816,0.02113]$} & \tblstack{$3.83\!\times\!10^{-3}$}{$[2.22,5.64]\!\times\!10^{-3}$} \\
LINCS: guided & 5 & Dose & \tblstack{867/868}{236 groups} & \tblstack{$0.04853$}{$[0.04158,0.05586]$} & \tblstack{$1.28\!\times\!10^{-2}$}{$[1.08,1.50]\!\times\!10^{-2}$} \\
LKCP: guided & 4 & Compound & \tblstack{351/405}{13 groups} & \tblstack{$0.00043$}{$[-0.00149,0.00239]$} & \tblstack{$1.19\!\times\!10^{-3}$}{$[-4.18,6.24]\!\times\!10^{-3}$} \\
LKCP: guided & 4 & Dose & \tblstack{405/405}{15 groups} & \tblstack{$0.02407$}{$[0.01154,0.04369]$} & \tblstack{$6.61\!\times\!10^{-2}$}{$[3.14,11.41]\!\times\!10^{-2}$} \\
LKCP: guided & 5 & Compound & \tblstack{324/432}{12 groups} & \tblstack{$0.00321$}{$[-0.00049,0.00728]$} & \tblstack{$1.19\!\times\!10^{-2}$}{$[-0.40,3.27]\!\times\!10^{-2}$} \\
LKCP: guided & 5 & Dose & \tblstack{432/432}{16 groups} & \tblstack{$0.03537$}{$[0.02124,0.04906]$} & \tblstack{$1.21\!\times\!10^{-1}$}{$[0.50,2.14]\!\times\!10^{-1}$} \\
\bottomrule
\end{tabular*}
\tblnote{Original-entry-weighted means: score-selected BBBC047 has 5 entries per fold; path-constrained and guided BBBC047 have 10 each; LINCS and LKCP have 50 each. LKCP's 50 entries contain 30 unique weight hashes. Path-constrained and guided checkpoints are fixed across the two boundary evaluations. BBBC047 has no registered observed-support dose contrast.}
\end{table}

\FloatBarrier
\subsection{Complete matched input-reference family}
\label{app:matched-input-references}
The four input subsets use the same masked-MLP family, loss, fit budget, and five paired training seeds. Each selected checkpoint is evaluated on both held-out folds. Here $c$ denotes context, $p$ compound representation, and $a$ dose. Dose improves the LINCS reference; its increment is small on BBBC and LKCP. On LKCP Fold 5, the full reference has a positive PCC increment over $g(c,a)$, while its MSE interval includes zero. These trained-subset comparisons describe the fixed reference family; within-checkpoint replacement measures use in the separately selected agent models.

\begin{table}[!htbp]
\centering
\caption{Complete four-input masked-MLP references. Each cell gives mean Global PCC (top; higher is better) and MSE (bottom; lower is better) across five paired seeds.}
\label{tab:matched-input-family}
\tblsetup
\begin{tabular*}{\textwidth}{@{\extracolsep{\fill}}llrrrr@{}}
\toprule
\rowcolor{csHeader}
Cohort & Fold & $g(c)$ & $g(c,a)$ & $g(c,p)$ & $g(c,p,a)$ \\
\midrule
BBBC036 & 4 & \tblstack{0.3112}{0.058572} & \tblstack{0.3108}{0.058582} & \tblstack{0.2981}{0.059153} & \tblstack{0.2981}{0.059154} \\
BBBC036 & 5 & \tblstack{0.2928}{0.059025} & \tblstack{0.2922}{0.059035} & \tblstack{0.2708}{0.059966} & \tblstack{0.2707}{0.059968} \\
BBBC047 & 4 & \tblstack{0.3033}{0.045777} & \tblstack{0.3031}{0.045781} & \tblstack{0.2736}{0.046620} & \tblstack{0.2737}{0.046619} \\
BBBC047 & 5 & \tblstack{0.3145}{0.045252} & \tblstack{0.3144}{0.045257} & \tblstack{0.2840}{0.046161} & \tblstack{0.2840}{0.046159} \\
LINCS & 4 & \tblstack{0.1702}{0.731582} & \tblstack{0.1912}{0.725847} & \tblstack{0.1636}{0.733309} & \tblstack{0.1643}{0.733138} \\
LINCS & 5 & \tblstack{0.1721}{0.590465} & \tblstack{0.1917}{0.586218} & \tblstack{0.1631}{0.593010} & \tblstack{0.1638}{0.592862} \\
LKCP & 4 & \tblstack{0.1855}{2.849319} & \tblstack{0.1856}{2.850595} & \tblstack{0.2031}{2.998213} & \tblstack{0.2081}{3.020705} \\
LKCP & 5 & \tblstack{0.2050}{4.550455} & \tblstack{0.2056}{4.548941} & \tblstack{0.2525}{4.510473} & \tblstack{0.2420}{4.536487} \\
\bottomrule
\end{tabular*}
\tblnote{Paired increments and their 95\% intervals appear in Table~\ref{tab:matched-input-increments}; the portable source preserves all per-reference and paired intervals at full precision.}
\end{table}

\begin{table}[!htbp]
\centering
\caption{Paired input-reference increments (mean and 95\% training-seed interval). Differences are the first predictor minus the second; positive PCC and negative MSE favor the first predictor.}
\label{tab:matched-input-increments}
\tblsetup
\begin{tabular*}{\textwidth}{@{\extracolsep{\fill}}llrrrr@{}}
\toprule
\rowcolor{csHeader}
Cohort & Fold & \tblhdr{$g(c,a)-g(c)$\\PCC} & \tblhdr{$g(c,a)-g(c)$\\MSE} & \tblhdr{$g(c,p,a)-g(c,a)$\\PCC} & \tblhdr{$g(c,p,a)-g(c,a)$\\MSE} \\
\midrule
BBBC036 & 4 & \tblstack{$-0.00035$}{$[-0.00127,0.00058]$} & \tblstack{$1.02\!\times\!10^{-5}$}{$[-3.25,5.30]\!\times\!10^{-5}$} & \tblstack{$-0.01279$}{$[-0.01490,-0.01068]$} & \tblstack{$5.71\!\times\!10^{-4}$}{$[4.74,6.69]\!\times\!10^{-4}$} \\
BBBC036 & 5 & \tblstack{$-0.00052$}{$[-0.00115,0.00011]$} & \tblstack{$9.87\!\times\!10^{-6}$}{$[-1.47,21.21]\!\times\!10^{-6}$} & \tblstack{$-0.02150$}{$[-0.02258,-0.02041]$} & \tblstack{$9.33\!\times\!10^{-4}$}{$[8.51,10.15]\!\times\!10^{-4}$} \\
BBBC047 & 4 & \tblstack{$-0.00011$}{$[-0.00016,-0.00006]$} & \tblstack{$3.86\!\times\!10^{-6}$}{$[2.48,5.23]\!\times\!10^{-6}$} & \tblstack{$-0.02948$}{$[-0.03342,-0.02553]$} & \tblstack{$8.38\!\times\!10^{-4}$}{$[7.45,9.31]\!\times\!10^{-4}$} \\
BBBC047 & 5 & \tblstack{$-0.00012$}{$[-0.00016,-0.00008]$} & \tblstack{$4.89\!\times\!10^{-6}$}{$[3.16,6.62]\!\times\!10^{-6}$} & \tblstack{$-0.03039$}{$[-0.03351,-0.02726]$} & \tblstack{$9.03\!\times\!10^{-4}$}{$[8.03,10.02]\!\times\!10^{-4}$} \\
LINCS & 4 & \tblstack{$0.02095$}{$[0.02026,0.02164]$} & \tblstack{$-5.73\!\times\!10^{-3}$}{$[-5.95,-5.52]\!\times\!10^{-3}$} & \tblstack{$-0.02687$}{$[-0.02779,-0.02595]$} & \tblstack{$7.29\!\times\!10^{-3}$}{$[7.05,7.54]\!\times\!10^{-3}$} \\
LINCS & 5 & \tblstack{$0.01954$}{$[0.01899,0.02009]$} & \tblstack{$-4.25\!\times\!10^{-3}$}{$[-4.33,-4.16]\!\times\!10^{-3}$} & \tblstack{$-0.02790$}{$[-0.02887,-0.02693]$} & \tblstack{$6.64\!\times\!10^{-3}$}{$[6.35,6.93]\!\times\!10^{-3}$} \\
LKCP & 4 & \tblstack{$0.00004$}{$[-0.00050,0.00057]$} & \tblstack{$1.28\!\times\!10^{-3}$}{$[-1.88,4.44]\!\times\!10^{-3}$} & \tblstack{$0.02256$}{$[-0.00440,0.04952]$} & \tblstack{$1.70\!\times\!10^{-1}$}{$[-0.70,4.10]\!\times\!10^{-1}$} \\
LKCP & 5 & \tblstack{$0.00061$}{$[0.00016,0.00107]$} & \tblstack{$-1.51\!\times\!10^{-3}$}{$[-2.53,-0.49]\!\times\!10^{-3}$} & \tblstack{$0.03637$}{$[0.00366,0.06908]$} & \tblstack{$-1.25\!\times\!10^{-2}$}{$[-10.17,7.68]\!\times\!10^{-2}$} \\
\bottomrule
\end{tabular*}
\end{table}

\FloatBarrier
\subsection{Linked source and behavior in real candidates}
\label{app:real-source-behavior}
A fixed, dataset--policy-balanced and source-structure-stratified sample links 48 source records to 96 same-checkpoint boundary evaluations. Implemented source consumption coexists with distinct fitted outcomes: an implemented \texttt{MultiTowerFusion} source is exactly invariant on both boundaries, whereas 47 sources change predictions and 20 have positive target-loss intervals on both boundaries. Unresolved source locations also admit complete-model behavioral tests.

\begin{table}[!htbp]
\centering
\caption{Descriptive source--behavior counts in 48 real candidates. Each case contributes its paired boundary pattern once.}
\label{tab:real-source-behavior-counts}
\tblsetup
\begin{tabular*}{\textwidth}{@{\extracolsep{\fill}}llrrrr@{}}
\toprule
\rowcolor{csHeader}
Cohort & Source status & Cases & \tblhdr{Invariant\\both folds} & \tblhdr{Sensitive; loss CI not\\positive on $\geq 1$ fold} & \tblhdr{Positive loss CI\\both folds} \\
\midrule
BBBC036 & Implemented & 22 & 0 & 22 & 0 \\
BBBC036 & Unresolved & 2 & 0 & 2 & 0 \\
BBBC047 & Implemented & 21 & 1 & 1 & 19 \\
BBBC047 & Unresolved & 3 & 0 & 2 & 1 \\
\midrule
Total &  & 48 & 1 & 27 & 20 \\
\bottomrule
\end{tabular*}
\tblnote{The middle behavioral category has positive prediction distance but lacks a positive target-loss interval on at least one fold. Source status describes the cited computation. Counts describe this structurally sampled set and its scripted source classifications. Claim origin remains \texttt{task\_contract} where no candidate-authored manifest was emitted.}
\end{table}

\subsection{Control-reference reuse on sci-Plex}
\label{app:sciplex-control-reuse}
The existing control-bank comparison holds the response target $Y-R$ fixed and changes whether the context input reuses reference bank $R$ or uses the other bank. Reference A/B assignment is fixed before fitting and counterbalanced within cell-line, replicate, and control-layout strata (24 plate groups per orientation). Three input sets, two bank assignments, and five paired seeds give 30 fitted checkpoints. All six shared-minus-disjoint PCC intervals include zero. This comparison measures control-reference reuse on sci-Plex.

\begin{table}[!htbp]
\centering
\caption{sci-Plex control-reference reuse: shared-minus-disjoint Global PCC (mean and paired-seed 95\% interval), with identical centered targets in both arms.}
\label{tab:sciplex-control-reuse}
\tblsetup
\begin{tabular*}{\textwidth}{@{\extracolsep{\fill}}lrr@{}}
\toprule
\rowcolor{csHeader}
Predictor & Fold 4 & Fold 5 \\
\midrule
$g(c)$ & \tblstack{$0.00092$}{$[-0.00215,0.00399]$} & \tblstack{$0.00053$}{$[-0.00136,0.00241]$} \\
$g(c,a)$ & \tblstack{$-0.00030$}{$[-0.00336,0.00277]$} & \tblstack{$-0.00006$}{$[-0.00473,0.00462]$} \\
$g(c,p,a)$ & \tblstack{$0.00511$}{$[-0.00241,0.01263]$} & \tblstack{$0.00402$}{$[-0.01467,0.02271]$} \\
\bottomrule
\end{tabular*}
\end{table}

\FloatBarrier

\section{Matched feedback: predictive outcomes and input contribution}
% Appendix presentation derived from ../generated_feedback_prediction.tex
\begin{table}[!htbp]
\centering
\caption{Frozen predictors after score or audit feedback on sci-Plex. Values are mean $\pm$ trajectory SD ($n=5$).}
\label{tab:feedback-prediction}
\tblsetup
\begin{lrbox}{\AppendixTableBox}
\begin{tabular}{@{}L{88pt}R{70pt}R{70pt}R{70pt}R{70pt}@{}}
\toprule
\rowcolor{csHeader}
\tblhdr{Predictor} & \tblhdr{F4 PCC $\uparrow$} & \tblhdr{F4 MSE $\downarrow$} & \tblhdr{F5 PCC $\uparrow$} & \tblhdr{F5 MSE $\downarrow$} \\
\midrule
Anchor $g(c,a)$ & \tblpm{0.9664}{0.0001} & \tblpm{0.0308}{0.0001} & \tblpm{0.9728}{0.0002} & \tblpm{0.0251}{0.0001} \\
Score & \tblpm{0.9756}{0.0027} & \tblpm{0.0225}{0.0025} & \tblpm{0.9764}{0.0015} & \tblpm{0.0219}{0.0013} \\

Audit & \tblpm{0.9779}{0.0004} & \tblpm{0.0204}{0.0004} & \tblpm{0.9778}{0.0002} & \tblpm{0.0205}{0.0002} \\
\bottomrule
\end{tabular}
\end{lrbox}
\AppendixTableOutput{generated_feedback_prediction-1}
\tblnote{The anchor is shared within each seed. F4 and F5 are held-out combination boundaries of the sci-Plex development source.}
\end{table}

% Appendix presentation derived from ../generated_feedback_input_use.tex
\begin{table}[!htbp]
\centering
\caption{Target-relevant input use after feedback. Mean $\pm$ SD across five endpoints.}
\label{tab:feedback-input-use}
\tblsetup
\begin{lrbox}{\AppendixTableBox}
\begin{tabular}{@{}L{16pt}L{34pt}R{73pt}R{73pt}R{73pt}R{73pt}@{}}
\toprule
\rowcolor{csHeader}
\tblhdr{Set} & \tblhdr{Feedback} & \tblhdr{Chemical\\PCC drop} & \tblhdr{Chemical\\loss gain} & \tblhdr{Dose\\PCC drop} & \tblhdr{Dose\\loss gain} \\
\midrule
F4 & Score & \tblpm{0.0203}{0.0046} & \tblpm{0.0186}{0.0042} & \tblpm{0.0068}{0.0026} & \tblpm{0.0062}{0.0024} \\

F4 & Audit & \tblpm{0.0240}{0.0016} & \tblpm{0.0220}{0.0015} & \tblpm{0.0090}{0.0006} & \tblpm{0.0082}{0.0006} \\
F5 & Score & \tblpm{0.0070}{0.0016} & \tblpm{0.0064}{0.0015} & \tblpm{0.0039}{0.0017} & \tblpm{0.0035}{0.0015} \\

F5 & Audit & \tblpm{0.0086}{0.0005} & \tblpm{0.0079}{0.0005} & \tblpm{0.0049}{0.0004} & \tblpm{0.0045}{0.0004} \\
\bottomrule
\end{tabular}
\end{lrbox}
\AppendixTableOutput{generated_feedback_input_use-1}
\tblnote{Positive values favor the correct input over its legal replacement. Each displayed coordinate metric is positive in 5/5 endpoints in both arms at both boundaries.}
\end{table}

% Appendix presentation derived from ../generated_feedback_statistics.tex
\begin{table}[!htbp]
\centering
\caption{Audit minus Score: paired predictive effects and two uncertainty summaries. Effect and interval entries are in $10^{-3}$ units.}
\label{tab:feedback-statistics}
\tblsetup
\begin{lrbox}{\AppendixTableBox}
\begin{tabular}{@{}L{20pt}L{30pt}R{39pt}R{108pt}R{108pt}R{24pt}@{}}
\toprule
\rowcolor{csHeader}
\tblhdr{Set} & \tblhdr{Metric} & \tblhdr{Mean $\Delta$} & \tblhdr{Paired $t$\\95\% CI} & \tblhdr{Joint bootstrap\\95\% CI} & \tblhdr{Wins} \\
\midrule
F4 & PCC & +2.364 & [-1.277, +6.005] & [+0.253, +5.619] & 4/5 \\
F4 & MSE & -2.159 & [-5.482, +1.164] & [-5.136, -0.234] & 4/5 \\
\midrule
F5 & PCC & +1.460 & [-0.496, +3.416] & [+0.202, +3.021] & 3/5 \\
F5 & MSE & -1.333 & [-3.116, +0.450] & [-2.748, -0.187] & 3/5 \\
\bottomrule
\end{tabular}
\end{lrbox}
\AppendixTableOutput{generated_feedback_statistics-1}
\tblnote{Descriptive paired comparison. Paired $t$: five trajectories, df=4. Joint bootstrap: 4,000 shared trajectory--compound draws. Wins favor higher PCC or lower MSE. Both uncertainty summaries retain all five trajectory pairs.}
\end{table}

% Appendix presentation derived from ../generated_feedback_costs.tex
\begin{table}[!htbp]
\centering
\caption{Discovery outcomes and realized resource use under matched candidate and training opportunities.}
\label{tab:feedback-costs}
\tblsetup
\begin{lrbox}{\AppendixTableBox}
\begin{tabular}{@{}L{37pt}R{52pt}R{32pt}R{24pt}R{30pt}R{30pt}R{61pt}R{48pt}@{}}
\toprule
\rowcolor{csHeader}
\tblhdr{Feedback} & \tblhdr{Selected\\PCC} & \tblhdr{Valid} & \tblhdr{Failed} & \tblhdr{Repairs} & \tblhdr{LLM\\calls} & \tblhdr{Reported\\tokens} & \tblhdr{Training\\wall s} \\
\midrule
Score & 0.9757 & 25/25 & 0 & 11 & 36 & $\geq$286,995 & 194.93 \\

Audit & 0.9781 & 25/25 & 0 & 13 & 38 & 592,347 & 217.41 \\
\bottomrule
\end{tabular}
\end{lrbox}
\AppendixTableOutput{generated_feedback_costs-1}
\tblnote{Selected PCC averages the five discovery-selection scores. Calls include repairs; the Score token total is a lower bound with one missing usage receipt. Five shared anchors contribute a further 32.61 training seconds, counted once.}
\end{table}

% Appendix presentation derived from ../generated_feedback_appendix.tex
\subsection{Score--audit feedback comparison: complete records}

\label{app:feedback-comparison}

\paragraph{Comparison and evaluation.} We analyze Score and Audit feedback using all five paired discovery trajectories and their shared context-plus-dose anchors. The descriptive comparison holds model space, trainer, selection rule, and candidate budget fixed, and reports every selected endpoint, paired effect, and uncertainty interval. Within the sci-Plex development source, compound--cell-line roles were reassigned and frozen before response preprocessing. F4 and F5 evaluate the same selected checkpoints on held-out combinations.

\paragraph{Data and splitting.} The fit, selection, F4, and F5 packages contain 2,666, 891, 438, and 442 treatment wells. Complete compound $\times$ cell-line groups, including their doses and replicate wells, remain within one role; evaluated compounds and cell lines each appear among the fit marginals. The 2,000 target genes are selected by variance among fit treatment wells, with fit-only feature scaling. Control bank A supplies inputs and bank B the reference; they contain distinct wells matched on plate, cell line, and replicate, with control pools reused across these development roles. The target is the complete gene-expression profile.

\paragraph{Feedback and model revision.} Score supplies selection metrics, sampled learning curves, model size, timing, and past designs. Audit adds grouped errors, component-source clues, and chemical/dose replacement effects. This comparison estimates the complete enhanced-feedback package. Both arms share $g(c,a)$ within seed, five new candidate slots, and three repair opportunities per slot. The language model may revise encoders, fusion, readout, capacity, regularization, and residual scale; a direct predictor or frozen-anchor residual is equally available. Each new component is trained from scratch. Completed selection diagnostics inform subsequent proposals; final checkpoints are frozen before F4/F5 evaluation.

\paragraph{Fixed training and selection.} AdamW uses learning rate $10^{-3}$, weight decay $10^{-5}$, batch size 128, gradient clipping at 5, a 100-epoch maximum, and early-stopping patience 15. The shared loss is fit-standardized target MSE plus $0.02(1-\mathrm{PCC})$, using flattened Pearson correlation. The common limit is five million trainable parameters. Epoch and endpoint selection prioritize selection Global PCC, then lower MSE; endpoint ties additionally prefer fewer parameters and earlier slots. The anchor stays frozen in residual models.

\paragraph{Uncertainty and continuous evidence.} SD uses denominator $n-1$ across five trajectories; paired two-sided Student-$t$ intervals use df=4. Compound-only bootstrap fixes the fitted trajectories, trajectory-only bootstrap fixes the evaluated compounds, and joint bootstrap resamples both; each uses 4,000 shared-index draws. Appendix~\ref{app:feedback-robustness} compares these intervals with complete leave-one-pair-out and exact sign-flip analyses. Maps are averaged within endpoints before trajectory inference. RMS measures sensitivity; PCC drop is correct-input minus replacement PCC, and loss gain is replacement minus correct-input MSE. Positive continuous effects are distinct from individual qualification or biological-mechanism evidence.

\paragraph{Model identifier and indexing.} Recorded model identifier: \texttt{deepseek-v41}. All compared calls use this serving identity. Trajectory indices 1--5 correspond, in order, to seeds 2026091201, 2026091202, 2026091203, 2026091204, 2026091205. Full-precision endpoint, cost, and paired-effect tables accompany the renderer under \texttt{results/feedback\_comparison/}.

\paragraph{Ratios and resources.} Relative MSE reduction against the anchor is computed within seed as $1-\mathrm{MSE}_{\mathrm{endpoint},s}/\mathrm{MSE}_{\mathrm{anchor},s}$ and then averaged. A ratio of arm mean MSEs is reported separately when used. Costs retain failed attempts and recorded interruptions. Shared anchor training totals 32.611546 seconds and is counted once, separately from each condition's candidate-training costs. Trainer and worker wall times overlap; missing token usage produces a lower bound.

\begin{table}[!htbp]
\centering
\caption{Every frozen endpoint at both held-out boundaries: all five paired trajectories and their shared anchors.}
\label{tab:feedback-endpoints-f4}
\label{tab:feedback-endpoints-f5}
\tblsetup

\begin{lrbox}{\AppendixTableBox}
\begin{tabular}{@{}R{12pt}L{27pt}R{13pt}R{33pt}R{33pt}R{33pt}R{33pt}R{33pt}R{33pt}R{33pt}R{33pt}@{}}
\toprule
\rowcolor{csHeader}
\tblhdr{Tr.} & \tblhdr{Model} & \tblhdr{Slot} & \tblhdr{PCC} & \tblhdr{MSE} & \tblhdr{C-R} & \tblhdr{C-P} & \tblhdr{C-L} & \tblhdr{D-R} & \tblhdr{D-P} & \tblhdr{D-L} \\
\midrule
\rowcolor{csHeader}\multicolumn{11}{@{}l}{\textbf{Fold 4}} \\
1 & Anchor & 0 & 0.96652 & 30.768 & 0.000 & 0.000 & 0.000 & 28.653 & 1.152 & 1.042 \\
1 & Score & 5 & 0.97557 & 22.551 & 141.301 & 20.580 & 18.805 & 84.343 & 6.798 & 6.212 \\
1 & Audit & 5 & 0.97806 & 20.278 & 142.171 & 23.162 & 21.173 & 94.176 & 8.838 & 8.081 \\
2 & Anchor & 0 & 0.96646 & 30.836 & 0.000 & 0.000 & 0.000 & 25.007 & 0.985 & 0.896 \\
2 & Score & 3 & 0.97740 & 20.889 & 136.127 & 21.774 & 19.882 & 84.750 & 7.580 & 6.923 \\
2 & Audit & 4 & 0.97809 & 20.252 & 155.045 & 25.294 & 23.142 & 101.848 & 9.453 & 8.650 \\
3 & Anchor & 0 & 0.96647 & 30.813 & 0.000 & 0.000 & 0.000 & 31.861 & 1.311 & 1.189 \\
3 & Score & 3 & 0.97570 & 22.432 & 138.732 & 20.644 & 18.823 & 96.003 & 7.982 & 7.278 \\
3 & Audit & 3 & 0.97817 & 20.181 & 138.099 & 22.698 & 20.763 & 91.844 & 8.613 & 7.880 \\
4 & Anchor & 0 & 0.96646 & 30.820 & 0.000 & 0.000 & 0.000 & 26.818 & 1.066 & 0.960 \\
4 & Score & 5 & 0.97807 & 20.274 & 157.142 & 25.704 & 23.535 & 96.377 & 9.070 & 8.307 \\
4 & Audit & 5 & 0.97725 & 21.023 & 145.497 & 22.784 & 20.858 & 90.603 & 8.193 & 7.502 \\
5 & Anchor & 0 & 0.96629 & 30.977 & 0.000 & 0.000 & 0.000 & 17.193 & 0.640 & 0.580 \\
5 & Score & 4 & 0.97115 & 26.587 & 117.806 & 13.009 & 11.887 & 48.269 & 2.392 & 2.186 \\
5 & Audit & 2 & 0.97815 & 20.203 & 159.745 & 26.108 & 23.941 & 102.759 & 9.697 & 8.893 \\
\addlinespace[3pt]
\rowcolor{csHeader}\multicolumn{11}{@{}l}{\textbf{Fold 5}} \\
1 & Anchor & 0 & 0.97265 & 25.253 & 0.000 & 0.000 & 0.000 & 27.811 & 0.712 & 0.647 \\
1 & Score & 5 & 0.97586 & 22.332 & 100.896 & 7.672 & 7.048 & 65.922 & 3.595 & 3.306 \\
1 & Audit & 5 & 0.97785 & 20.495 & 90.349 & 8.720 & 7.978 & 71.763 & 5.036 & 4.612 \\
2 & Anchor & 0 & 0.97297 & 24.955 & 0.000 & 0.000 & 0.000 & 24.562 & 0.606 & 0.553 \\
2 & Score & 3 & 0.97777 & 20.567 & 80.270 & 7.558 & 6.905 & 71.301 & 4.866 & 4.452 \\
2 & Audit & 4 & 0.97772 & 20.619 & 90.797 & 8.556 & 7.835 & 74.695 & 4.693 & 4.303 \\
3 & Anchor & 0 & 0.97297 & 24.957 & 0.000 & 0.000 & 0.000 & 31.737 & 0.839 & 0.764 \\
3 & Score & 3 & 0.97626 & 21.947 & 92.154 & 7.588 & 6.919 & 75.540 & 4.604 & 4.202 \\
3 & Audit & 3 & 0.97815 & 20.230 & 86.705 & 8.379 & 7.691 & 71.783 & 4.883 & 4.487 \\
4 & Anchor & 0 & 0.97270 & 25.230 & 0.000 & 0.000 & 0.000 & 26.502 & 0.688 & 0.621 \\
4 & Score & 5 & 0.97771 & 20.623 & 89.824 & 8.217 & 7.526 & 77.179 & 5.209 & 4.776 \\
4 & Audit & 5 & 0.97758 & 20.744 & 88.060 & 7.995 & 7.320 & 70.291 & 4.508 & 4.131 \\
5 & Anchor & 0 & 0.97292 & 24.999 & 0.000 & 0.000 & 0.000 & 17.108 & 0.399 & 0.363 \\
5 & Score & 4 & 0.97425 & 23.789 & 75.832 & 4.122 & 3.769 & 40.870 & 1.079 & 0.988 \\
5 & Audit & 2 & 0.97786 & 20.504 & 99.006 & 9.405 & 8.635 & 83.693 & 5.505 & 5.060 \\
\bottomrule
\end{tabular}
\end{lrbox}
\AppendixTableOutput{generated_feedback_appendix-1}

\tblnote{PCC is unscaled; all other numeric outcomes are in $10^{-3}$ units. C/D: chemical/dose; R: prediction RMS; P: PCC drop; L: target-loss gain. Slot 0 is the shared anchor. RMS has no preferred direction.}
\end{table}

\begin{table}[!htbp]
\centering
\caption{All ten discovery trajectories and their realized costs.}
\label{tab:feedback-all-costs}
\tblsetup

\begin{lrbox}{\AppendixTableBox}
\begin{tabular}{@{}R{12pt}L{28pt}R{15pt}R{34pt}R{24pt}R{24pt}R{24pt}R{50pt}R{34pt}R{34pt}R{34pt}@{}}
\toprule
\rowcolor{csHeader}
\tblhdr{Tr.} & \tblhdr{Arm} & \tblhdr{Slot} & \tblhdr{Sel. PCC} & \tblhdr{Repair} & \tblhdr{Calls} & \tblhdr{HTTP} & \tblhdr{Tokens} & \tblhdr{Train s} & \tblhdr{Work s} & \tblhdr{Diag. s} \\
\midrule
1 & Score & 5 & 0.97491 & 1 & 6 & 7 & $\geq$50,685 & 48.09 & 67.26 & 7.10 \\
1 & Audit & 5 & 0.97828 & 2 & 7 & 7 & 106,239 & 41.24 & 60.03 & 12.10 \\
2 & Score & 3 & 0.97757 & 3 & 8 & 8 & 62,703 & 41.96 & 60.07 & 7.50 \\
2 & Audit & 4 & 0.97783 & 3 & 8 & 8 & 125,047 & 42.32 & 61.01 & 12.94 \\
3 & Score & 3 & 0.97628 & 4 & 9 & 9 & 70,709 & 40.04 & 58.06 & 7.36 \\
3 & Audit & 3 & 0.97825 & 3 & 8 & 8 & 136,386 & 44.25 & 61.58 & 12.15 \\
4 & Score & 5 & 0.97852 & 3 & 8 & 8 & 68,757 & 35.40 & 52.75 & 8.73 \\
4 & Audit & 5 & 0.97798 & 3 & 8 & 8 & 129,227 & 39.60 & 55.93 & 12.24 \\
5 & Score & 4 & 0.97145 & 0 & 5 & 5 & 34,141 & 29.44 & 46.61 & 6.75 \\
5 & Audit & 2 & 0.97830 & 2 & 7 & 7 & 95,448 & 50.00 & 67.66 & 11.96 \\
\bottomrule
\end{tabular}
\end{lrbox}
\AppendixTableOutput{generated_feedback_appendix-2}
\tblnote{Every trajectory completed 5/5 new candidates and had zero fully failed slots. Train is known completed trainer wall time; Work is candidate worker wall time; Diag. is postfit diagnostic wall time. Logical calls and physical HTTP attempts are distinct. Complete token breakdowns and missing-cost flags are in the CSV.}
\end{table}

\begin{table}[!htbp]
\centering
\caption{All eight Audit--Score outcome contrasts at both boundaries. Effects and intervals are in $10^{-3}$ units.}
\label{tab:feedback-all-contrasts}
\tblsetup
\begin{lrbox}{\AppendixTableBox}
\begin{tabular}{@{}L{22pt}L{104pt}R{58pt}R{125pt}R{43pt}@{}}
\toprule
\rowcolor{csHeader}
\tblhdr{Set} & \tblhdr{Metric} & \tblhdr{Mean $\Delta$} & \tblhdr{Paired 95\% CI} & \tblhdr{+/--/0} \\
\midrule
F4 & PCC & +2.364 & [-1.277, +6.005] & 4/1/0 \\
F4 & MSE & -2.159 & [-5.482, +1.164] & 1/4/0 \\
F4 & Chem. RMS & +9.890 & [-16.189, +35.969] & 3/2/0 \\
F4 & Chem. PCC drop & +3.667 & [-3.577, +10.912] & 4/1/0 \\
F4 & Chem. loss gain & +3.389 & [-3.269, +10.048] & 4/1/0 \\
F4 & Dose RMS & +14.298 & [-16.033, +44.628] & 3/2/0 \\
F4 & Dose PCC drop & +2.194 & [-1.639, +6.028] & 4/1/0 \\
F4 & Dose loss gain & +2.020 & [-1.496, +5.536] & 4/1/0 \\
F5 & PCC & +1.460 & [-0.496, +3.416] & 3/2/0 \\
F5 & MSE & -1.333 & [-3.116, +0.450] & 2/3/0 \\
F5 & Chem. RMS & +3.189 & [-13.712, +20.089] & 2/3/0 \\
F5 & Chem. PCC drop & +1.580 & [-1.069, +4.228] & 4/1/0 \\
F5 & Chem. loss gain & +1.458 & [-0.979, +3.896] & 4/1/0 \\
F5 & Dose RMS & +8.283 & [-16.535, +33.100] & 3/2/0 \\
F5 & Dose PCC drop & +1.054 & [-1.483, +3.592] & 3/2/0 \\
F5 & Dose loss gain & +0.974 & [-1.355, +3.303] & 3/2/0 \\
\bottomrule
\end{tabular}
\end{lrbox}
\AppendixTableOutput{generated_feedback_appendix-3}
\tblnote{Descriptive effect estimates over all five paired trajectories. Signs count positive, negative, and exactly zero differences, not wins. Lower MSE is favorable; RMS has no favorable direction.}
\end{table}

\begin{table}[!htbp]
\centering
\caption{Predictive uncertainty under the three original sampling schemes; interval entries in $10^{-3}$ units.}
\label{tab:feedback-bootstrap}
\tblsetup
\begin{lrbox}{\AppendixTableBox}
\begin{tabular}{@{}L{18pt}L{28pt}R{106pt}R{106pt}R{106pt}@{}}
\toprule
\rowcolor{csHeader}
\tblhdr{Set} & \tblhdr{Metric} & \tblhdr{Paired $t$ CI} & \tblhdr{Compound-only CI} & \tblhdr{Joint CI} \\
\midrule
F4 & PCC & [-1.277, +6.005] & [+1.300, +3.704] & [+0.253, +5.619] \\
F4 & MSE & [-5.482, +1.164] & [-3.389, -1.197] & [-5.136, -0.234] \\
F5 & PCC & [-0.496, +3.416] & [+1.006, +1.975] & [+0.202, +3.021] \\
F5 & MSE & [-3.116, +0.450] & [-1.800, -0.924] & [-2.748, -0.187] \\
\bottomrule
\end{tabular}
\end{lrbox}
\AppendixTableOutput{generated_feedback_appendix-4}
\tblnote{Bootstrap intervals use 4,000 draws. No coordinate-effect bootstrap interval is substituted from predictive statistics, and no new confirmatory tests are introduced.}
\end{table}

\begin{table}[!htbp]
\centering
\caption{Legal replacement support shared by all compared endpoints.}
\label{tab:feedback-map-coverage}
\tblsetup
\begin{lrbox}{\AppendixTableBox}
\begin{tabular}{@{}L{24pt}L{67pt}R{66pt}R{57pt}R{62pt}R{62pt}@{}}
\toprule
\rowcolor{csHeader}
\tblhdr{Set} & \tblhdr{Coordinate} & \tblhdr{Eligible wells} & \tblhdr{Maps} & \tblhdr{Unique} & \tblhdr{First 32\\unique} \\
\midrule
F4 & Chemical & 438/438 & 128 & 128 & 32 \\
F4 & Dose & 428/438 & 128 & 1 & 1 \\
F5 & Chemical & 434/442 & 128 & 128 & 32 \\
F5 & Dose & 436/442 & 128 & 1 & 1 \\
\bottomrule
\end{tabular}
\end{lrbox}
\AppendixTableOutput{generated_feedback_appendix-5}
\tblnote{Replacements match plate, cell line, replicate, and treatment time. Chemical replacement holds dose fixed; dose replacement holds compound fixed and requires at least 0.1 log-dose difference. No cross-group fallback is used. Discovery used 32 maps; final evaluation used 128, with only one distinct dose map. Unsupported wells are excluded, not assigned zero effect.}
\end{table}

\FloatBarrier
\subsection{Robustness across the complete five trajectory pairs}
\label{app:feedback-robustness}
% Appendix presentation derived from ../generated_feedback_fivepair_appendix.tex
% Generated by tools/render_validation_evidence.py; do not edit by hand.
\paragraph{All five matched feedback trajectories.}
This retrospective sensitivity analysis retains seeds 2026091201--2026091205 and all eight metrics at both boundaries. Audit minus Score is computed within each selected-trajectory pair; F4 and F5 reuse the same selected checkpoints and do not make $N=10$. Means and descriptive two-sided 95\% $t_4$ intervals use all five pairs. All 32 sign assignments are enumerated for the absolute unstudentized mean under a pair-exchangeability/sign-symmetry null, not a randomized-assignment guarantee. The smallest attainable two-sided tail probability is $2/32=.0625$; no multiplicity-adjusted confirmatory claim is made.

\begin{table}[!htbp]
\centering
\caption{All five-pair Audit-minus-Score endpoints. Numerical effects and intervals are $\times10^3$.}
\label{tab:feedback-fivepair-seed}
\tblsetup
\begin{lrbox}{\AppendixTableBox}
\begin{tabular}{@{}C{16pt}L{83pt}R{43pt}R{104pt}R{43pt}C{35pt}R{30pt}@{}}
\toprule
\rowcolor{csHeader}
Fold & Metric & Mean & Paired-$t$ 95\% CI & Median & $+/-/0$ & Tail \\
\midrule
F4 & Global PCC & $2.364$ & $[-1.277,\,6.005]$ & $2.465$ & 4/1/0 & 6/32 \\
F4 & MSE & $-2.159$ & $[-5.482,\,1.164]$ & $-2.251$ & 1/4/0 & 6/32 \\
F4 & Chemical RMS & $9.890$ & $[-16.189,\,35.969]$ & $0.870$ & 3/2/0 & 12/32 \\
F4 & Chemical loss gain & $3.389$ & $[-3.269,\,10.048]$ & $2.368$ & 4/1/0 & 8/32 \\
F4 & Chemical PCC drop & $3.667$ & $[-3.577,\,10.912]$ & $2.582$ & 4/1/0 & 8/32 \\
F4 & Dose RMS & $14.298$ & $[-16.033,\,44.628]$ & $9.833$ & 3/2/0 & 10/32 \\
F4 & Dose loss gain & $2.020$ & $[-1.496,\,5.536]$ & $1.727$ & 4/1/0 & 6/32 \\
F4 & Dose PCC drop & $2.194$ & $[-1.639,\,6.028]$ & $1.873$ & 4/1/0 & 6/32 \\
F5 & Global PCC & $1.460$ & $[-0.496,\,3.416]$ & $1.888$ & 3/2/0 & 8/32 \\
F5 & MSE & $-1.333$ & $[-3.116,\,0.450]$ & $-1.716$ & 2/3/0 & 8/32 \\
F5 & Chemical RMS & $3.189$ & $[-13.712,\,20.089]$ & $-1.763$ & 2/3/0 & 24/32 \\
F5 & Chemical loss gain & $1.458$ & $[-0.979,\,3.896]$ & $0.930$ & 4/1/0 & 4/32 \\
F5 & Chemical PCC drop & $1.580$ & $[-1.069,\,4.228]$ & $0.999$ & 4/1/0 & 4/32 \\
F5 & Dose RMS & $8.283$ & $[-16.535,\,33.100]$ & $3.394$ & 3/2/0 & 20/32 \\
F5 & Dose loss gain & $0.974$ & $[-1.355,\,3.303]$ & $0.285$ & 3/2/0 & 12/32 \\
F5 & Dose PCC drop & $1.054$ & $[-1.483,\,3.592]$ & $0.279$ & 3/2/0 & 12/32 \\
\bottomrule
\end{tabular}
\end{lrbox}
\AppendixTableOutput{generated_feedback_fivepair_appendix-1}
\tblnote{Tail is the exact sign-flip count out of 32. All 16 paired-$t$ intervals contain zero and all exact tails are at least .125. RMS measures sensitivity; target-loss gain and PCC drop measure changes in loss and correlation.}
\end{table}

\begin{table}[!htbp]
\centering
\caption{All percentile-bootstrap 95\% intervals ($\times10^3$).}
\label{tab:feedback-fivepair-bootstrap}
\tblsetup
\begin{lrbox}{\AppendixTableBox}
\begin{tabular}{@{}C{16pt}L{80pt}R{91pt}R{91pt}R{91pt}@{}}
\toprule
\rowcolor{csHeader}
Fold & Metric & Compound only & Trajectory only & Joint \\
\midrule
F4 & Global PCC & $[1.300,\,3.704]$ & $[0.144,\,4.828]$ & $[0.253,\,5.619]$ \\
F4 & MSE & $[-3.389,\,-1.197]$ & $[-4.408,\,-0.132]$ & $[-5.136,\,-0.234]$ \\
F4 & Chemical RMS & $[6.425,\,13.160]$ & $[-4.738,\,28.126]$ & $[-5.652,\,27.694]$ \\
F4 & Chemical loss gain & $[2.032,\,5.060]$ & $[-0.482,\,8.094]$ & $[-0.456,\,8.402]$ \\
F4 & Chemical PCC drop & $[2.201,\,5.452]$ & $[-0.534,\,8.787]$ & $[-0.508,\,9.067]$ \\
F4 & Dose RMS & $[4.676,\,22.279]$ & $[-2.007,\,34.959]$ & $[-3.228,\,38.642]$ \\
F4 & Dose loss gain & $[0.689,\,3.753]$ & $[0.039,\,4.490]$ & $[0.066,\,5.631]$ \\
F4 & Dose PCC drop & $[0.750,\,4.083]$ & $[0.028,\,4.884]$ & $[0.068,\,6.187]$ \\
F5 & Global PCC & $[1.006,\,1.975]$ & $[0.302,\,2.618]$ & $[0.202,\,3.021]$ \\
F5 & MSE & $[-1.800,\,-0.924]$ & $[-2.392,\,-0.274]$ & $[-2.748,\,-0.187]$ \\
F5 & Chemical RMS & $[-1.073,\,7.471]$ & $[-6.751,\,14.920]$ & $[-7.875,\,15.416]$ \\
F5 & Chemical loss gain & $[0.824,\,2.180]$ & $[0.248,\,3.260]$ & $[0.143,\,3.537]$ \\
F5 & Chemical PCC drop & $[0.892,\,2.364]$ & $[0.266,\,3.528]$ & $[0.146,\,3.843]$ \\
F5 & Dose RMS & $[1.196,\,15.103]$ & $[-3.716,\,25.621]$ & $[-5.151,\,28.995]$ \\
F5 & Dose loss gain & $[0.324,\,1.866]$ & $[-0.261,\,2.557]$ & $[-0.253,\,3.122]$ \\
F5 & Dose PCC drop & $[0.349,\,2.021]$ & $[-0.294,\,2.768]$ & $[-0.284,\,3.397]$ \\
\bottomrule
\end{tabular}
\end{lrbox}
\AppendixTableOutput{generated_feedback_fivepair_appendix-2}
\tblnote{4000 draws, seed 69313. Compound weights are shared across all paired arms and trajectories; trajectory draws resample the five paired indices. Joint draws resample both units. The 128 donor maps are fixed. Sufficient statistics are pooled before PCC; compound PCCs are not averaged. Compound-only intervals condition on five selected trajectories; trajectory-only intervals condition on the observed compound set, while joint intervals vary both empirical units.}
\end{table}

\paragraph{Why inference methods disagree.}
Trajectory-only percentile bootstrap and paired-$t$ inference target the same conditional mean paired contrast; their different conclusions reflect the five-point empirical bootstrap distribution versus Student-$t$ sampling assumptions, not different estimands. Compound-only and joint intervals additionally have different resampling scopes. All predictive bootstrap intervals exclude zero, whereas the paired-$t$ intervals and exact sign-flip tails do not provide the same evidence. Increasing bootstrap draws does not add independent trajectories or establish broad generalization.

\begin{table}[!htbp]
\centering
\caption{Fifth-pair contribution and complete leave-one-pair-out mean ranges ($\times10^3$). No pair is removed from the primary summary.}
\label{tab:feedback-fifth-pair}
\tblsetup
\begin{lrbox}{\AppendixTableBox}
\begin{tabular}{@{}C{16pt}L{90pt}R{48pt}R{54pt}R{54pt}R{102pt}@{}}
\toprule
\rowcolor{csHeader}
Fold & Metric & All-five mean & \tblhdr{Fifth-pair\\$d_5/5$} & \tblhdr{Omit-fifth\\mean} & All five LOO means \\
\midrule
F4 & Global PCC & $2.364$ & $1.399$ & $1.206$ & $[1.206,\,3.160]$ \\
F4 & MSE & $-2.159$ & $-1.277$ & $-1.103$ & $[-2.886,\,-1.103]$ \\
F4 & Chemical RMS & $9.890$ & $8.388$ & $1.877$ & $[1.877,\,15.273]$ \\
F4 & Chemical loss gain & $3.389$ & $2.411$ & $1.223$ & $[1.223,\,4.906]$ \\
F4 & Chemical PCC drop & $3.667$ & $2.620$ & $1.309$ & $[1.309,\,5.314]$ \\
F4 & Dose RMS & $14.298$ & $10.898$ & $4.249$ & $[4.249,\,19.315]$ \\
F4 & Dose loss gain & $2.020$ & $1.341$ & $0.848$ & $[0.848,\,2.726]$ \\
F4 & Dose PCC drop & $2.194$ & $1.461$ & $0.916$ & $[0.916,\,2.962]$ \\
F5 & Global PCC & $1.460$ & $0.722$ & $0.922$ & $[0.922,\,1.859]$ \\
F5 & MSE & $-1.333$ & $-0.657$ & $-0.845$ & $[-1.697,\,-0.845]$ \\
F5 & Chemical RMS & $3.189$ & $4.635$ & $-1.808$ & $[-1.808,\,6.622]$ \\
F5 & Chemical loss gain & $1.458$ & $0.973$ & $0.606$ & $[0.606,\,1.875]$ \\
F5 & Chemical PCC drop & $1.580$ & $1.057$ & $0.654$ & $[0.654,\,2.030]$ \\
F5 & Dose RMS & $8.283$ & $8.565$ & $-0.352$ & $[-0.352,\,12.075]$ \\
F5 & Dose loss gain & $0.974$ & $0.815$ & $0.199$ & $[0.199,\,1.379]$ \\
F5 & Dose PCC drop & $1.054$ & $0.885$ & $0.211$ & $[0.211,\,1.493]$ \\
\bottomrule
\end{tabular}
\end{lrbox}
\AppendixTableOutput{generated_feedback_fivepair_appendix-3}
\tblnote{The fifth-pair column is its contribution to the five-pair mean, not its whole paired difference. Every leave-one-out estimate and $t_3$ interval is retained in the portable CSV. F5 chemical and dose RMS have positive all-five means but negative omit-fifth means; the F5 chemical-RMS median is negative, whereas the dose-RMS median is positive. All five pairs remain in the primary estimate.}
\end{table}

\paragraph{Verification boundary.}
Retained full predictions independently verify global PCC/MSE and eligible-source correct statistics. Full counterfactual prediction arrays were not retained for this legacy feedback package, so coordinate reconstruction verifies stored sufficient statistics but is not a new independent full-array counterfactual check. These additional descriptions do not reselect checkpoints, retrain models, or revise historical qualification labels.

% Appendix presentation derived from ../generated_feedback_case_appendix.tex
\section{Recorded feedback, source revisions, and frozen endpoints}
\label{app:feedback-case}
The two cases are the first two registered trajectory seeds (2026091201 and 2026091202), selected by registration order. Each retains the common initial predictor and all five new audit-feedback proposals. Parent links denote design ancestry: every new candidate is trained from scratch, while the shared context-plus-dose predictor $g(c,a)$ is frozen. This anchor concatenates 2,000 context features with dose and uses a 2001--256--256--2000 MLP with 1,092,816 trainable parameters. Local metrics describe the development-selection partition. The public case package contains the hash-verified anchor source, actual system/user prompts, proposals, candidate source, diffs, and received feedback.
\begin{figure}[!t]
\centering
\input{figures/case_panels}
\caption{\textbf{A recorded revision from received feedback to frozen-model evaluation.} The first registered trajectory supplies (a) all parent links and selection scores, (b) the slot-2 curve and best epoch received before slot 4, and (c) verified source edits, parameter counts and selection MSE; shading marks changed entries. (d) All five new candidates: arrows show ancestry, marker area scales with trainable parameters, and the diamond denotes selected slot 5. (e) The matched anchor, Score and Audit checkpoints on Fold 4 (open circles) and Fold 5 (filled diamonds). (f) Compound and dose loss gains of the same selected slot-5 checkpoint. These are individual recorded measurements; both held-out partitions belong to the sci-Plex development source.}
\label{fig:feedback-case}
\end{figure}
\paragraph{What the first revision actually changes.}
In the first trajectory, slot 2 is the incumbent after its child slot 3 fails to improve selection PCC. Before slot 4, the actual prompt contains slot 2's best epoch (5), sampled learning curve, group-error diagnostics, source-check findings, and positive compound and dose replacement effects. The model hypothesizes overfitting and proposes reducing encoder capacity; compound memorization is a proposed explanation, not a separately measured biological or learning mechanism. Inspection of the emitted source verifies that each chemical/context encoder changes from two Linear layers to one, with dropout 0.1 added in the encoders and fusion. Dose-conditioned FiLM and the rank-128 bilinear interaction remain. Trainable parameters decrease from 4,056,400 to 2,754,896; total parameters including the fixed anchor are 5,149,216 and 3,847,712.
Slot 2 versus slot 4 selection PCC is 0.97808647 versus 0.97818047; MSE is 0.02028296 versus 0.02019429; compound target-loss gain is 0.01846193 versus 0.01949683. The static checker reports syntactic return-dependency hints while leaving functional usefulness unresolved; fixed-model input replacement supplies the separate behavioral evidence.
\paragraph{Initialization and the final endpoint.}
Both first-case sources zero-initialize their output readout. Under the fixed residual wrapper this initializes the newly fitted predictor at $g(c,a)$. Slot 5 adds a dose-conditioned scalar interaction gate and an extra projected chemical feature before fusion, then wins by selection PCC. The emitted \texttt{chem\_direct} feature enters the nonlinear fusion, whereas its proposal calls it a direct additive target-space path. Slot 5 has higher selection PCC but lower compound target-loss gain than slot 4; selection and input contribution therefore remain distinct. The same frozen slot-5 checkpoint is subsequently evaluated on the two held-out sci-Plex partitions.
\begin{table}[!htbp]
\centering
\tblsetup
\caption{Complete records of the first two registered audit-feedback trajectories. PCC and MSE use the selection partition; chemical gain is the selection-set MSE increase after compound replacement. Bold PCC identifies the selected endpoint within each trajectory.}
\label{tab:feedback-case-2026091201}
\label{tab:feedback-case-2026091202}
\setlength{\tabcolsep}{3pt}
\begin{tabularx}{\linewidth}{@{}rr>{\raggedright\arraybackslash}Xrrrr@{}}
\toprule
\rowcolor{csHeader}
Slot & Parent & Recorded design change & PCC & MSE & \shortstack{Chemical\\gain} & Repairs \\
\midrule
\rowcolor{csHeader}\multicolumn{7}{@{}l}{\textbf{Trajectory 2026091201}} \\
0 & -- & Common context-plus-dose predictor & 0.968005 & 0.029461 & 0.000000 & 0 \\
1 & 0 & Chemical residual with dose FiLM and multiplicative fusion & 0.977915 & 0.020443 & 0.017382 & 0 \\
2 & 1 & Rank-128 bilinear interaction; 256-dimensional encoders & 0.978086 & 0.020283 & 0.018462 & 0 \\
3 & 2 & Four rank-64 interaction heads and dose gates & 0.977964 & 0.020405 & 0.018360 & 0 \\
4 & 2 & Single-layer encoders and dropout 0.1 & 0.978180 & 0.020194 & 0.019497 & 1 \\
5 & 4 & Scalar interaction gate; extra chemical feature before fusion & \textbf{0.978276} & 0.020115 & 0.017901 & 1 \\
\addlinespace[3pt]
\rowcolor{csHeader}\multicolumn{7}{@{}l}{\textbf{Trajectory 2026091202}} \\
0 & -- & Common context-plus-dose predictor & 0.967972 & 0.029506 & 0.000000 & 0 \\
1 & 0 & Gated multiplicative residual; scale 0.2 & 0.977817 & 0.020529 & 0.016928 & 0 \\
2 & 1 & Low-rank chemical FiLM and bilinear context interaction & 0.977703 & 0.020634 & 0.018378 & 0 \\
3 & 2 & Triple-product fusion plus auxiliary FiLM/bilinear terms & 0.977740 & 0.020608 & 0.017446 & 0 \\
4 & 1 & Return to slot 1; bottleneck and auxiliary FiLM & \textbf{0.977826} & 0.020533 & 0.017235 & 1 \\
5 & 4 & Add three pairwise projections to slot 4 & 0.977506 & 0.020827 & 0.017108 & 2 \\
\bottomrule
\end{tabularx}
\end{table}

\paragraph{The second registered trajectory retains non-improvements.}
For seed 2026091202, slot 2 decreases selection PCC from its slot-1 parent (0.977817 to 0.977703). Slot 3 improves on slot 2 but remains below slot 1. Slot 4 explicitly returns to slot 1 as its design parent and is selected at PCC 0.977826; its MSE (0.02053288) is slightly worse than slot 1's (0.02052894), consistent with PCC-first selection. Slot 5 adds explicit pairwise projections but falls to PCC 0.977506 and is not selected. Its rationale claims that zero-initialized new terms preserve the parent, but the shared from-scratch training contract does not inherit fitted parent weights: this claim is preserved as an agent statement in the public proposal, not endorsed as a measured guarantee.
\paragraph{Recorded execution repairs.}
The first case uses one JSON-contract repair at each of slots 4 and 5; the second uses one at slot 4 and two at slot 5. These logged failures are completion-token-limit/JSON-contract failures. Lower-scoring, successfully trained candidates are retained without score-driven repair. The case package preserves these outcomes and the actual branch structure.
\begin{table}[!htbp]
\centering
\tblsetup
\caption{Frozen endpoints of the first two registered audit-feedback trajectories. The same selected checkpoint is used at both boundaries within the sci-Plex development source. These are individual observations, not policy-mean intervals or qualification counts.}
\label{tab:feedback-case-endpoints}
\begin{tabularx}{\linewidth}{@{}lXrrrr@{}}
\toprule
\rowcolor{csHeader}
Seed / slot & Boundary & PCC & MSE & \shortstack{Chemical\\loss gain} & \shortstack{Dose\\loss gain} \\
\midrule
2026091201 / 5 & Held-out audit & 0.978061 & 0.020278 & 0.021173 & 0.008081 \\
2026091201 / 5 & Final held-out & 0.977854 & 0.020495 & 0.007978 & 0.004612 \\
2026091202 / 4 & Held-out audit & 0.978089 & 0.020252 & 0.023142 & 0.008650 \\
2026091202 / 4 & Final held-out & 0.977718 & 0.020619 & 0.007835 & 0.004303 \\
\bottomrule
\end{tabularx}
\end{table}

\FloatBarrier

\end{document}